\documentclass[letterpaper, 10pt, journal, twoside]{IEEEtran}

\IEEEoverridecommandlockouts                              % This command is only needed if 
\usepackage{cite}
\usepackage{amsmath,amssymb,amsfonts}
\usepackage{algorithmic}
\usepackage{graphicx}
\usepackage{textcomp}
\usepackage[table]{xcolor}

\usepackage{graphics} % for pdf, bitmapped graphics files
\usepackage{epsfig} % for postscript graphics files
\usepackage{times} % assumes new font selection scheme installed
\usepackage{amsmath} % assumes amsmath package installed
\usepackage{amssymb}  % assumes amsmath package installed
\usepackage{amsfonts}
\usepackage{BOONDOX-calo}
\usepackage[colorlinks,linkcolor=red,anchorcolor=blue,citecolor=blue,urlcolor=black]{hyperref}
\usepackage[switch]{lineno} % 里面的选项代表双栏
\usepackage{booktabs}
\usepackage{multirow}
\usepackage{graphicx}
\usepackage{subfigure}
\usepackage{cite}
\usepackage{arydshln}
\usepackage[ruled]{algorithm2e}
\usepackage{booktabs}
\usepackage{babel}
\usepackage{threeparttable}    %这行要添加
\newtheorem{remark}{\textbf{Remark}}

\def\bfa{\mathbf{a}}
\def\bfb{\mathbf{b}}

\def\bff{\mathbf{f}}
\def\bfg{\mathbf{g}}
\def\bfh{\mathbf{h}}

\def\bfn{\mathbf{n}}
\def\bfp{\mathbf{p}}

\def\bfl{\mathbcal{l}}

\def\bfu{\mathbf{u}}
\def\bfv{\mathbf{v}}
\def\bfx{\mathbf{x}}
\def\bfy{\mathbf{y}}

\def\bfA{\mathbf{A}}

\def\bfC{\mathbf{C}}
\def\bfE{\mathbf{E}}
\def\bfF{\mathbf{F}}
\def\bfG{\mathbf{G}}
\def\bfH{\mathbf{H}}
\def\bfI{\mathbf{I}}
\def\bfJ{\mathbf{J}}
\def\bfK{\mathbf{K}}
\def\bfN{\mathbf{N}}
\def\bfM{\mathbf{M}}

\def\bfP{\mathbf{P}}
\def\bfQ{\mathbf{Q}}
\def\bfR{\mathbf{R}}

\def\bfT{\mathbf{T}}
\def\bfV{\mathbf{V}}

\def\bfZo{\mathbf{0}}

\def\bfXi{\boldsymbol{\Xi}}
\def\bfPi{\boldsymbol{\Pi}}

\def\bfomega{\boldsymbol{\omega}}
\def\bftheta{\boldsymbol{\theta}}

\def\bfepsilon {\boldsymbol{\epsilon}}
\def\bfphi{\boldsymbol{\phi}}
\def\bfPhi{\boldsymbol{\Phi}}
\def\upG{}

\def\skew#1{\big [#1 \big]_\times}
\def\textrr#1{{#1}}
\def\textrj#1{{#1}}
\def\textac#1{{#1}}

\def\R3{{\mathbb{R}^3}}
\def\RS3{{\mathbb{R}^{3\times 3}}}
\def\SO3{{\mathbf{SO}(3)}}
\def\so3{\mathfrak{s} \mathfrak{o}(3)}

\usepackage{etoolbox}
\makeatletter
\patchcmd{\@makecaption}
  {\scshape}
  {}
  {}
  {}
\makeatletter
\patchcmd{\@makecaption}
  {\\}
  {.\ }
  {}
  {}
\makeatother

\makeatletter
\renewcommand\normalsize{%
 \@setfontsize\normalsize\@xpt\@xiipt
 \abovedisplayskip 8\p@ \@plus2\p@ \@minus8\p@
 \abovedisplayshortskip \z@ \@plus10\p@
 \belowdisplayshortskip 10\p@ \@plus10\p@ \@minus3\p@
 \belowdisplayskip \abovedisplayskip
 \let\@listi\@listI}
\makeatother

\UseRawInputEncoding
\begin{document}

\title{T-ESKF: Transformed Error-State Kalman Filter for Consistent Visual-Inertial Navigation}

% November 22 

\author{Chungeng Tian\textsuperscript{*}, Ning Hao\textsuperscript{*}, and Fenghua He 
% This work was supported by (organizations/grants which supported the work.) $^{\ast}$
\thanks{* Equal contribution. Corresponding author: {Fenghua He}. The authors are with School of Astronautics, Harbin Institute of Technology, Harbin, 150000, China. {(email: tcghit@outlook.com; haoning0082022@163.com; hefenghua@hit.edu.cn)}}%
}

\markboth{IEEE Robotics and Automation Letters. Preprint Version.}
{Tian \MakeLowercase{\textit{et al.}}: T-ESKF} 

\maketitle
%%%%%%%%%%%%%%%%%%%%%%%%%%%%%%%%%%%%%%%%%%%%%%%%%%%%%%%%%%%%%%%%%%%%%%%%%%%%%%%%
\begin{abstract}

This paper presents a novel approach to address the inconsistency problem caused by observability mismatch in visual-inertial navigation systems (VINS). The key idea involves applying a linear time-varying transformation to the error-state within the Error-State Kalman Filter (ESKF). This transformation ensures that \textrr{the unobservable subspace of the transformed error-state system} becomes independent of the state, thereby preserving the correct observability of the transformed system against variations in linearization points. We introduce the Transformed ESKF (T-ESKF), a consistent VINS estimator that performs state estimation using the transformed error-state system. \textrj{Furthermore, 
% by utilizing the transformation relationship between the original and transformed Jacobians,
we develop an efficient propagation technique to accelerate the covariance propagation based on the transformation relationship between the transition and accumulated matrices of T-ESKF and ESKF.
% by reducing the dimensionality of the matrices involved in multiplication via the designed transformation.
% in light of the derived transformation relationship between ESKF and 
}
We validate the proposed method through extensive simulations and experiments, 
demonstrating better (or competitive at least) performance compared to state-of-the-art methods.
The code is available at
        \href{https://github.com/HITCSC/T-ESKF}{github.com/HITCSC/T-ESKF}.
\end{abstract}


\section{Introduction}

In recent years, visual-inertial navigation systems (VINS) have found widespread applications in fields such as robotics, virtual reality, and augmented reality due to their compact size and low cost. By leveraging the complementary characteristics of inertial measurement units (IMUs) and cameras in navigation and localization tasks, VINS can provide high-precision pose information.
VINS can be categorized into two main approaches: optimization-based methods and filter-based methods. Optimization-based methods achieve high localization accuracy through nonlinear optimization, while filter-based methods prioritize computational efficiency. Notably, filter-based methods have recently made significant progress, achieving accuracy levels comparable to those of optimization-based methods \cite{genevaOpenVINSResearchPlatform2020}.
% Thus, in this paper, we focus on filter-based methods, especially those based on the Extended Kalman Filter (EKF).

Classical filter-based VINS estimators,
such as the Error-State Kalman Filter (ESKF)\cite{eskf}, represent robot orientation using the {\it special orthogonal group}, $\SO3$. The remaining variables, including the robot's position, velocity, and landmark positions, are structured within a vector space. \textrr{While ESKF models the state concisely, it is inherently inconsistent.}
In the absence of absolute information, the VINS system possesses four unobservable dimensions: global translation and rotation around the gravity direction. 
\textrr{However, ESKF mistakenly treats the rotation, which is
unobservable, as observable.}
{As a result, ESKF gains spurious information about this falsely observable state, leading to inconsistency} \cite{huangAnalysisImprovementConsistency2008}.

To address the inconsistency issue in VINS, two main categories of methods have been developed: observability-constrained methods\cite{huangAnalysisImprovementConsistency2008,heschConsistencyAnalysisImprovement2014,chenFEJ2ConsistentVisualInertial2022} and Lie group-based methods\cite{barrauInvariantExtendedKalman2017a,
        vangoorEquivariantFilterEqF2023,
        zhangConvergenceConsistencyAnalysis2017,
        eqvio,
        yangDecoupledRightInvariant2022
}. Among the observability-constrained approaches, the First Estimate Jacobian (FEJ) \cite{huangAnalysisImprovementConsistency2008} is quite popular due to its simplicity and ease of implementation. FEJ evaluates the linearized system state transition matrix and Jacobians at the same estimate (i.e., the initial estimates) over all time periods. However, this dependency on the initial estimates may introduce non-negligible first-order linearization errors, particularly when compared to methods that utilize the latest state estimate as the linearization point. To mitigate linearization errors, FEJ2 \cite{chenFEJ2ConsistentVisualInertial2022} was introduced.
 
In the realm of Lie group-based consistent estimation methods, the Right Invariant EKF (RI-EKF)\cite{barrauInvariantExtendedKalman2017a} and the Equivariant Filter (EqF)\cite{vangoorEquivariantFilterEqF2023}, have gained prominence within the robotics community very recently. When applied to VINS, both RI-EKF and EqF model the extended IMU poses (orientation, position, and velocity) on the {\it extended special Euclidean group}, \(\mathbf{SE}_2(3)\). A primary distinction between them lies in their treatment of landmark position variables. In RI-EKF, the extended IMU state, along with \(m\) landmark positions, is modeled on \(\mathbf{SE}_{2+m}(3)\), which inherently satisfies observability constraints without requiring any modifications to the Jacobian matrix\cite{zhangConvergenceConsistencyAnalysis2017}. In contrast, the EqF models landmark states on the {\it scaled orthogonal transform group}, $\mathbf{SOT}(3)$, and effectively eliminates the second-order linearization errors of visual measurement functions\cite{eqvio}.

% However, due to the particularity of these manifolds, it is quite challenging to maintain consistency without significantly increasing computational cost. 
% To address these issue, \cite{Chen2024}  decouples the error and state representations (DES), which demonstrates signiﬁcant
% improvements in computational eﬃciency.
% In \cite{Chen2024}, 
% while the computational cost is not significantly increased
% utilize arbitrary state representations while maintaining consistency\cite{ccchen2023}. 
% For instance, {\it global} feature representation requires special acceleration techniques \cite{yangDecoupledRightInvariant2022,Chen2024}.
% (for instance, a global feature representation \cite{yangDecoupledRightInvariant2022} requires special attention)\cite{ccchen2023}.

\textrr{
Beyond the above two main categories, Robo-Centric \cite{huai2022robocentric} ensures consistency by reformulating the VINS with respect to a moving local frame, rather than a fixed global frame.
A common point between Robo-Centric and Lie group-based methods in ensuring consistency is that they redefine states on manifolds, making the unobservable subspace independent of the states.}
\textac{
It is noteworthy that \cite{Chen2024} identifies the connections among these formulations and decouples the error and state representations (DES) to broaden the design space for the VINS estimator, which also significantly improves the computational efficiency of RI-EKF.}

In this paper, we propose a novel approach to address the inconsistency caused by observability mismatch in VINS.
The key idea is to apply a transformation to the error-state system to ensure that the transformed error-state system has correct observability. \textrr{Specifically, a transformation is designed to ensure that the unobservable subspace of the transformed system is independent of the state, thus not being affected by changes in linearization points.}
We further present Transformed ESKF (T-ESKF), a consistent VINS estimator that performs state estimation utilizing the transformed error-state system. 
\textrj{
Due to the coupling of landmark uncertainties with IMU pose, T-ESKF encounters the same computational bottleneck in covariance propagation as RI-EKF.
Based on the transformation relationship between the transition and accumulated matrices of T-ESKF and ESKF,
we reduce the dimensions of the matrices involved in multiplication to a fixed and manageable size, thereby achieving efficient covariance propagation.}
% decouping large transition and accumulation noise matrices into smaller fixed-size matrices through transformation.} 

\textrj{Compared to the Lie group-based method and the Robo-Centric method, the proposed approach offers a novel perspective to address the inconsistency issue arising from the observability mismatch.} Our method directly transforms the error-state system, which is fundamentally linked to achieving consistency. It only requires the unobservable subspace of the transformed system to be independent of the state, thereby enabling various degrees of freedom for designing transformations.

%only requires the unobservable subspace of the transformed system to be independent of the state, which 

% to sovle its computational bottleneck caused by coupling landmarks' uncertainties with IMU, since T-ESKF and RI-EKF share the same linearized error-state system.
% Extensive simulations and experiments demonstrate that the proposed approach exhibits competitive performance compared to state-of-the-art methods.

The main contributions of this paper are summarized as follows:
\begin{itemize}
        \item We propose a transformation-based approach to address
        the inconsistency issue in VINS. A linear time-varying
        transformation is designed to make \textrr{the unobservable subspace of the transformed system} state-independent.
        \item We present T-ESKF, a consistent VINS estimator that performs state estimation based on the transformed linearized error-state system. 
        \item \textac{ We derive an efficient propagation technique from the basis definition of transformations, which is also applicable to RI-EKF.}
        % by reducing the dimensionality of the matrices involved in multiplication via the designed transformation.based on the transformation relationship between the transition and accumulated matrices of T-ESKF and ESKF}
        \item \textrr{We analytically prove that T-ESKF has correct observability with the optimal Jacobians.
        Thus, T-ESKF mitigates the observability mismatch issue and guarantees consistent estimation results.}
\end{itemize}

% The remainder of the paper is organized as follows:  Section \ref{sec:2} outlines the VINS model and the inconsistency issue. 
%  Section \ref{sec:tless} presents the transformed error-state system. Section \ref{sec:teskf} proposes T-ESKF, a consistent VINS estimator. Monte-Carlo simulations and real-world experiments are conducted in Section \ref{sec:simulations} and Section \ref{sec:exp}, respectively. Section \ref{sec:concl} concludes this paper.

\section{Problem Statement}
\label{sec:2}
% \subsection{System state vector}
\subsection{System model}
\label{sys:VINS}
The system state is denoted as follows:
% \footnotetext[1]{
% Throughout this paper, lowercase letters (e.g., $\bfa$) denote vectors, and uppercase letters (e.g., $\bfA$) denote matrices. $\dot{\bfa}$ and $\dot{\bfA}$ represents their derivatives with respect to time. $\hat{\bfa}$ indicates the estimate of $\bfa$.
% }
\begin{align}
        \bfx = (\upG{\bfR},\upG\bfp,\upG\bfv,\upG\bfl_1,\cdots,\upG\bfl_m)
\end{align}
where $\upG{\bfR} \in \SO3$  and $\upG\bfp \in \R3$  are IMU's orientation and position in the global frame, respectively, $\upG\bfv \in \R3$ is the IMU's velocity expressed in the global frame, and $\upG\bfl_1, \dots, \bfl_m \in \R3$ are the landmark positions in the global frame. For simplicity, only one landmark $\bfl$ is employed to formulate the state in the sequel. Besides, to make the model more concise, the IMU bias is not included as a state variable, as it is unrelated to the inconsistency issue under consideration. We provide the 
model encompassing the consideration of IMU bias in our supplemental material\footnotemark[1].
\footnotetext[1]{
Online at \href{https://github.com/HITCSC/T-ESKF/tree/main/teskf_doc}{https://github.com/HITCSC/T-ESKF/tree/main/teskf\_doc}.
}
%The results of this paper can be easily extended to encompass the consideration of IMU bias (please refer to the supplemental material).

\subsubsection*{Motion model}
The continuous-time motion model is
\begin{subequations}
        \begin{align}
                \dot{\bfx} & = \begin{pmatrix}
                        \upG{\bfR} \skew{\bfomega_m- \bfn_{g}} \\
                        \bfv                                   \\
                        \upG{\bfR}(\bfa_m - \bfn_{a}) + \bfg   \\
                        \bf0
                \end{pmatrix}      \\
                           & \triangleq  \bff(\bfx,\bfu,\bfn)
        \end{align}
        \label{sys:f}%
\end{subequations}
where
$\bfu = \begin{bmatrix}
                \bfomega_m^\top,  \bfa_m^\top
        \end{bmatrix}^\top$, $\bfomega_m \in \R3 $ and $\bfa_m \in \R3 $ are the gyroscope and accelerometer measurements in the IMU frame, $\bfn= \begin{bmatrix}
                \bfn_{g} ^\top, \bfn_{a}^\top
        \end{bmatrix} ^\top$, $\bfn_{g} \in \R3 $ and $\bfn_{a} \in \R3 $ are the zero-mean Gaussian white noises with $\bfE(\bfn\bfn ^\top )=\bfQ $,
$\bfg$ is the gravity vector in the global frame.

\subsubsection*{Landmark observation}
Let ${^I}\bfp_L$ denote the landmark position in the IMU frame, expressed as
\begin{equation}
        {^I}\bfp_L = \bfR^\top (\bfl - \bfp).
\end{equation}
As the camera explores the environment, the visual measurement of the landmark is available after the data association and rectification, formulated as
\begin{align}
        \bfy =\bfh({^I}\bfp_{L}) + \bfepsilon
        \label{sys:h}%   
\end{align}
where $\bfh = \pi \circ \varUpsilon $,
$ \varUpsilon:  \mathbb{R}^3 \to \mathbb{R}^3$ transforms points from the IMU frame to the camera frame, and
$\pi: \mathbb{R}^3 \to \mathbb{R}^2$ is the camera perspective projection function,
$\bfepsilon \in \mathbb{R}^2$ is the zero-mean Gaussian noise with $\bfE(\bfepsilon \bfepsilon^\top) = \bfV$.

\subsection{Linearized error-state system}

Let $\hat{\bfx}$ denote the state estimate, which is propagated following \eqref{sys:f} by setting the process noise to zero:
\begin{equation}
        \dot{\hat{\bfx}}= \bff(\hat{\bfx},\bfu,\bf0).
        \label{equ:ode1}
\end{equation}
Let $\tilde{\bfx}$ represent the error-state, describing the difference between the true state $\bfx$ and its estimate $\hat{\bfx}$. In ESKF, the error-state of orientation is defined using the logarithm map on $\SO3$ while the errors of other variables are defined in vector space:
\begin{align}
        \tilde{\bfx}= \bfx \ominus \hat{\bfx}=
        \begin{bmatrix}
                \text{Log}(\bfR \hat{\bfR}^{-1}) \\
                \bfp- \hat{\bfp}                 \\
                \bfv- \hat{\bfv}                 \\
                \bfl- \hat{\bfl}                 \\
        \end{bmatrix}
        \triangleq \begin{bmatrix}
                \tilde{\bftheta} \\
                \tilde{\bfp}     \\
                \tilde{\bfv}     \\
                \tilde{\bfl}     \\
        \end{bmatrix}.
        \label{equ:ominus}
\end{align}
The measurement residual is denoted by
\begin{equation}
        \tilde{\bfy} =  \bfy- \bfh({^I}\hat{\bfp}_L).
\end{equation}
Linearizing \eqref{sys:f} and \eqref{sys:h} at the current state estimate, we obtain the linearized error-state system given by (details please refer to the supplementary material):
\begin{equation}
        \label{sys:original}
        \left\{
        \begin{aligned}
                \dot{\tilde{\bfx}} & = \bfF \tilde{\bfx} + \bfG \bfn  \\
                \tilde{\bfy}       & =  \bfH\tilde{\bfx} + \bfepsilon
        \end{aligned}
        \right.
\end{equation}
where the state propagation Jacobian is
\begin{align}
        {\bfF} & =
        \left[
                \begin{array}[]{c:ccc}
                        \bf0                                                  & \bf0 & \bf0   & \bf0 \\
                        \hdashline
                        \cellcolor{green!20}  \bf0                            & \bf0 & \bfI_3 & \bf0 \\
                        \cellcolor{green!20}-\skew{ \upG{\hat{\bfR}}\bfa_m  } & \bf0 & \bf0   & \bf0 \\
                        \cellcolor{green!20}  \bf0                            & \bf0 & \bf0   & \bf0 \\
                \end{array}
                \right],
        \label{equ:F}
\end{align}
the noise propagation Jacobian is
\begin{equation}
        {\bfG}   = \begin{bmatrix}
                -\upG{\hat{\bfR}} & \bf0              \\
                \bf0              & \bf0              \\
                \bf0              & -\upG{\hat{\bfR}} \\
                \bf0              & \bf0              \\
        \end{bmatrix},
\end{equation}
and the measurement Jacobian is
\begin{equation}
        \bfH   = \bfPi  \bfH_{{{e}}}
        \label{equ:H}
\end{equation}
with $\bfPi=  \frac{\partial \bfh}{\partial ^I\bfp_L}
        \hat{ \bfR}^\top$ and
\begin{equation}
        \bfH_{{{e}}} =       \Big[
                \begin{array}[]{c:ccc}
                        \cellcolor{green!20} \skew{\upG \hat\bfl- \upG \hat\bfp } & -\bfI_{3} & \bf0 & \bfI_3
                \end{array}
                \Big].
        \label{equ:bfH_v}
\end{equation}
We call $\bfH_{{{e}}}$ the {\it essential measurement Jacobian}. Note that the green-highlighted block in $\bfF$ and $\bfH_{{{e}}}$ are related to states.

\subsection{Inconsistency of Error-State Kalman Filter}
\label{sec:inconsistency}
Observability refers to the ability of a system to recover its initial states using all available measurements. The set of states that cannot be recovered from measurements constitutes the unobservable subspace of the system. For the system described in \eqref{sys:original}, we can utilize the local observability matrix \cite[P. 180]{chenLinearSystemTheory1999} to conduct the observability analysis. Let $\bfM$ denote the local observability matrix of system \eqref{sys:original}, then 
\begin{equation}
        \bfM
        = \begin{bmatrix}
                \bfM_0 ^\top    &
                \bfM_1 ^\top    &
                \cdots     &
                \bfM_{n-1}^\top 
                
        \end{bmatrix}^\top
        \label{equ:M}
\end{equation}
where $\bfM_0 = \bfPi  \bfH_{{{e}}}$ and
\begin{equation}
        \bfM_{k+1} =   \bfM_{k} \bfF + \dot{\bfM}_k, \quad k = 0, 1,..., n-1.
        \label{equ:Mk}
\end{equation}

We first assume that both $\bfF$ and $\bfH_{{{e}}}$ are evaluated at the same linearization point. While this assumption is idealized, it is widely used to demonstrate the inconsistency issue in VINS \cite{huangAnalysisImprovementConsistency2008,ZhangMapBased}.
The right null space $\bfN$ of the observability matrix $\bfM$, satisfying $ \bfM \bfN=\bf0$, represents the unobservable subspace of the system. The unobservable subspace is:
\begin{equation}
        \bfN = \left[
                \begin{array}[pos]{c:c}
                        \bfZo_{3\times 3 } & \bfg                    \\
                        \bfI_{3}          & -\skew{\hat{\bfp}} \bfg \\
                        \bfZo_{3\times 3 } & -\skew{\hat{\bfv}} \bfg \\
                        \bfI_{3}          & -\skew{\hat{\bfl}} \bfg \\
                \end{array}
                \right].
\end{equation}
$\bfN$ indicates that VINS has four unobservable directions: the global position (represented by the first three columns) and the global orientation around the $z$-axis (represented by the last column).

However, the above assumption is not practical for ESKF. ESKF evaluates $\bfF$ at the posterior estimates and $\bfH_{{{e}}}$ at the prior estimates, respectively. According to \cite{ZhangMapBased}, evaluating the Jacobians at two different state trajectories would lead to
the unobservable direction, which depends on the state, to disappear from the unobservable subspace. Thus, the unobservable subspace of ESKF is
\begin{equation}
        \bfN_{\text{ESKF}} = \left[
                \begin{array}[pos]{cccc}
                        \bfZo_{3\times 3 } &
                        \bfI_{3}          &
                        \bfZo_{3\times 3 } &
                        \bfI_{3}            \\
                \end{array}\right]^\top,
\end{equation}
where the global orientation around the $z$-axis is falsely observable in ESKF, leading to inconsistency. Note that the unobservable directions unrelated to states remain invariant to changes in linearization points.

\begin{figure}[!ht]
        \centering
        \includegraphics[width=0.8\linewidth]{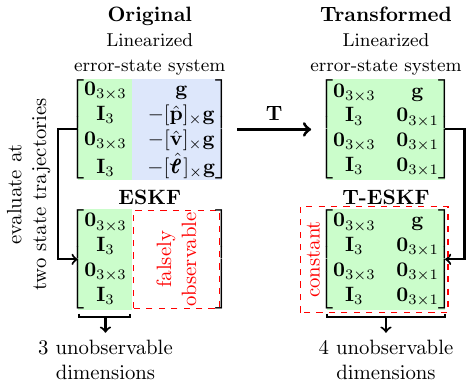}
        \caption{The unobservable subspace of the original system depends on the state, while that of the transformed system is \emph{state-independent}. The dimension of ESKF unobservable subspace is reduced by one (with the global rotation in ESKF becoming erroneously observable). In contrast, the unobservable subspace of T-ESKF remains invariant to changes in linearization points.}
        \label{fig:rank}
\end{figure}

The aim of this paper is to propose a transformation-based approach to address inconsistency caused by observability mismatch in VINS.
We design a linear time-varying transformation applied to the error-state, as shown in Fig. \ref{fig:rank}, which makes the unobservable subspace independent of states, thereby preserving the correct observability against variations in linearization points.

\section{Transformed Linearized Error-State System}
\label{sec:tless}

In this section, we address the inconsistencies in VINS by transforming the linearized error-state system to achieve a state-independent unobservable subspace. We begin by deriving the transformed linearized error-state system.  \textrr{Subsequently, to prevent the observability from being affected by variations in linearization points, we design a transformation that makes the unobservable subspace of the transformed system state-independent.}

% In this section, we tackle the inconsistency in VINS by transforming the linearized error-state system to achieve a state-independent unobservable subspace. We first derive the transformed linearized error-state system.
% Then, to prevent the observability being affected by variations in linearization points,
% a transformation is designed to make the unobservable subspace of the transformed system state-independent. 
% In this section, we tackle the inconsistency issue in VINS by transforming the linearized error-state system. 
% We start by utilizing an undetermined linear time-varying transformation on the linearized error-state system. 
% Then, the transformation is designed to ensure the independence of the state propagation Jacobian and essential measurement Jacobian from the state variables.

\subsection{Linear time-varying transformation on error-states}
Denote the linear time-varying transformation by $\bfT(\hat{\bfx})$, where $\bfT(\cdot)$ is a $12 \times 12$ nonsingular matrix function that remains to be chosen. The transformed error-state is obtained by multiplying $\bfT(\hat{\bfx})$ with the original error-state:
\begin{equation}
        \tilde{ \bfx}^* = \bfT(\hat{\bfx}) \tilde{ \bfx}.
        \label{equ:trans_x}
\end{equation}
For simplicity, the variable $\hat{\bfx}$ of $\bfT(\hat{\bfx})$ is omitted in the sequel if there is no ambiguity. Taking the derivative of both sides of \eqref{equ:trans_x} with respect to time, we have
\begin{equation}
        \dot{\tilde{ \bfx}}^* = \dot{\bfT}\tilde{ \bfx} + {\bfT}\dot{\tilde{{ \bfx}}}.
        \label{equ:trans_der}
\end{equation}
% where $\dot{\bfT}$ is the derivative of $\bfT$ with respect to time.
Substituting  \eqref{equ:trans_x} and \eqref{equ:trans_der} into  system \eqref{sys:original} yields the transformed linearized error-state system as follows:
% \begin{subequations}
%         \begin{align}
%                 \dot{  \tilde{ \bfx}}^* & = {\bfF^*}\tilde{ \bfx}^* +{\bfG^*} \bfn   \\
%                 \tilde{\bfy}            & = {\bfH^*} \tilde{ \bfx}^*+ \bfepsilon .
%         \end{align}
%         \label{sys:trans}%
% \end{subequations}
\begin{equation}
        \left\{
        \begin{aligned}
                \dot{  \tilde{ \bfx}}^* & = {\bfF^*}\tilde{ \bfx}^* +{\bfG^*} \bfn \\
                \tilde{\bfy}            & = {\bfH^*} \tilde{ \bfx}^*+ \bfepsilon
        \end{aligned}
        \right.       \label{sys:trans}%
\end{equation}
where
\begin{align}
        \bfF^* & = \dot{\bfT} \bfT^{-1}+\bfT \bfF  \bfT^{-1}  \label{equ:kdF} \\
        \bfG^* & = \bfT \bfG                                      \label{equ:kdG}            \\
        \bfH^* & = \bfH \bfT^{-1}.  \label{equ:kdH}
\end{align}
According to \eqref{equ:H} and \eqref{equ:kdH}, the essential measurement Jacobian $\bfH_{{{e}}}$ is transformed into
\begin{equation}
        \bfH_{{{e}}}^* =\bfH_{{{e}}} \bfT^{-1}. \label{equ:kdHv}
\end{equation}

There are various possible selections for the transformation, leading to different $\bfF^*$ and $\bfH_{{{e}}}^*$.
\textrr{To make the unobservable subspace independent of states, 
one effective method is to make $\bfF^*$ and $\bfH_{{{e}}}^*$ state-independent.}

\subsection[]{Transformation design}
\label{sec:design}
% The transformation design is divided into two steps: designing the structure for the transformation matrix and determining the blocks.
% First, we present a transformation matrix structure designed for VINS. Following that, the sub-matrices of the transformation are chosen based on the requirement for state independence.
% \subsubsection[short]{\textbf{Structure design}}
We first establish the structure of the transformation matrix. Observing \eqref{equ:F} and \eqref{equ:bfH_v}, obviously only the green-highlighted sub-matrices, relating to orientation error, are state-dependent. To make the transformed Jacobians $\bfF^*$ and $\bfH_{{{e}}}^*$ independent of states, an intuitive idea is to transform these green-highlighted sub-matrices exclusively while leaving the remaining sub-matrices unchanged. Thus, the transformation can be designed as a lower triangular matrix with the diagonal being the identity:
\begin{equation}
        \bfT = \left[
                \begin{array}{c:ccc}
                        \bfI_3                         & \bf0   & \bf0   & \bf0   \\
                        \hdashline
                        \cellcolor{green!0}  \bfT_\bfp & \bfI_3 & \bf0   & \bf0   \\
                        \cellcolor{green!0}  \bfT_\bfv & \bf0   & \bfI_3 & \bf0   \\
                        \cellcolor{green!0}  \bfT_\bfl & \bf0   & \bf0   & \bfI_3 \\
                \end{array}
                \right]
        \label{equ:Tk},
\end{equation}
where $\bfT_\bfp , \bfT_\bfv, \bfT_\bfl$ are sub-matrices to be determined such that $\bfF^*$ and $\bfH_{{{e}}}^*$ are independent of the states.
Substituting \eqref{equ:Tk} into \eqref{equ:kdF} and \eqref{equ:kdHv}, we have 
\begin{align}
        \bfF^*   & = \left[
                \begin{array}{c:ccc}
                        \bf0                          & \bf0 & \bf0   & \bf0 \\
                        \hdashline
                        \cellcolor{green!20}   \dot{\bfT}_{\bfp} - {\bfT}_{\bfv}& \bf0 & \bfI_3 & \bf0 \\
                        \cellcolor{green!20}    \dot{\bfT}_{\bfv} - \skew{\hat{\bfR}\bfa_m}   & \bf0 & \bf0   & \bf0 \\
                        \cellcolor{green!20}  \dot{\bfT}_{\bfl} & \bf0 & \bf0   & \bf0 \\
                \end{array}
                \right],  \label{equ:bfF1} \\
        \bfH_{{{e}}}^* & =\left[
                \begin{array}{c:ccc}
                        \cellcolor{green!20} \skew{\upG \hat\bfl- \upG \hat\bfp } +{\bfT}_{\bfp} -{\bfT}_{\bfl} & -\bfI_3 & \bf0 & \bfI_3
                \end{array} \label{equ:bfHv2}
                \right].
\end{align}

To ensure $\bfF^*$ and $\bfH_{{{e}}}^*$ independent of states,  
the green-highlighted blocks in \eqref{equ:bfF1} and \eqref{equ:bfHv2} are required to be constant. 
A feasible solution to achieve this target is (details please refer to the supplementary material):
\begin{equation}
    \bfT_\bfp = [\hat{\bfp}]_\times, \quad  \bfT_\bfv= [\hat{\bfv}]_\times,\quad  \bfT_\bfl = [\hat{\bfl}]_\times.
    \label{equ:Tpvl}
\end{equation}
Substituting \eqref{equ:Tpvl} into \eqref{equ:Tk} yields the transformation matrix:
\begin{equation}
        \bfT = \left[
                \begin{array}{c:ccc}
                        \bfI_3                         & \bf0   & \bf0   & \bf0   \\
                        \hdashline
                        \cellcolor{green!0}  [\hat{\bfp}]_\times & \bfI_3 & \bf0   & \bf0   \\
                        \cellcolor{green!0}  [\hat{\bfv}]_\times & \bf0   & \bfI_3 & \bf0   \\
                        \cellcolor{green!0}  [\hat{\bfl}]_\times & \bf0   & \bf0   & \bfI_3 \\
                \end{array}
                \right].
        \label{equ:Tk2}
\end{equation}
The transformed Jacobians are obtained by substituting \eqref{equ:Tpvl} into \eqref{equ:bfF1}-\eqref{equ:bfHv2}: 
\begin{align}
        \bfF^*   & = \left[
                \begin{array}{c:ccc}
                        \bf0                          & \bf0 & \bf0   & \bf0 \\
                        \hdashline
                        \cellcolor{green!20}   \bf0& \bf0 & \bfI_3 & \bf0 \\
                        \cellcolor{green!20}    \skew{\bfg}   & \bf0 & \bf0   & \bf0 \\
                        \cellcolor{green!20}  \bf0 & \bf0 & \bf0   & \bf0 \\
                \end{array}
                \right],   \label{equ:bfF1final} \\
        \bfH_{{{e}}}^* & =\left[
                \begin{array}{c:ccc}
                        \cellcolor{green!20} \bf0 & -\bfI_3 & \bf0 & \bfI_3
                \end{array} \label{equ:bfHv2final}
                \right].
\end{align}

\textrr{
Now, we can verify that the unobservable subspace of the transformed system \eqref{sys:trans} is independent of states.
The local observability matrix $\bfM^*$ of the transformed system is computed by 
replacing $(\bfF, \bfH_{{e}})$ in \eqref{equ:M}-\eqref{equ:Mk} with $(\bfF^*, \bfH^*_{{{e}}})$ in \eqref{equ:bfF1final}-\eqref{equ:bfHv2final}. The unobservable subspace of the transformed system with $\bfM^*\bfN^*=\bf0$ is (details please refer to the supplementary material):
\begin{equation}
        \bfN^* = \begin{bmatrix}
                \bf0_{3\times 3 } & \bfI_{3} & \bf0_{3\times 3 }     &\bfI_{3}         \\
                \bfg &\bfZo_{1\times 3} &\bfZo_{1\times 3} &\bfZo_{1\times 3} \\
        \end{bmatrix}^\top.
        \label{equ:N*}
\end{equation}
\eqref{equ:N*} indicates that the designed transformation successfully ensures 
a state-independent unobservable subspace.}

\textrr{
\begin{remark}
Designing a transformation that makes the Jacobians independent of states is not the only approach for ensuring the state-independent unobservable subspace. Actually, various solutions can be further explored to maintain the constancy of the unobservable subspace.
\end{remark}}

The transformed linearized error-state system \eqref{sys:trans} along with these transformed Jacobians \eqref{equ:bfF1final} and \eqref{equ:bfHv2final} will be used for state estimation in Section \ref{sec:teskf}.

\section{Transformed Error-State Kalman Filter}
\label{sec:teskf}
\begin{figure}[ht]
        \centerline{\includegraphics[width=0.475\textwidth]{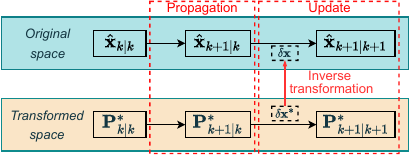}}
        \caption{Pipeline of T-ESKF.
                It propagates and updates the covariance estimates in the transformed space. The state estimate is propagated by integrating \eqref{equ:ode1} and updated using the state correction derived from the transformed system. 
        }
        \label{fig:frames}
\end{figure}

In this section, we present T-ESKF, a consistent VINS estimator based on the transformed linearized error-state system \eqref{sys:trans}.
The pipeline of T-ESKF is shown in Fig. \ref{fig:frames}.
\textrr{In T-ESKF, we only store and compute the estimate and the transformed error-state covariance. According to \eqref{equ:trans_x}, the covariance associated with the original error-state can be obtained through the following inverse transformation:
\begin{equation}
        \bfP = \bfT^{-1} \bfP^* \bfT^{-\top} .
        \label{equ:P2P}
\end{equation}}

\subsection{Propagation}
\textrj{Before updating with the visual measurements at $t_{k+1}$, T-ESKF needs to propagate the state estimate to the current time
through the Runge-Kutta numerical integration of \eqref{equ:ode1}
using the IMU measurements accumulated from $t_k$ to $t_{k+1}$.} 
Correspondingly, the transformed error-state covariance is propagated as follows:
\begin{equation}
        {\bfP_{k+1|k}^*}    =\bfPhi^*_{k}{\bfP_{k|k}^*} {\bfPhi^*_{k}}^\top + \bfQ_{k}^*
        \label{equ:proP}
\end{equation}
where the transition matrix $\bfPhi^*_{k}\triangleq\bfPhi^*(t_{k+1},t_k)$ is computed by integrating the differential equation:
\begin{equation}
        \dot{\bfPhi}^*(\tau,t_k) = \bfF^*{\bfPhi}^*(\tau,t_k)
        \label{equ:int_PHI}
\end{equation}
with the initial condition ${\bfPhi}^*(t_{k},t_k)= \bfI.$
The accumulated noise matrix is computed by
\begin{equation}
        \bfQ_{k}^* = \int_{t_k}^{t_{k+1}} \bfPhi ^* (t_{k+1},\tau) \bfG^*_{\tau} \bfQ {\bfG^*_{\tau} }^\top {\bfPhi ^*(t_{k+1},\tau)}^\top \text{d} \tau.
        \label{equ:intG}
\end{equation}

\textrj{
When IMU bias and a large number of landmarks are included in the state vector, directly integrating \(\bfPhi^*_k\) and \(\bfQ_{k}^*\) using \eqref{equ:int_PHI} and \eqref{equ:intG} requires performing matrix multiplication of size \((15+3m)×(15+3m)\) multiple times, where $m$ represents the number of landmarks.
To address this computational problem, 
we develop an efficient propagation technique to compute \(\bfPhi^*_k\) and \(\bfQ_{k}^*\). Specifically,
by leveraging the designed transformation, \(\bfPhi^*_k\) and \(\bfQ_{k}^*\) are transformed into the following computationally efficient forms (the details please refer to the supplementary material):
\begin{align}
      \bfPhi^*_{k} &= \bfT(\hat{\bfx}_{k+1|k}) \begin{bmatrix}
         \bfPhi_{k}^I &\bf0 \\
         \bf0 &\bfI_{3m}\\
     \end{bmatrix}\bfT(\hat{\bfx}_{k|k})^{-1}, \label{equ:38}  \\
     \bfQ_{k}^* &= \bfT(\hat{\bfx}_{k+1|k}) \begin{bmatrix}
         \bfQ_k^I &\bf0 \\
         \bf0 &\bfZo_{3m \times 3m}\\
     \end{bmatrix}\bfT(\hat{\bfx}_{k+1|k}) ^\top. \label{equ:39}
\end{align}
where $\bfPhi_{k}^I$ and $\bfQ_k^I \in \mathbb{R}^{15\times 15}$ are the IMU transition matrix and the IMU accumulated noise matrix of ESKF, respectively. 
Regardless of the number of landmarks included in the state vector, $\bfPhi_{k}^I$ and $\bfQ_k^I$ always 
maintain a fixed and manageable dimensionality, enabling their computation within a short timeframe. Additionally, the sparsity of the transformation $\bfT(\cdot)$ is fully
exploited to accelerate the matrix multiplications in \eqref{equ:38} and \eqref{equ:39}.
Consequently, T-ESKF achieves efficient covariance propagation comparable to ESKF.}

\textrj{
\begin{remark}
Equations \eqref{equ:38} and \eqref{equ:39} are derived in a continuous form and do not presuppose any assumptions regarding the IMU model, such as the piecewise constant acceleration. Therefore, we can directly apply existing IMU integration theories, such as \cite{ic2}, to compute $\bfPhi_k^I$ and $\bfQ_k^I$.
\end{remark}}

\textrj{
\begin{remark}
The proposed efficient propagation technique is also applicable to RI-EKF, as both RI-EKF and T-ESKF share the same linearized error-state system,  which implies that their transition matrices and accumulated noise matrices are identical. 
\end{remark}}

\subsection{Update}
\textrr{In T-ESKF, we begin with deriving the Kalman state correction in the transformed space, then we inversely transform it back to the original space to correct the estimate.
\subsubsection{Correction in transformed space}
Let $\delta \bfx ^*  \triangleq \bfK^*\tilde{\bfy}_{k+1}$ denote the Kalman state correction in the transformed space.
We compute the Kalman gain following the classical EKF:}
\begin{equation}
        \bfK^*  = \bfP^*_{k+1|k} {\bfH^*_{k+1}}^\top (\bfH^*_{k+1}\bfP^*_{k+1|k}{\bfH^*_{k+1}}^\top  + \bfV ) ^{-1}.
\end{equation}
\textrr{where $\bfH^*_{k+1}$ is evaluated at the current best state estimate $\hat{\bfx}_{k+1|k}$.}
The transformed error-state covariance is updated as 
\vspace{-0.2cm}
\begin{subequations}
        \begin{align}
                \bfP_{k+1|k+1}^* &= \bfE\left( (\tilde{\bfx}_{k|k+1}^* - \delta  \bfx ^* )(\tilde{\bfx}_{k|k+1}^* - \delta  \bfx ^* )^\top 
                \right) \\
                & =  \bfP_{k+1|k}^* - \bfK^* \bfH^*_{k+1} \bfP_{k+1|k}^* .
        \end{align}
\end{subequations}

\textrr{
\subsubsection{Correction in original space}
Let $\delta \bfx$ denote the correction in original space.
The error-state $\tilde{\bfx}$ is corrected by:
\begin{equation}
        \tilde{\bfx}_{k+1|k+1} =  \tilde{\bfx}_{k+1|k}  - \delta \bfx.
        \label{equ:x-x-x}
\end{equation}
According to \eqref{equ:P2P}, we have the covariance matrix associated with $ \tilde{\bfx}_{k+1|k+1}$ as follows:
\begin{equation}
        \bfP_{k+1|k+1} = \bfT(\hat{\bfx}_{k+1|k+1})^{-1} \bfP^*_{k+1|k+1} \bfT(\hat{\bfx}_{k+1|k+1})^{-\top} .
        \label{equ:P2P2}
\end{equation}
To ensure that $\bfP_{k+1|k+1}$ in \eqref{equ:x-x-x} matches $\tilde{\bfx}_{k+1|k+1}$ in \eqref{equ:P2P2}, $\delta \bfx$ should satisfy (details please refer to the supplementary material):}
\begin{equation}
        \delta {\bfx}  =  \bfT(\hat{\bfx}_{k+1|k})^{-1}  \delta {\bfx}^* \label{equ:tekf1}.
\end{equation}
Finally, the state estimate is corrected by:
\begin{align}
        \hat{\bfx}_{k+1|k+1} =   \hat{\bfx}_{k+1|k} \oplus \delta{\bfx}  = \begin{pmatrix}
                {\text{Exp}(\delta{\bftheta})}\upG\hat{\bfR}_{k+1|k} \\
                \upG\hat{{\bfp}}_{k+1|k} + \delta{\bfp}              \\
                \upG\hat{{\bfv}}_{k+1|k} + \delta{\bfv}              \\
                \upG\hat{\bfl}_{k+1|k}+ \delta\bfl
        \end{pmatrix}.
\end{align}
% \vspace{-0.5cm}

\begin{remark}
T-ESKF also integrates several widely used techniques in filter-based VINS, such as delayed feature initialization, multi-state constraint nullspace projection, and measurement compression  \cite{mourikisMultiStateConstraintKalman2007}.
The details are omitted due to space limitations.
\end{remark}

\textrr{
\subsection{T-ESKF properties}
\label{sec:property}
We now show that T-ESKF has the optimal Jacobians and correct observability.
Since the Jacobians of T-ESKF (in discrete form), $\bfPhi^*$ and $\bfH^*$, are evaluated at the current best estimates,
the optimality of Jacobians is automatically preserved.
The unobservable subspace of T-ESKF is (details please refer to the supplementary material):
\begin{equation}
        \bfN_{\text{T-ESKF}} = \begin{bmatrix}
                \bfZo_{3\times 3 } & \bfI_{3} & \bfZo_{3\times 3 }     &\bfI_{3}         \\
                \bfg &\bfZo_{1\times 3} &\bfZo_{1\times 3} &\bfZo_{1\times 3} \\
        \end{bmatrix}^\top.
        \label{equ:N_teskf}
\end{equation}
\eqref{equ:N*} and \eqref{equ:N_teskf} indicate that erroneous reductions of unobservable dimensions are prevented and  T-ESKF has the correct observability. Consequently, T-ESKF does not encounter estimation inconsistencies arising from observability mismatches.}

\section{Monte-Carlo Simulations}
\label{sec:simulations}
We assess the accuracy, consistency, and efficiency of the proposed approach using OpenVINS\cite{genevaOpenVINSResearchPlatform2020} as the testing platform. Specifically, we compare T-ESKF with the following estimators: ESKF\cite{eskf}, FEJ-ESKF\cite{huangAnalysisImprovementConsistency2008}, \textrr{Robo-Centric \cite{huai2022robocentric},} and RI-EKF\cite{zhangConvergenceConsistencyAnalysis2017}. ESKF and FEJ-ESKF are natively integrated into OpenVINS, while \textrr{Robo-Centric,} RI-EKF, and T-ESKF are implemented by ourselves.

\begin{figure}[ht]
        % \vspace{-0.5cm}
        \centering
        \includegraphics[width=0.88\linewidth]{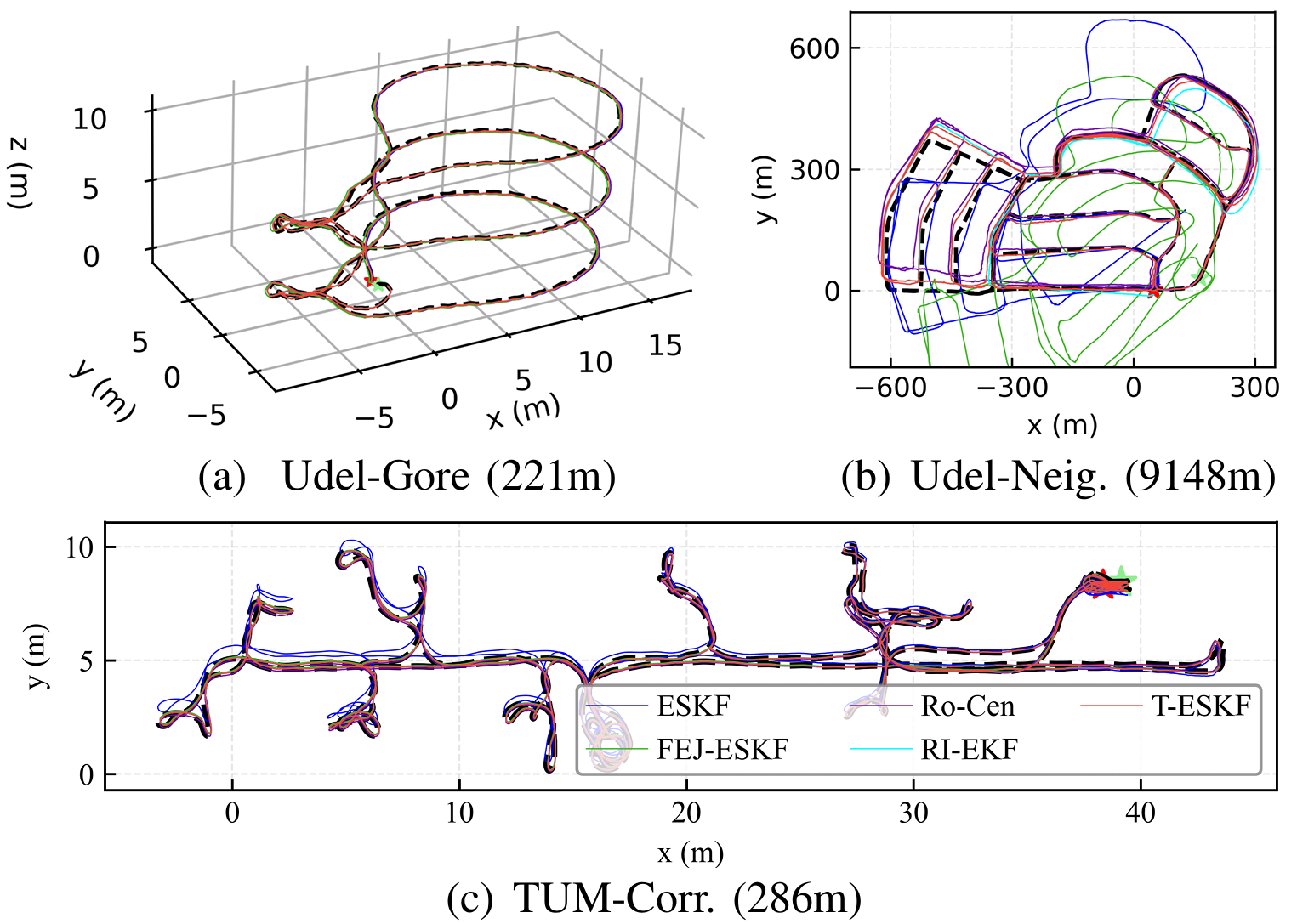}
        \caption{\textrr{Simulated trajectories (black dashed lines) with green and red stars marking the starting and ending points, respectively, and estimated trajectories (solid lines).}
        }
        \label{fig:traj_sim}
        % \vspace{-0.5cm}
\end{figure}

\begin{figure*}[t]
        \centering
        \includegraphics[width=0.99\textwidth]{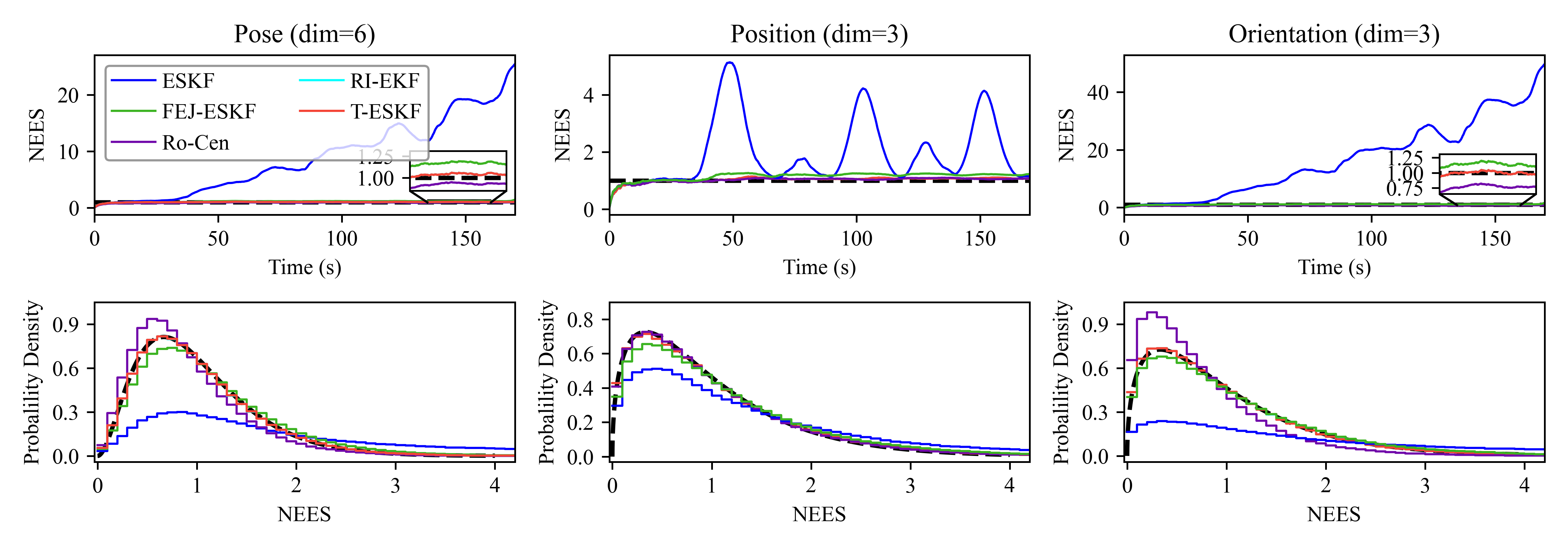}
        \vspace{-0.4cm}
        \caption{The NEES of 1000 Monte-Carlo simulation runs.  The upper three subfigures show the average NEES over time for 1000 runs. The lower three subfigures depict the frequency distribution of the NEES for 1000 runs throughout the simulation time, with the black dashed lines representing the theoretical value of the chi-square distribution. 
        The lines of T-ESKF and RI-EKF coincide. 
        }
        \label{fig:nees}
\end{figure*}

The evaluations involve testing these estimators on three trajectories collected by OpenVINS, as shown in Fig. \ref{fig:traj_sim}.
 Inertial readings and camera pixel data are generated based on these trajectories using OpenVINS's built-in simulator. Table \ref{tab:sim_params} outlines the sensor properties and key parameters of OpenVINS.

\begin{table}[htp]
        \caption{Monte-Carlo simulation configuration}
        \centering
        \setlength\tabcolsep{4.5pt}
        \begin{tabular}[]{cccc}
                \toprule
                { \textbf{Parameter}} & \textbf{Value} & \textbf{Parameter} & \textbf{Value} \\
                \midrule
                Accel. White Noise    & 2.00e-03       & Gyro. White Noise  & 1.70e-04       \\
                Accel. Random Walk    & 3.00e-03       & Gyro. Random Walk  & 2.00e-05       \\
                Pixel Noise           & 2              & IMU Freq.          & 400            \\
                Max Cam Pts/Frame     & 100            & Cam Freq.          & 10             \\
                Max SLAM feats/Frame       & 40             & Cam Number         & Mono           \\
                Max MSCKF feats/Frame        & 10             & Max Clone Size        & 11           \\
                \bottomrule
        \end{tabular}
        \label{tab:sim_params}
\end{table}

\subsection{Accuracy evaluation}
To evaluate the accuracy of these estimators, we perform 100 Monte-Carlo simulations for each trajectory. Fig. \ref{fig:traj_sim} displays the trajectories estimated by these estimators.
Table \ref{tab:sim_rmse} reports the average RMSE resulting from the 100 Monte-Carlo runs.

\begin{table}[!ht]
        \caption{Average orientation (deg) and position (meter) RMSE over 100 Monte-Carlo runs}\label{tab:tablenotes}
        \centering
        \setlength\tabcolsep{2.3pt}
        \begin{threeparttable}          %这行要添加
                \begin{tabular}[]{cccccc}
                        \toprule
                        Trajectory   & ESKF          & FEJ-ESKF               & RI-EKF                            & {T-ESKF}              & \textrr{Ro-Cen}           \\
                        \midrule
                        {Udel-Gore}  & 1.13 / 0.26 & 0.67 / 0.21          & \textbf{0.58} / \textbf{0.20} & \textbf{0.58} / \textbf{0.20} &\textrr{0.59 / 0.21} \\
                        {Udel-Neig.} & 11.6 / 72.7 & 24.3 / 153. & 6.92 / 42.2                     & {6.83} / {41.8}  & \textrr{\textbf{6.30} / \textbf{39.2}}\\
                        { TUM-Corr.} & 0.68 / 0.28 & 0.36 / 0.16          & \textbf{0.34} /  \textbf{0.15}           & \textbf{0.34} /  \textbf{0.15} &\textrr{0.36 / \textbf{0.15}}\\
                        \bottomrule
                \end{tabular}\label{tab:sim_rmse}
        %        \begin{tablenotes}    %这行要添加， 从这开始
        %                \footnotesize               %这行要添加
        %        \end{tablenotes}            %这行要添加
        \end{threeparttable}       %这行要添加，到这里结束
\end{table}
\textrr{As seen, T-ESKF demonstrates competitive performance with RI-EKF and Robo-Centric. 
Since the \textit{Udel-Neig.} trajectory is much longer than the other two trajectories, and the RMSE results within it are larger.}
Moreover, FEJ-ESKF fails in most runs on \textit{Udel-Neig.}, primarily due to its larger first-order linearization error from poor feature initialization.

\begin{figure}[!htbp]
        \centerline{\includegraphics[width=0.475\textwidth]{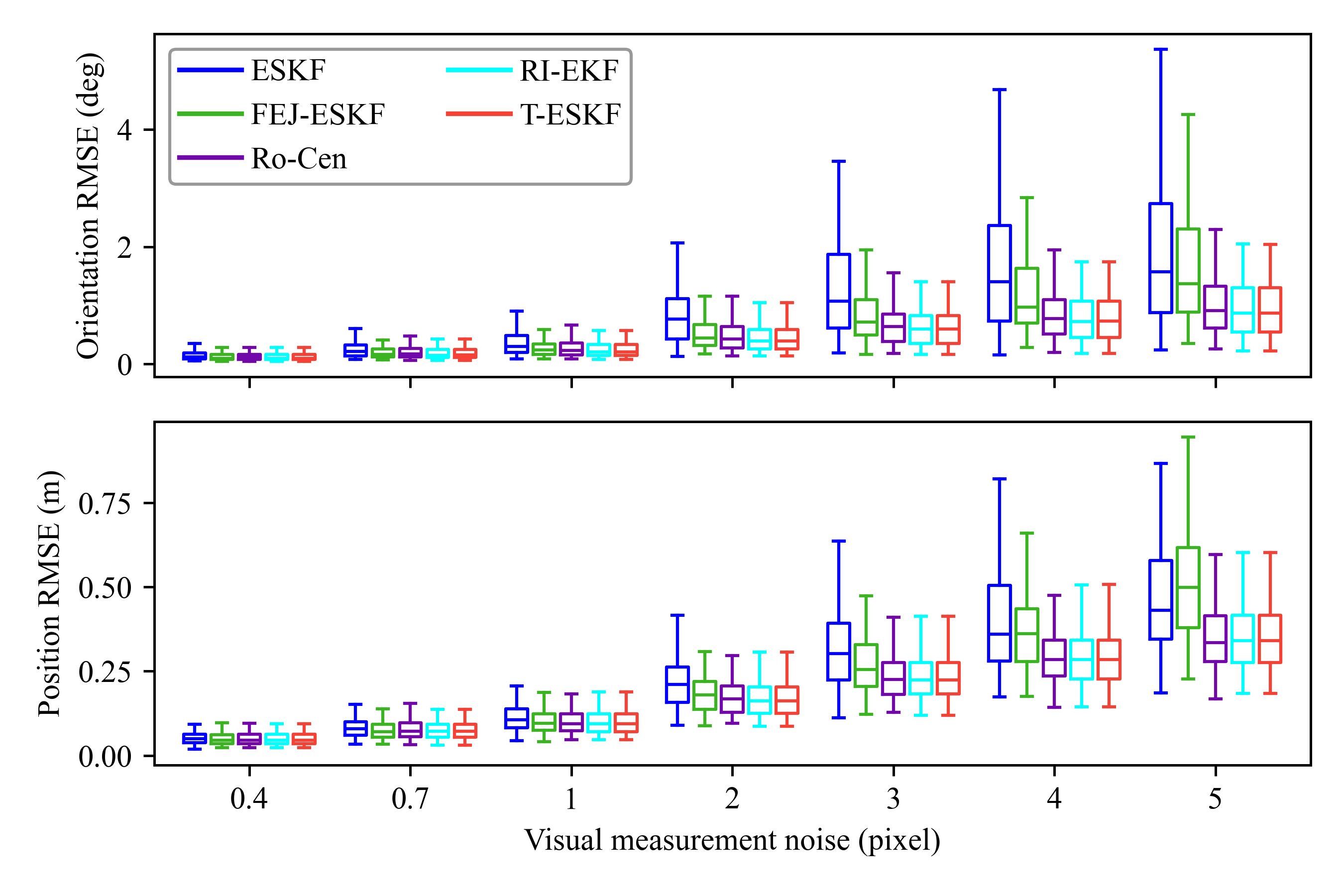}}
        \caption{\textrr{Orientation and position RMSE of 100 Monte-Carlo runs
                with various measurement noises on Udel-Gore.}}
        \label{fig:rmse}
    \vspace{-0.5cm}
\end{figure}

To compare the performance of the estimators under different measurement noise levels, visual measurement noise levels are set at [0.4, 0.7, 1, 2, 3, 4, 5] pixels. Each estimator undergoes 100 Monte Carlo simulation runs based on the \textit{Udel-Gore} trajectory. The RMSE results for these 100 simulation runs are depicted in Fig. \ref{fig:rmse}.
\textrr{
FEJ-ESKF uses the initial estimated landmark positions as linearization points to ensure consistency. However, as the noise level increases, the accuracy of landmark position initialization decreases significantly. As a result, the Jacobians of FEJ-ESKFs deviate from the optimal linearization point, leading to a significant decline in performance.
In contrast, T-ESKF evaluates the Jacobians using the current best estimates, thereby maintaining its Jacobian optimality. Consequently,  T-ESKF demonstrates better accuracy under varying measurement noise conditions.}

\subsection{Consistency evaluation}

\label{sec:cons}
To assess the consistency, 1000 Monte-Carlo runs are conducted using the {\it Udel-Gore} trajectory. The simulation parameters align with those presented in Table \ref{tab:sim_params}. The NEES results across the 1000 Monte-Carlo runs are visualized in Fig. \ref{fig:nees}. The upper three subfigures display the average NEES over time. Except for ESKF, the average NEES of the other estimators approaches 1. In particular, T-ESKF demonstrates NEES values closer to 1, indicating a higher consistency level than FEJ-ESKF. The bottom three subfigures illustrate the NEES frequency distribution across the 1000 runs throughout the simulation. \textrr{The NEES histograms of T-ESKF and RI-EKF closely match the theoretical chi-square distribution, further indicating their better consistency.}

\textrr{It is worth noting that T-ESKF and RI-EKF exhibit similar performance, as the error-state system utilized by T-ESKF is exactly identical to that of RI-EKF.
Compared to T-ESKF, Robo-Centric demonstrates a slightly conservative orientation estimation. We speculate that this may be due to its estimation of gravitational direction in the local frame, which couples gravity and orientation, thereby adding extra uncertainty to the orientation estimation.}

\begin{figure}[!htb]
        \centering
        {\includegraphics[width=0.95\linewidth]{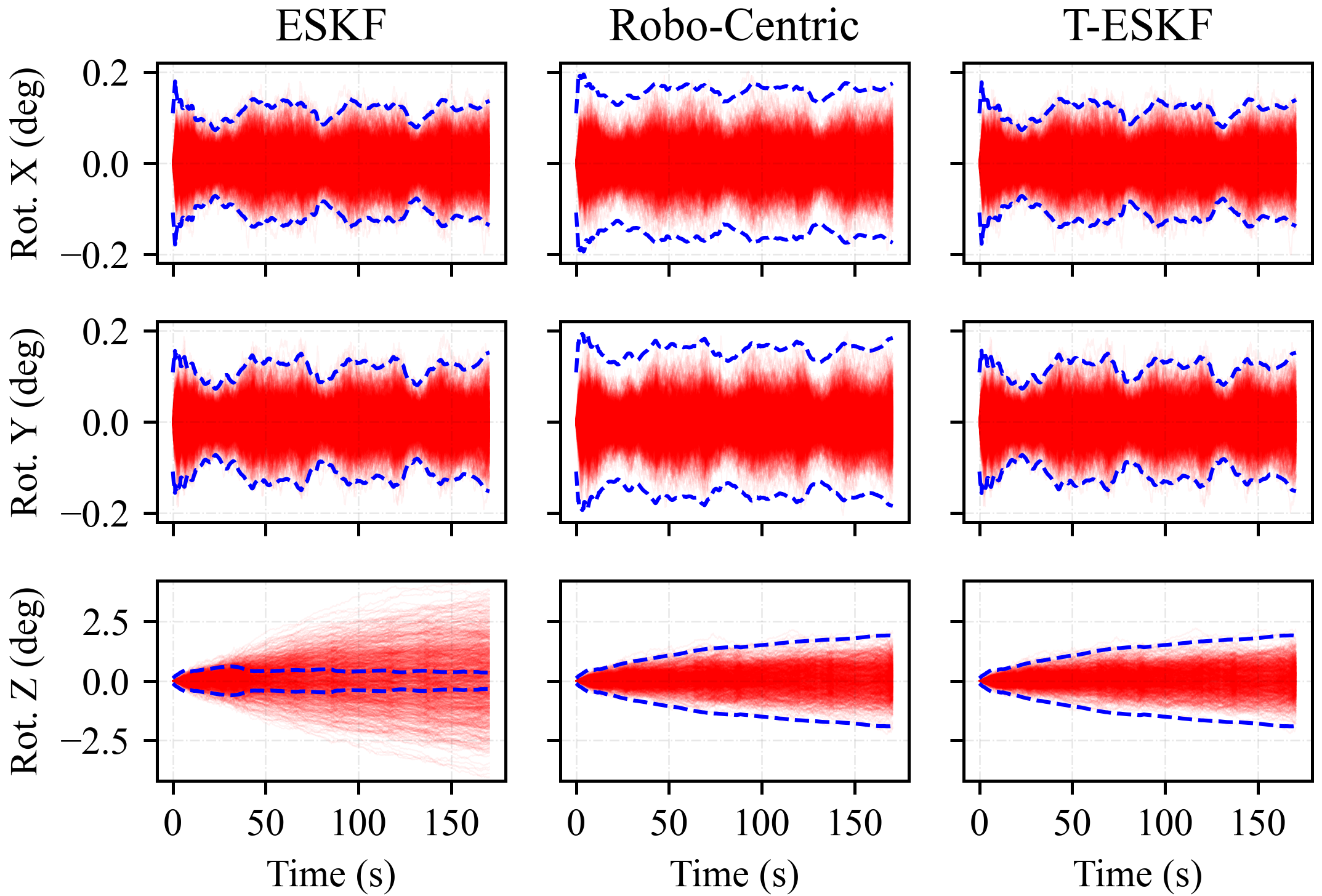}}
        \caption{\textrr{Orientation error over time for the 1000 Monte-Carlo runs. The blue dashed lines represent the $\pm 3\sigma$  bounds estimated by the estimators. The plots of RI-EKF and FEJ-ESKF are omitted since they are similar to T-ESKF's.}}
        \label{fig:rmse2}
\vspace{-0.2cm}
\end{figure}

\begin{figure}[!htb]
        \centering
        {\includegraphics[width=0.95\linewidth]{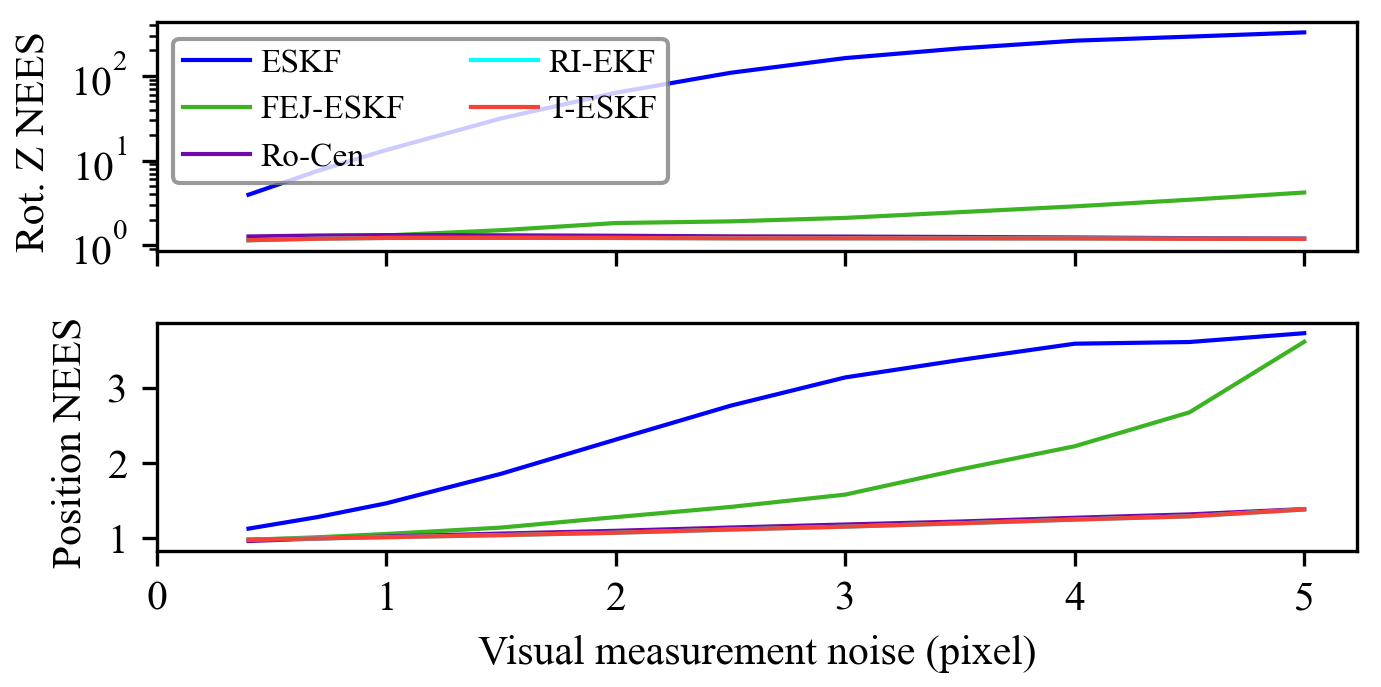}}
        \caption{\textrj{Average orientation and position NEES of 100 Monte-Carlo runs
        under various measurement noise conditions on Udel-Gore (the closer the NEES to 1, the better the consistency). The lines for Robo-Centric, RI-EKF, and T-ESKF coincide.}}
        \label{fig:nees2}
\end{figure}

Additionally, the orientation error \textrr{around each axis} is depicted in Fig. \ref{fig:rmse2}.
In ESKF, the orientation error around the $z$-axis exceeds the corresponding $\pm 3\sigma$ bounds, due to the orientation erroneously becoming observable. \textrr{Compared to T-ESKF, the orientation estimates of Robo-Centric around $x$ and $y$-axis are more conservative, which corresponds to the smaller NEES value in Fig. \ref{fig:nees}.} \textrj{Fig \ref{fig:nees2} reports the average NEES values of the four unobservable states under various measurement noises. T-ESKF exhibits a level of consistency comparable to that of the Robo-Centric and RI-EKF. Since Robo-Centric, RI-EKF, and T-ESKF methods utilize optimal Jocabians, their NEES does not deteriorate as significantly as that of FEJ-ESKF when the measurement noise increases.}

\vspace{-0.2cm}

\subsection{Computational efficiency evaluation}
\begin{figure}[!htbp]
        \vspace{-0.3cm}
                \centerline{\includegraphics[width=0.5\textwidth]{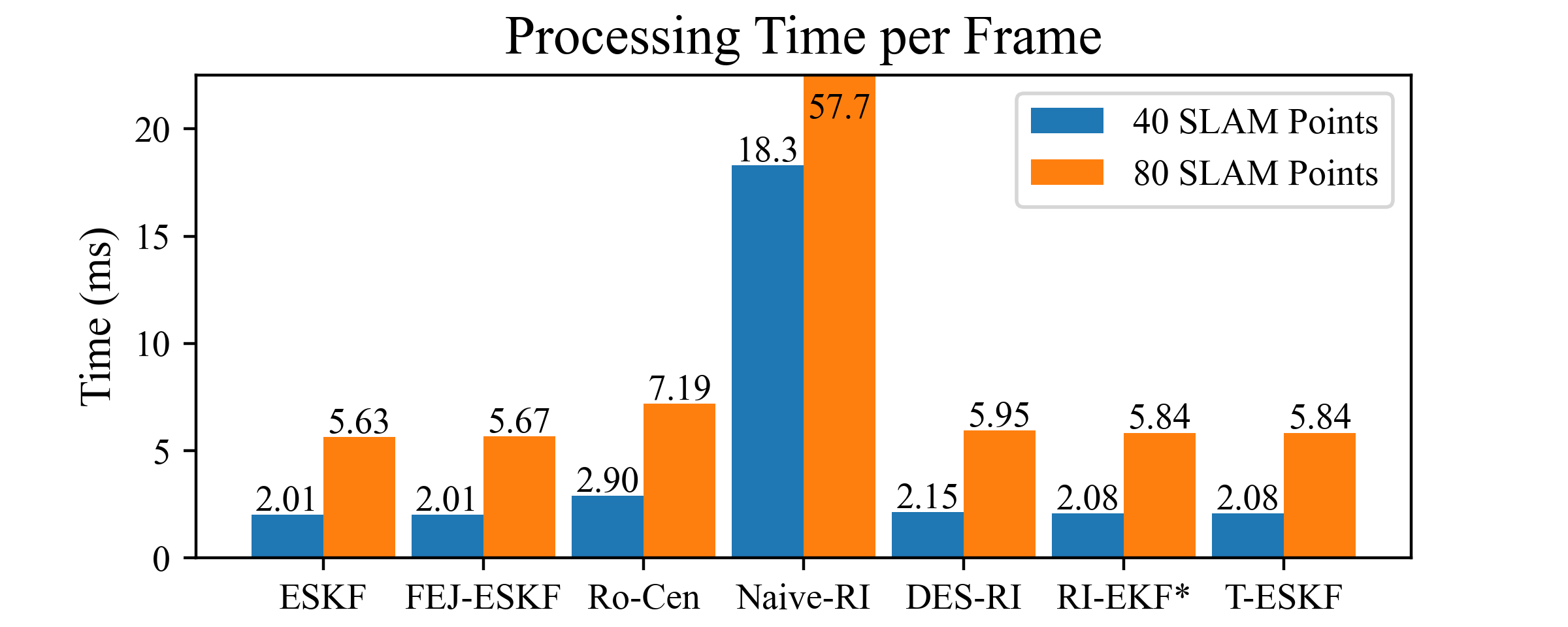}}
                \caption{\textac{Average time to process one frame tested on the Udel trajectory. The tests run on R9 7950X @ 4.5GHz. 
                RI-EKF* propagates the covariance utilizing equations \eqref{equ:38} and \eqref{equ:39} derived from T-ESKF.}}
                \label{fig:timing}
\end{figure}

Figure \ref{fig:timing} presents the average time required to process a single frame using these estimators.
One can see that T-ESKF achieves comparable computational efficiency as the ESKF while maintaining consistency. 
\textac{Additionally, we compare three different implementations of covariance propagation for RI-EKF: Naive-RI, which directly propagates covariance based on \eqref{equ:int_PHI} and \eqref{equ:intG}; DES-RI, which incorporates the DES concept from \cite{Chen2024}; and RI-EKF*, which employs the proposed efficient propagation technique.
The result shows that Naive-RI encounters substantial computational overhead in covariance propagation, while DES-RI and RI-EKF improve the computational efficiency significantly.} 

\vspace{-0.2cm}
\section{Real-World Experiments}
\label{sec:exp}

\begin{figure}[!b]
        \vspace{-0.6cm}
        \centerline{\includegraphics[width=0.475\textwidth]{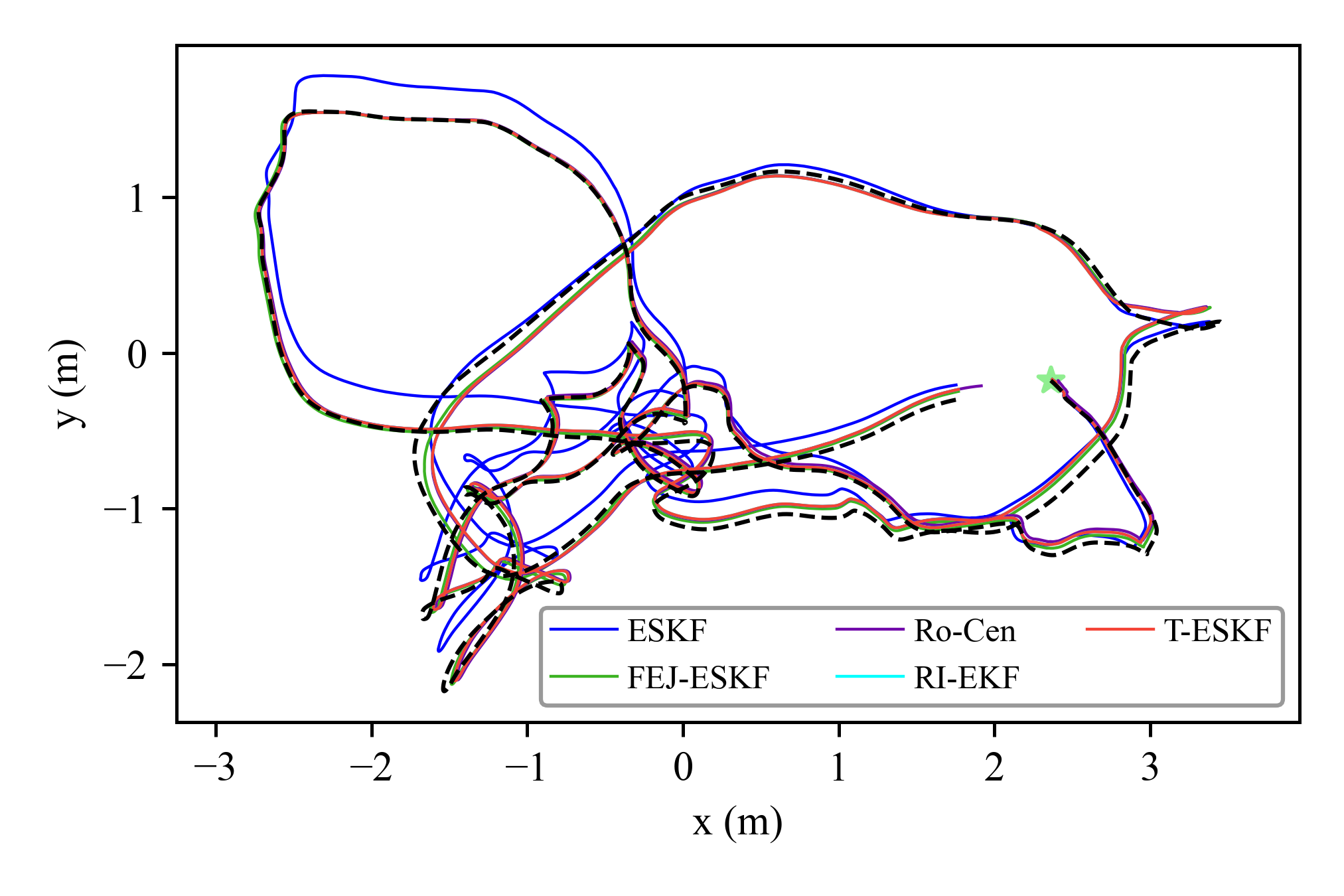}}
        \vspace{-0.3cm}
        \caption{
                \textrr{Estimated trajectories during the beginning 100 seconds of the EuRoC V1\_01 dataset. The black dashed line represents the ground truth trajectory, with a green star marking the starting point. The estimated trajectories are aligned to the ground truth trajectory (black dashed line) using the initial frame.}}
        \label{fig:v11}
        \vspace{-0.3cm}
\end{figure}

We further test these estimators using real-world datasets, including EuRoC and TUM-VI. Due to space limitations, additional experiments on our customized sensor platform are given in the supplemental material. 
For a fair comparison, all parameters are kept at their default values in OpenVINS.

The absolute trajectory error (ATE) is displayed in Table \ref{tab:ET}. Compared to ESKF, the other four estimators show a substantial enhancement in accuracy.
\begin{table}[!tb]
        \centering
        \setlength\tabcolsep{3.5pt}
        \caption{\textrr{Aabsolute trajectory error (ATE) on the EuRoC dataset and TUM-VI dataset}}
        \begin{tabular}{cccccc}
                \toprule
                Dataset  & ESKF                             & FEJ-ESKF                & \textrr{Ro-Cen}                    & RI-EKF                  & T-ESKF      \\
               \toprule
                V1\_01 & 0.75 / 0.06& 0.81 / 0.05& 1.21 / 0.06& 0.75 / 0.06& 0.75 / 0.06\\
                V1\_02 & 2.08 / 0.07& 1.78 / 0.06& 1.77 / 0.06& 1.70 / 0.05& 1.70 / 0.05\\
                V1\_03 & 2.48 / 0.06& 2.71 / 0.06& 2.76 / 0.07& 2.66 / 0.06& 2.65 / 0.06\\
                V2\_01 & 1.53 / 0.11& 1.04 / 0.12& 1.33 / 0.10& 1.06 / 0.09& 1.06 / 0.09\\
                V2\_02 & 1.76 / 0.08& 1.46 / 0.07& 1.80 / 0.08& 1.43 / 0.07& 1.43 / 0.07\\
                V2\_03 & 1.04 / 0.16& 1.39 / 0.16& 1.04 / 0.14& 1.11 / 0.15& 1.09 / 0.15\\
                \midrule
                Room1  & 1.31 / 0.07& 0.71 / 0.07& 0.96 / 0.06& 0.61 / 0.06& 0.61 / 0.06\\
                Room2  & 6.56 / 0.14& 0.90 / 0.09& 2.02 / 0.09& 1.00 / 0.09& 1.00 / 0.09\\
                Room3  & 6.91 / 0.14& 2.08 / 0.08& 2.45 / 0.09& 2.28 / 0.09& 2.28 / 0.09\\
                Room4  & 2.27 / 0.07& 1.05 / 0.05& 1.58 / 0.06& 1.00 / 0.06& 1.00 / 0.06\\
                Room5  & 2.00 / 0.11& 0.88 / 0.09& 1.15 / 0.10& 0.87 / 0.10& 0.88 / 0.10\\
                Room6  & 5.09 / 0.10& 1.86 / 0.09& 2.89 / 0.08& 2.40 / 0.07& 1.99 / 0.06\\
                \midrule
                Average               & 2.82 / 0.10 & 1.39 / \textbf{0.08} & 1.75 / \textbf{0.08} & 1.41 / \textbf{0.08} & \textbf{1.37} / \textbf{0.08} \\
                \bottomrule
        \end{tabular} \label{tab:ET}
\end{table}

\begin{figure}[!htbp]
        \vspace{-0.2cm}
                \centerline{\includegraphics[width=0.45\textwidth]{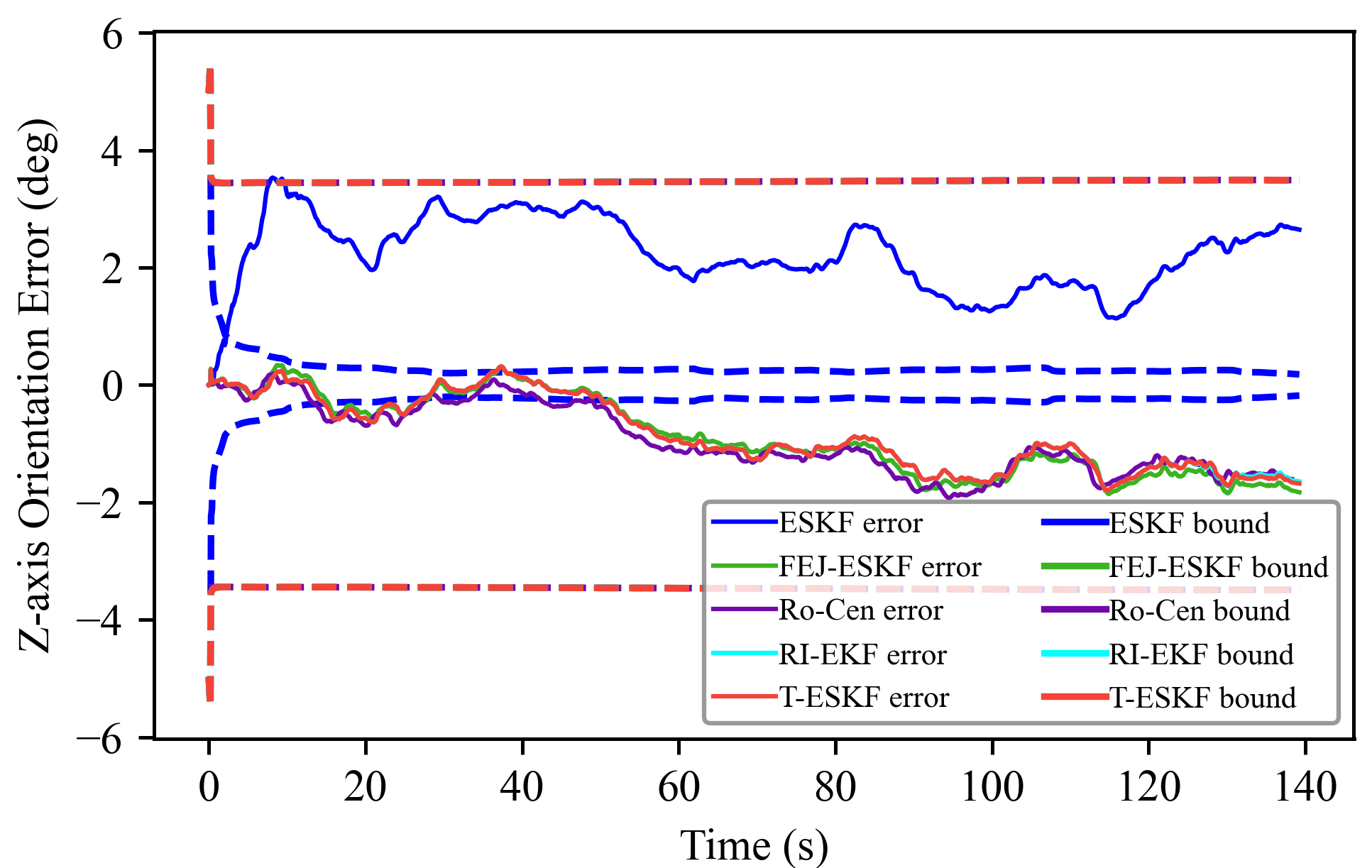}}
                \vspace{-0.2cm}
                \caption{\textrr{Orientation errors around the $z$-axis and the $\pm 3\sigma$ bounds estimated by the estimators on EuRoC V1\_01. The $\pm 3\sigma$  bounds of FEJ-ESKF, Robo-Centric, RI-EKF, and T-ESKF coincide and gradually increase.}}
                \label{fig:bound}
\end{figure}

To assess the consistency of the estimators, the estimated and ground truth trajectories are aligned based on the initial frame, as shown in Fig. \ref{fig:v11}.
    Fig. \ref{fig:bound} illustrates the orientation errors around the $z$-axis and the corresponding $\pm 3\sigma$ bounds on EuRoC V1\_01, providing a visual comparison of consistency among the estimators. The error for T-ESKF stays within its $\pm 3\sigma$ bound, showcasing its consistency. However, ESKF exhibits inconsistency as its estimation error promptly surpasses the $\pm 3\sigma$ bound. \textrj{Table \ref{tab:NE} presents the NEES values of the four unobservable states across various datasets. Real-world complexities, such as incorrect data associations, can cause the NEES value of the estimator to diverge from 1. The exceptionally high NEES value of ESKF's orientation demonstrates its overconfidence in orientation estimation.}

        \begin{table}[!h]
                \caption{ \textrj{NEES values of orientation around $z$-axis (dim = 1) and position (dim=3) on the EuRoC dataset and TUM-VI dataset}}
                \centering
                \setlength\tabcolsep{3pt}
                \begin{threeparttable}          %这行要添加
                        \begin{tabular}{cccccc}
                                \toprule
                                Dataset  & ESKF                                         & FEJ-ESKF                   &Ro-Cen                  & RI-EKF                        & T-ESKF                        \\
                                \toprule
                                {V1\_01}  & 680. /   2.90 &   1.05 /   0.60 &   1.09 /   0.61 &   0.96 /   0.93 &   0.96 /   0.94 \\
                                {V1\_02} & 264. /   1.93 &   1.78 /   2.27 &   0.53 /   1.01 &   0.61 /   1.63 &   0.60 /   1.63 \\
                                {V1\_03} & 482. /   5.05 &   1.51 /   3.96 &   1.43 /   2.98 &   2.08 /   6.67 &   2.02 /   6.69  \\
                                {V2\_01}  & 2909. /   5.98 &   0.35 /   2.44 &   0.20 /   1.59 &   0.09 /   1.52 &   0.09 /   1.52       \\
                                {V2\_02} & 323. /   1.49 &   0.79 /   1.33 &   0.41 /   1.13 &   0.75 /   1.47 &   0.69 /   1.47  \\
                                {V2\_03} &  20.1 /   7.71 &   2.57 /   6.22 &   0.24 /   3.53 &   1.12 /   6.04 &   1.09 /   5.94 \\
                                \midrule
                                Room1   & 163. /   1.86 &   1.12 /   1.54 &   0.26 /   0.86 &   0.17 /   1.02 &   0.17 /   0.99         \\
                                Room2    & 5147. /   8.25 &   0.57 /   3.08 &   1.29 /   5.76 &   1.27 /   6.00 &   1.24 /   5.63        \\
                                Room3   & 4187. /   5.17 &   3.01 /   1.70 &   4.66 /   1.72 &   4.05 /   2.58 &   4.04 /   2.58            \\
                                Room4   & 2468. /   3.14 &   2.79 /   1.09 &   5.03 /   1.94 &   6.17 /   1.42 &   6.38 /   1.38               \\
                                Room5    & 247. /   1.85 &   0.68 /   1.22 &   0.31 /   1.57 &   0.40 /   1.92 &   0.40 /   1.85   \\
                                Room6   & 3564. /   3.59 &   6.68 /   3.56 &   7.38 /   2.32 &   7.95 /   2.41 &   7.87 /   2.23         \\
                                \bottomrule
                        \end{tabular} \label{tab:NE}
                        % \begin{tablenotes}    %这行要添加， 从这开始
                        %         \footnotesize               %这行要添加
                        %         \item[1] The extremely overconfident (value$>$5) NEES  are in red.
                        % \end{tablenotes}
                \end{threeparttable}       %这行要添加，到这里结束
        \end{table}

\section{Conclusion}
\label{sec:concl}
In this paper, we propose a novel approach to address the inconsistency issue caused by observability mismatch in VINS. We design and apply a linear time-varying transformation to the linearized error-state system of ESKF, guranteeing that the \textrr{unobservable subspace of the transformed system} becomes state-independent. Through this transformation, the system's observability remains invariant to changes in linearization points, thereby ensuring consistency. Based on the transformed error-state system, we present a consistent VINS estimator referred to as T-ESKF. \textrj{Furthermore, we develop an efficient propagation technique to accelerate covariance propagation.} Simulations and experiments demonstrate that T-ESKF has competitive performance with state-of-the-art estimators.

\bibliographystyle{IEEEtran}
\bibliography{ref.bib}

\onecolumn

\begin{center}
    \fontsize{18pt}{24pt}\selectfont % 24pt 字号，行距 28pt（通常比字号大一些）
    Supplementary Material for ``T-ESKF: Transformed Error-State Kalman Filter for Consistent Visual-Inertial Navigation''
\end{center}

% The relation between this supplementary material and the manuscript is displayed in Figure \ref{fig_sup:1}. 
In this supplementary material, Section \ref{sec_sup:pre} provides fundamental knowledge of $\mathbf{SO}(3)$. Sections \ref{sec_sup:imu}-\ref{sec_sup:teskf} present the model and method encompassing the consideration of IMU bias. The real-world experiments on our customized platform are detailed in Section \ref{sec_sup:real}.  Appendix \ref{app:a} contains the derivation of \eqref{equ:Tpvl}, which is omitted from the manuscript.
Appendix \ref{app:a} provides the state update equations of T-ESKF.

% \begin{figure}[!h]
%     \centering
%     \includegraphics*[width=0.65\textwidth]{figures_sup/relationship.pdf}
%     \caption{Relation between this supplementary material and the manuscript.}
%     \label{fig_sup:1}
% \end{figure}

\section{Preliminaries: Special Orthogonal Group}
\label{sec_sup:pre}

\subsection{Lie algebra}
The Lie algebra corresponding to $\mathbf{SO}(3)$ is denoted by $\mathfrak{s}\mathfrak{o} (3)$. It consists of all $3 \times 3$ skew-symmetric matrices, i.e.,
\begin{equation}
    \mathfrak{s}  \mathfrak{o} (3) =\{\bftheta^{\land}| \bftheta \in \mathbb{R}^3 \},
\end{equation}
where a general element of $\mathfrak{s}\mathfrak{o} (3)$ can be written as $\bftheta^{\land} = \skew{\bftheta}$ (the operator 
$(\cdot)^{\land}$ converts a vector to its corresponding skew-symmetric matrix form) with
\begin{equation}
    \skew{\bftheta} = \skew{\begin{array}[]{c}
            \theta_1 \\
            \theta_2 \\
            \theta_3 \\
        \end{array}} = \begin{bmatrix}
        0         & -\theta_3 & \theta_2  \\
        \theta_3  & 0         & -\theta_1 \\
        -\theta_2 & \theta_1  & 0         \\
    \end{bmatrix}.      
\end{equation}
%For $\mathbf{SO}(3)$, 

\subsection{Exponential and logarithmic map}
Let $\bftheta$ represent a rotation vector, and $\theta \triangleq |\bftheta|$ be the rotaion angle and $\bfu \triangleq \frac{\bftheta}{|\bftheta|}$ the rotation axis.
The exponential map  $\text{exp}: \mathfrak{s}  \mathfrak{o} (3) \to \mathbf{SO}(3)$ transforms the space of $\mathfrak{s}  \mathfrak{o} (3)$ to the space of the rotations represented by rotation matrices
\begin{subequations}
    \begin{align}
        \text{exp}(\skew{\bftheta}) & = \bfI_3 + \frac{1}{1!}\skew{\bftheta} + \frac{1}{2!}\skew{\bftheta}^2 + \frac{1}{3!}\skew{\bftheta}^3 + \dots                                                                            \\
                                    & = \bfI_3 +  (  \frac{\theta}{1!} - \frac{\theta^3}{3!} +\frac{\theta^5}{5!} -\dots  )\skew{\bfu} + (\frac{\theta^2}{2!} -\frac{\theta^4}{4!} + \frac{\theta^6}{6!} -\dots  )\skew{\bfu}^2 \\
                                    & = \bfI_3 +  \text{sin}\theta \skew{\bfu} + (1- \text{cos}\theta )\skew{\bfu}^2                                                                                                            \\
                                    & = \bfI_3 \text{cos}\theta  + \skew{\bfu}\text{sin}\theta + \bfu \bfu^\top(1- \text{cos}\theta )                                                                                           \\
                                    & \triangleq \bfR.
    \end{align}
\end{subequations}
The exponential map $\text{Exp}: \mathbb{R}^3 \to \mathbf{SO}(3)$ is defined as
\begin{equation}
    \text{Exp}(\bftheta) =  \text{exp}(\skew{\bftheta}).
\end{equation}
The logarithmic map is defined as the inverse of the exponential map. Specifically, the map $\text{log}: \mathbf{SO}(3) \to  \mathfrak{s}  \mathfrak{o} (3)  $ is given by
\begin{align}
    \text{log}(\bfR)=\skew{\bfu \theta}
\end{align}
and the map $\text{Log}: \mathbf{SO}(3) \to  \mathbb{R}^3 $ is given by
\begin{align}
    \text{Log}(\bfR)=\bfu \theta.
\end{align}

For a given rotation matrix $\bfR \in \mathbf{SO}(3)$, the logarithmic map can be expressed as:
\begin{align}
    \theta & = \arccos\left( \frac{\text{trace}{(\bfR)}-1}{2} \right) \\
    \bfu   & = \frac{(\bfR-\bfR^\top)^{\lor}}{2 \sin{\theta}}
\end{align}
where the operator $(\cdot)^{\lor}$ is the inverse of the operator $(\cdot)^{\land}$.

\subsection{Adjoint operation}
For $\bfR\in  \mathbf{SO}(3)$, the adjoint operation $Ad_{\bfR}$ is defined as follows:
\begin{align}
Ad_{\bfR}(\boldsymbol{\phi})=\bfR \boldsymbol{\phi},
\end{align}
where $\boldsymbol{\phi} \in \mathbb{R}^3$ is a vector, and $Ad_{\bfR}=\bfR$ denotes the operation which applies the rotation matrix $\bfR$ to a vector.

For completeness, the adjoint operation can also be expressed in terms of the matrix representation:
\begin{align}
\skew{ Ad_{\bfR} \boldsymbol{\phi}}=\bfR \skew{\boldsymbol{\phi}} \bfR ^\top.\label{equ_sup:ad_so3}
\end{align}
%This means that the adjoint operation can rotate both the vector $\boldsymbol{\phi}}$ and its corresponding skew-symmetric matrix representation $\skew{\boldsymbol{\phi}}}$.
Then the exponential map satisfies the following relationship:
\begin{align}
    \text{Exp}( Ad_{\bfR}\boldsymbol{\phi}) =\bfR \text{Exp}({\boldsymbol{\phi}}) \bfR ^\top. 
\end{align}

%This means that applying the adjoint operation to the vector $\boldsymbol{\phi}$ before the exponential map is equivalent to conjugating the resulting rotation matrix by $\bfR$

%\begin{align}
%    \bfR \skew{\boldsymbol{\phi}} \bfR ^\top        & = \skew{ Ad_{\bfR} \boldsymbol{\phi}}     \label{equ_sup:ad_so3} \\
%    \bfR \text{Exp}({\boldsymbol{\phi}}) \bfR ^\top & = \text{Exp}( Ad_{\bfR}\boldsymbol{\phi})                    \\
%    Ad_{\bfR}                                       & =  \bfR
%\end{align}

\subsection{Jacobians}
Assuming that $\bfphi$ is small enough, the exponential map can be approximated by
\begin{align}
    \text{Exp}(\bftheta + \bfphi) & \approx  \text{Exp}(\bftheta) \text{Exp}( \bfJ_r(\bftheta)\bfphi) \\
    \text{Exp}(\bftheta + \bfphi) & \approx  \text{Exp}( \bfJ_l(\bftheta)\bfphi) \text{Exp}(\bftheta)
\end{align}
where \begin{equation}
    \bfJ_{r}(\bftheta) = \frac{\sin{\theta}}{\theta} \bfI_3 + \left( 1- \frac{\sin{\theta}}{\theta} \right) \bfu \bfu^\top - \frac{1- \cos \theta}{\theta} \skew{\bfu}
\end{equation} and $\bfJ_{l}(\bftheta) = \bfJ_{r}(-\bftheta) $. $\bfJ_{r}$ and $\bfJ_{l}$ are called right and left Jacobians of $\mathbf{SO}(3)$, respectively.

\section{IMU model and error-state kinematics}
\label{sec_sup:imu}
This section presents the IMU model and its error-state kinematics. The main result is equivalent to what is described in {\it The ESKF using global angular errors} \cite[Chapter 7]{eskf}. If you are familiar with ESKF, you can skip this section.

\subsection{IMU model}
The IMU state vector is defined as
\begin{align}
    \bfx_I & = (\bfR,\bfp,\bfv,\bfb_g,\bfb_a),
\end{align}
where $\bfR \in \mathbf{SO}(3)$, $\bfp \in \mathbb{R}^3$ and $\bfv \in \mathbb{R}^3$ are the orientation, position, and velocity of IMU in the global frame, $\bfb_g \in \mathbb{R}^3$ and $\bfb_a \in \mathbb{R}^3 $ are the gyroscope bias and accelerometer bias, respectively.

The continuous motion model for the IMU state vector is given by the following differential equations:
\begin{subequations}
    \begin{align}
        \dot{\bfR}     & = \bfR\skew{\bfomega} \\
        \dot{\bfp}     & = \bfv                \\
        \dot{\bfv}     & = \bfa                \\
        \dot{\bfb}_{g} & = \bfn_{gw}           \\
        \dot{\bfb}_{a} & = \bfn_{aw}
    \end{align}
    \label{equ_sup:f}
\end{subequations}
where
\begin{align}
     \bfomega & = \bfomega_m-\bfb_g- \bfn_{g}  \\
    \bfa & =\bfR (\bfa_m - \bfb_{a}- \bfn_{a}) + \bfg
\end{align}
$\bfomega_m \in \R3 $ and $\bfa_m \in \R3 $ are the gyroscope and accelerometer measurements in the IMU frame, 
$\bfn_{gw} \sim \mathcal{N}(0,\sigma_{gw}^2 \bfI_3)$,  $\bfn_{aw} \sim \mathcal{N}(0,\sigma_{aw}^2 \bfI_3)$,
$\bfn_g \sim \mathcal{N}(0,\sigma_{g}^2 \bfI_3)$, and $\bfn_a \sim \mathcal{N}(0,\sigma_{a}^2 \bfI_3)$ are assumed to be white Gaussian noises.

\subsection{IMU state estimate and error-state}
Let $\hat{\bfx}_I$ denote the IMU state estimate,
which is propagated following \eqref{equ_sup:f} by setting the process noise to zero:
\begin{subequations}
    \begin{align}
        \dot{\hat{\bfR}}   & = \hat\bfR\skew{\hat{\bfomega}} \\
        \dot{\hat\bfp}     & = \hat\bfv                      \\
        \dot{\hat\bfv}     & = \hat\bfa                      \\
        \dot{\hat\bfb}_{g} & = \bf0                          \\
        \dot{\hat\bfb}_{a} & = \bf0
    \end{align}
\end{subequations}
with
\begin{align}
    \hat \bfomega & = \bfomega_m - \hat{\bfb}_g                \\
    \hat \bfa     & = \hat{\bfR}(\bfa_m - \hat{\bfb}_a) + \bfg.
\end{align}

In the classical VINS estimator, the error-state Kalman filter (ESKF), the error-state representation is used to handle uncertainties in the state estimates:
\begin{equation}
    \tilde{\bfx}_I = {\bfx}_I \ominus \hat{{\bfx}}_I
    \label{equ_sup:error_x}
\end{equation} where the error-state of orientation is defined using the logarithm of $\mathbf{SO}(3)$ while the errors of other variables are defined on vector space, the general minus $\ominus$ is defined as
\begin{subequations}
    \begin{align}
        \tilde{\bftheta} & = \text{Log}(\bfR \hat \bfR^{-1})    \label{equ_sup:dR} \\
        \tilde\bfp       & = \bfp -   \hat{\bfp}           \label{equ_sup:dp}      \\
        \tilde\bfv       & = \bfv -   \hat{\bfv}                               \\
        \tilde\bfb_g     & =  \bfb_g - \hat{\bfb}_g                            \\
        \tilde\bfb_a     & =  \bfb_a - \hat{\bfb}_a.
    \end{align}
\end{subequations}
This error-state formulation allows for the use of standard Kalman filter techniques to estimate the uncertainties and correct the state estimates.

\subsection{IMU error-state kinematics}

The IMU error-state kinematics is 
\begin{subequations}
    \begin{align}
        \dot{\tilde{\bftheta}} & = -\hat{\bfR} \tilde{\bfb}_g -\hat{\bfR} \bfn_g                                                                  \\
        \dot{\tilde{\bfp}}     & = \tilde{\bfv}                                                                                           \\
        {\dot{\tilde{\bfv}}}   & =   -\skew{\hat\bfR (\bfa_m - \hat{\bfb}_a)} \tilde{\bftheta} -\hat{\bfR}\tilde\bfb_a - \hat{\bfR}\bfn_a \\
        \dot{\tilde{\bfb}}_g   & =  \bfn_{gw}                                                                                             \\
        \dot{\tilde{\bfb}}_a   & =  \bfn_{aw}.
    \end{align}
\end{subequations}
The derivation of the equations will be given in Section \ref{sec_sup:ori}-\ref{sec_sup:bias}.
The above equations can be written in matrix form:
\begin{equation}
    \dot{\tilde{\bfx}}_I = \bfF_I\tilde{\bfx}_I + \bfG_I \bfn \label{sys_sup:imu}
\end{equation}
where
\begin{align}
    \bfF_I & = \begin{bmatrix}
        \bf0                                      & \bf0 & \bf0   & -\hat{\bfR} & \bf0        \\
        \bf0                                      & \bf0 & \bfI_3 & \bf0        & \bf0        \\
        -\skew{\hat{\bfR}(\bfa_m - \hat{\bfb}_a)} & \bf0 & \bf0   & \bf0        & -\hat{\bfR} \\
        \bf0                                      & \bf0 & \bf0   & \bf0        & \bf0        \\
        \bf0                                      & \bf0 & \bf0   & \bf0        & \bf0        \\
    \end{bmatrix},  \\
    \bfG_I & = \begin{bmatrix}
        -\hat{\bfR} & \bf0        & \bf0 & \bf0 \\
        \bf0        & \bf0        & \bf0 & \bf0 \\
        \bf0        & -\hat{\bfR} & \bf0 & \bf0 \\
        \bf0        & \bf0        & \bfI & \bf0 \\
        \bf0        & \bf0        & \bf0 & \bfI \\
    \end{bmatrix},  \\
    \bfn   & = \begin{bmatrix}
        \bfn_g    \\
        \bfn_a    \\
        \bfn_{gw} \\
        \bfn_{aw} \\
    \end{bmatrix},
\end{align}
with
\begin{equation}
    \bfE(\bfn \bfn^\top) = \begin{bmatrix}
        \sigma_{g}^2\bfI_3 &\bf0 &\bf0 &\bf0\\ 
        \bf0 &\sigma_{a}^2\bfI_3&\bf0 &\bf0\\ 
        \bf0 &\bf0 &\sigma_{gw}^2\bfI_3&\bf0\\ 
        \bf0 &\bf0 &\bf0 &\sigma_{aw}^2\bfI_3\\ 
    \end{bmatrix}\triangleq \bfQ .
\end{equation}

\subsubsection{Orientation}
\label{sec_sup:ori}

We proceed by computing $\dot{\bfR}$ by two different means (left and right developments)
%\begin{eqnarray}
%    \dot{\bfR}  &=&\bfR \skew{\bfomega}    \nonumber \\
%&=& \overset{.}{\left(\text{Exp}(\tilde\bftheta)\hat{\bfR}   \right) } \nonumber \\
%&=& \overset{.}{\text{Exp}(\tilde\bftheta)}\hat{\bfR}  + {\text{Exp}(\tilde\bftheta)\dot{\hat{\bfR}}} \nonumber \\
%&=& \text{Exp}(\tilde\bftheta) \skew{\dot{\tilde{\bftheta}}} \hat{\bfR} + \text{Exp}(\tilde\bftheta) \hat{\bfR} \skew{\hat\bfomega}        \nonumber \\
%&=&\text{Exp}(\tilde\bftheta)\hat{\bfR}\skew{\bfomega}
%\end{eqnarray}

\begin{equation}
    \begin{array}[pos]{rcl}
        \overset{.}{\left(\text{Exp}(\tilde\bftheta)\hat{\bfR}   \right) }    =                                                                                                                          & \dot{\bfR} & =   \bfR \skew{\bfomega}                                            \\
        \overset{.}{\text{Exp}(\tilde\bftheta)}\hat{\bfR}  + {\text{Exp}(\tilde\bftheta)\dot{\hat{\bfR}}}  =                                                                                             &            & =  \left(\text{Exp}(\tilde\bftheta)\hat{\bfR}\right)\skew{\bfomega} \\
        \text{Exp}(\tilde\bftheta) \skew{\dot{\tilde{\bftheta}}} \hat{\bfR} + \text{Exp}(\tilde\bftheta) \hat{\bfR} \skew{\hat\bfomega}                                                                = &            & =  \text{Exp}(\tilde\bftheta)\hat{\bfR}\skew{\bfomega}
    \end{array}
\end{equation}
Having $\bfomega - \hat{\bfomega} = -\tilde\bfb_g -\bfn_g $, this reduces to
\begin{align}
    \skew{\dot{\tilde{\bftheta}}} \hat{\bfR} & = \hat{\bfR} \skew{-\tilde\bfb_g -\bfn_g}
\end{align}
Right-multiplying left and right terms by $\hat{\bfR}^{-1}$, and recalling the adjoint operation on $\mathbf{SO}(3)$, i.e., \eqref{equ_sup:ad_so3}, we have
\begin{align}
    \skew{\dot{\tilde{\bftheta}}} & = \hat{\bfR} \skew{-\tilde\bfb_g -\bfn_g}\hat{\bfR}^{-1} \\
                                  & = \skew{-\hat{\bfR} \tilde\bfb_g -\hat{\bfR} \bfn_g}
\end{align}
Finally, the error-state kinematic equation for orientation is given by
\begin{equation}
    \dot{\tilde{\bftheta}}= -\hat{\bfR}\tilde \bfb_g -\hat{\bfR} \bfn_g
\end{equation}

\subsubsection{Position}
\begin{equation}
    \begin{array}[pos]{rcl}
        \overset{.}{\left(\hat{\bfp}+ \tilde{\bfp}  \right) }    = & \dot{\bfp} & =   \bfv                     \\
        \hat\bfv  + {\dot{\tilde{\bfp}}}  =                        &            & =   \hat{\bfv}+\tilde{\bfv}.
    \end{array}
\end{equation}
The position error-state  kinematics is :
\begin{equation}
    \dot{\tilde{\bfp}} = \tilde{\bfv}
\end{equation}

\subsubsection{Velocity}
\begin{equation}
    \begin{array}[pos]{rcl}
        \overset{.}{\left(\hat{\bfv}+ \tilde{\bfv}  \right) }    =          & \dot{\bfv} & =   \bfa                                                                    \\
        \hat\bfa  + {\dot{\tilde{\bfv}}}  =                                 &            & =  {\bfR}(\bfa_m - {\bfb}_a - \bfn_a) + \bfg                                \\
        \hat{\bfR}(\bfa_m - \hat{\bfb}_a) + \bfg  + {\dot{\tilde{\bfv}}}  = &            & = \text{Exp}(\tilde{\bftheta}) \hat {\bfR}(\bfa_m - {\bfb}_a - \bfn_a)+\bfg
    \end{array}
\end{equation}
Simplifying this equation yields the velocity error-state  kinematics:
\begin{subequations}
    \begin{align}
        {\dot{\tilde{\bfv}}} & = \text{Exp}(\tilde{\bftheta}) \hat {\bfR}(\bfa_m - {\bfb}_a - \bfn_a) - \hat{\bfR}(\bfa_m - \hat{\bfb}_a)                                                 \\
                             & \simeq \left(\bfI_3+\skew{\tilde{\bftheta}}\right)\hat {\bfR}(\bfa_m - {\bfb}_a - \bfn_a) - \hat{\bfR}(\bfa_m - \hat{\bfb}_a) \\
                             & = -\hat{\bfR}\tilde\bfb_a - \hat{\bfR}\bfn_a +\skew{\tilde{\bftheta}} \hat\bfR (\bfa_m-\bfb_a - \bfn_a)                                                    \\
                             & \simeq -\skew{\hat\bfR (\bfa_m - \hat{\bfb}_a)} \tilde{\bftheta} -\hat{\bfR}\tilde\bfb_a - \hat{\bfR}\bfn_a
    \end{align}
\end{subequations}

\subsubsection{Bias}
\label{sec_sup:bias}
\begin{align}
    \dot{\tilde{\bfb}}_g & = \dot{\bfb}_g -  \dot{\hat{\bfb}}_g = \bfn_{gw} - \bf0 = \bfn_{gw} \\
    \dot{\tilde{\bfb}}_a & = \dot{\bfb}_a -  \dot{\hat{\bfb}}_a = \bfn_{aw} - \bf0 = \bfn_{aw} 
\end{align}

\section{Visual-Inertial Navigation System}
\subsection{System model}
The system state is defined as

\begin{equation}
    \bfx = \left( \bfx_I,\bfl \right)
    % \bfx = \left( \bfx_I,\bfl \right) = (\bfR,\bfp,\bfv,\bfb_g,\bfb_a,\bfl) 
\end{equation}
where $\bfl\in \mathbb{R}^3$ is the landmark position in the global frame.

The system kinematic equations are
\begin{subequations}
    \begin{align}
        \dot{\bfR}     & = \bfR\skew{\bfomega} \\
        \dot{\bfp}     & = \bfv                \\
        \dot{\bfv}     & = \bfa                \\
        \dot{\bfb}_{g} & = \bfn_{gw}           \\
        \dot{\bfb}_{a} & = \bfn_{aw}           \\
        \dot{\bfl}     & = \bf0
    \end{align}
    \label{equ_sup:VINS_f}
\end{subequations}

Let ${^I}\bfp_L$ denote the landmark position in the IMU frame, expressed as
\begin{equation}
        {^I}\bfp_L = \bfR^\top (\bfl - \bfp).
\end{equation}
As the camera explores the environment, the visual measurement of the landmark is available after the data association and rectification, formulated as
\begin{align}
        \bfy =\bfh({^I}\bfp_{L}) + \bfepsilon
        \label{sys_sup:h}%   
\end{align}
where $\bfh = \pi \circ \varUpsilon $,
$ \varUpsilon:  \mathbb{R}^3 \to \mathbb{R}^3$ transforms points from the IMU frame to the camera frame, and
$\pi: \mathbb{R}^3 \to \mathbb{R}^2$ is the camera perspective projection function,
$\bfepsilon \in \mathbb{R}^2$ is the zero-mean Gaussian noise with $\bfE(\bfepsilon \bfepsilon^\top) = \bfV$.

\subsection{Linearized error-state system}
Let $\hat{\bfx}$ denote the VINS state estimate,
which is propagated following \eqref{equ_sup:VINS_f} by setting the process noise to zero:
\begin{subequations}
    \begin{align}
        \dot{\hat{\bfR}}   & = \hat\bfR\skew{\hat{\bfomega}} \\
        \dot{\hat\bfp}     & = \hat\bfv                      \\
        \dot{\hat\bfv}     & = \hat\bfa                      \\
        \dot{\hat\bfb}_{g} & = \bf0                          \\
        \dot{\hat\bfb}_{a} & = \bf0                          \\
        \dot{\hat\bfl}     & = \bf0
    \end{align}
    \label{equ_sup:dyn_nomi}
\end{subequations}

The error-state of VINS is
\begin{equation}
    \tilde\bfx = \bfx \ominus \hat{\bfx} =  \begin{bmatrix}
        \bfx_I \ominus \hat{\bfx}_I \\
        \bfl- \hat{\bfl}
    \end{bmatrix}\triangleq\begin{bmatrix}
        \tilde{\bfx}_I \\
        \tilde{\bfl}
    \end{bmatrix}
    %  = \begin{bmatrix}
    %     \tilde{\bftheta} \\
    %     \tilde{\bfp} \\
    %     \tilde{\bfv} \\
    %     \tilde{\bfb}_g \\
    %     \tilde{\bfb}_a \\
    %     \tilde{\bfl} \\
    % \end{bmatrix}
\end{equation}
According to \eqref{sys_sup:imu}, the VINS error-state kinematics can be written as
\begin{equation}
    \dot{\tilde{\bfx}} = \bfF\tilde{\bfx} + \bfG \bfn \label{sys_sup:dyn}
\end{equation}
where
\begin{align}
    \bfF & = \begin{bmatrix}
        \bfF_I            & \bf0_{15\times 3} \\
        \bf0_{3\times 15} & \bf0_{3\times 3}  \\
    \end{bmatrix} = \begin{bmatrix}
        \bf0                                      & \bf0 & \bf0   & -\hat{\bfR} & \bf0        & \bf0 \\
        \bf0                                      & \bf0 & \bfI_3 & \bf0        & \bf0        & \bf0 \\
        -\skew{\hat{\bfR}(\bfa_m - \hat{\bfb}_a)} & \bf0 & \bf0   & \bf0        & -\hat{\bfR} & \bf0 \\
        \bf0                                      & \bf0 & \bf0   & \bf0        & \bf0        & \bf0 \\
        \bf0                                      & \bf0 & \bf0   & \bf0        & \bf0        & \bf0 \\
        \bf0                                      & \bf0 & \bf0   & \bf0        & \bf0        & \bf0 \\
    \end{bmatrix}
    \\
    \bfG & = \begin{bmatrix}
        \bfG_I \\
        \bf0_{3\times 12}
    \end{bmatrix} =\begin{bmatrix}
        -\hat{\bfR} & \bf0        & \bf0 & \bf0 \\
        \bf0        & \bf0        & \bf0 & \bf0 \\
        \bf0        & -\hat{\bfR} & \bf0 & \bf0 \\
        \bf0        & \bf0        & \bfI & \bf0 \\
        \bf0        & \bf0        & \bf0 & \bfI \\
        \bf0        & \bf0        & \bf0 & \bf0 \\
    \end{bmatrix}
\end{align}
The measurement residual is :
\begin{subequations}
    \begin{align}
        \tilde{\bfy} & =  \bfy-  \bfh({^I}\hat\bfp_L)                                                                                                                                               \\
                     & = \bfh({^I}\bfp_L) - \bfh({^I}\hat\bfp_L) + \bfepsilon                                                                                                                       \\
                     & = \frac{\partial \bfh}{\partial {^I}\bfp_L} \text{d}  {^I}\bfp_L + \bfepsilon                                                                                                \\
                     & = \frac{\partial \bfh}{\partial {^I}\bfp_L} \left(\bfR^\top (\bfl - \bfp) - \hat\bfR^\top (\hat\bfl - \hat\bfp)\right) + \bfepsilon                                          \\
                     & \simeq \frac{\partial \bfh}{\partial {^I}\bfp_L} \left(\hat\bfR^\top(\bfI_3-\skew{\tilde{\bftheta}}) (\bfl - \bfp) - \hat\bfR^\top (\hat\bfl - \hat\bfp)\right) + \bfepsilon \\
                     & = \frac{\partial \bfh}{\partial {^I}\bfp_L} \hat{\bfR}^\top \left(\skew{ \bfl-\bfp} \tilde{\bftheta}+ \tilde\bfl-\tilde\bfp  \right) + \bfepsilon                                   \\
                     & \simeq \frac{\partial \bfh}{\partial {^I}\bfp_L}\hat\bfR^\top \left(\skew{\hat \bfl-\hat\bfp} \tilde{\bftheta}+ \tilde\bfl-\tilde\bfp  \right) + \bfepsilon \\
                     & =  \bfH \tilde{\bfx} + \bfepsilon
    \end{align}
    \label{sys_sup:mea}
\end{subequations}
where
\begin{align}
    \bfH   & = \bfPi \bfH_{{e}}                                  \label{equ_sup:H} \\
    \bfPi  & =  \frac{\partial \bfh}{\partial {^I}\bfp_L}\bfR^\top         \\
    \bfH_{{e}} & = \begin{bmatrix}
        \skew{\hat \bfl-\hat\bfp} & -\bfI_3 & \bf0 & \bf0 & \bf0 & \bfI_3
    \end{bmatrix}.
\end{align}
We call $\bfH_{{e}}$  the {\it essential measurement Jacobian}.

By combining \eqref{sys_sup:dyn} and \eqref{sys_sup:mea}, we obtain the linearized error-state system of VINS:
\begin{equation}
    \left\{
    \begin{array}{rl}
        \dot{\tilde{\bfx}} & = \bfF\tilde{\bfx} + \bfG \bfn   \\
        \tilde{\bfy}       & = \bfH \tilde{\bfx} + \bfepsilon
    \end{array}
    \right.
    \label{sys_sup:original}
\end{equation}

\section{Transformed Linearized Error-State System}

\subsection{Linear time-varying transformation on error-state}

Denote the linear time-varying transformation by $\bfT(\hat{\bfx})$, where $\bfT(\cdot)$ is a $18 \times 18$ nonsingular matrix function and remains to be chosen.
The transformed error-state is obtained by  multiplying the transformation with the original error-state:
\begin{equation}
    \tilde{ \bfx}^* = \bfT(\hat{\bfx}) \tilde{ \bfx}.
    \label{equ_sup:trans_x}
\end{equation}
For simplicity,  the variable $\hat{\bfx}$ of $\bfT(\hat{\bfx})$ is omitted in the following if there is no ambiguity.
Taking the derivative of both sides of \eqref{equ_sup:trans_x} with respect to time, we have
\begin{equation}
    \dot{\tilde{ \bfx}}^* = \dot{\bfT}\tilde{ \bfx} + {\bfT}\dot{\tilde{{ \bfx}}}.
    \label{equ_sup:trans_der}
\end{equation}
% where $\dot{\bfT}$ is the derivative of $\bfT$ with respect to time.
Substituting  \eqref{equ_sup:trans_x} and \eqref{equ_sup:trans_der} into  the linearized error-state system \eqref{sys_sup:original} yields the transformed linearized error-state system as follows
\begin{equation}
    \left\{
    \begin{aligned}
        \dot{  \tilde{ \bfx}}^* & = {\bfF^*}\tilde{ \bfx}^* +{\bfG^*} \bfn \\
        \tilde{\bfy}            & = {\bfH^*} \tilde{ \bfx}^*+ \bfepsilon
    \end{aligned}
    \right.       \label{sys_sup:trans}%
\end{equation}
where
\begin{align}
    \bfF^* & = \dot{\bfT} \bfT^{-1}+\bfT \bfF  \bfT^{-1}  \label{equ_sup:kdF} \\
    \bfG^* & = \bfT \bfG                                      \label{equ_sup:kdG}            \\
    \bfH^* & = \bfH \bfT^{-1}.  \label{equ_sup:kdH}
\end{align}
According to \eqref{equ_sup:H} and \eqref{equ_sup:kdH}, the essential measurement Jacobian $\bfH_{{e}} $ is transformed into
\begin{equation}
    \bfH_{{e}} ^* =\bfH_{{e}} \bfT^{-1}. \label{equ_sup:kdHv}
\end{equation}
The transformation is designed as follows:
\begin{equation}
    \bfT = \begin{bmatrix}
        \bfI_3            & \bf0   & \bf0   & \bf0   & \bf0   & \bf0   \\
        \skew{\hat{\bfp}} & \bfI_3 & \bf0   & \bf0   & \bf0   & \bf0   \\
        \skew{\hat{\bfv}} & \bf0   & \bfI_3 & \bf0   & \bf0   & \bf0   \\
        \bf0              & \bf0   & \bf0   & \bfI_3 & \bf0   & \bf0   \\
        \bf0              & \bf0   & \bf0   & \bf0   & \bfI_3 & \bf0   \\
        \skew{\hat{\bfl}} & \bf0   & \bf0   & \bf0   & \bf0   & \bfI_3 \\
    \end{bmatrix}
\end{equation}
Correspondingly, its inverse and derivative are
\begin{equation}
    \bfT^{-1} = \begin{bmatrix}
        \bfI_3             & \bf0   & \bf0   & \bf0   & \bf0   & \bf0   \\
        -\skew{\hat{\bfp}} & \bfI_3 & \bf0   & \bf0   & \bf0   & \bf0   \\
        -\skew{\hat{\bfv}} & \bf0   & \bfI_3 & \bf0   & \bf0   & \bf0   \\
        \bf0               & \bf0   & \bf0   & \bfI_3 & \bf0   & \bf0   \\
        \bf0               & \bf0   & \bf0   & \bf0   & \bfI_3 & \bf0   \\
        -\skew{\hat{\bfl}} & \bf0   & \bf0   & \bf0   & \bf0   & \bfI_3 \\
    \end{bmatrix}
\end{equation}
\begin{equation}
    \dot\bfT = \begin{bmatrix}
        \bf0              & \bf0 & \bf0 & \bf0 & \bf0 & \bf0 \\
        \skew{\hat{\bfv}} & \bf0 & \bf0 & \bf0 & \bf0 & \bf0 \\
        \skew{\hat{\bfa}} & \bf0 & \bf0 & \bf0 & \bf0 & \bf0 \\
        \bf0              & \bf0 & \bf0 & \bf0 & \bf0 & \bf0 \\
        \bf0              & \bf0 & \bf0 & \bf0 & \bf0 & \bf0 \\
        \bf0              & \bf0 & \bf0 & \bf0 & \bf0 & \bf0 \\
    \end{bmatrix}.
\end{equation}
The transformed Jacobians are
\begin{align}
    \bfF^*   & =\begin{bmatrix}
        \bf0        & \bf0 & \bf0   & -\hat{\bfR}                 & \bf0        & \bf0 \\
        \bf0        & \bf0 & \bfI_3 & -\skew{\hat\bfp} \hat{\bfR} & \bf0        & \bf0 \\
        \skew{\bfg} & \bf0 & \bf0   & -\skew{\hat\bfv}\hat{\bfR}  & -\hat{\bfR} & \bf0 \\
        \bf0        & \bf0 & \bf0   & \bf0                        & \bf0        & \bf0 \\
        \bf0        & \bf0 & \bf0   & \bf0                        & \bf0        & \bf0 \\
        \bf0        & \bf0 & \bf0   & -\skew{\hat\bfl} \bfR       & \bf0        & \bf0 \\
    \end{bmatrix}  \\
    \bfG^*   & =\begin{bmatrix}
        -\hat{\bfR}                 & \bf0        & \bf0 & \bf0 \\
        -\skew{\hat\bfp} \hat{\bfR} & \bf0        & \bf0 & \bf0 \\
        -\skew{\hat\bfv} \hat{\bfR} & -\hat{\bfR} & \bf0 & \bf0 \\
        \bf0                        & \bf0        & \bfI & \bf0 \\
        \bf0                        & \bf0        & \bf0 & \bfI \\
        -\skew{\hat\bfl} \hat{\bfR} & \bf0        & \bf0 & \bf0 \\
    \end{bmatrix}  \\
    \bfH^*   & =\bfPi \bfH^*_v              \\
    \bfH_{{e}} ^* & = \begin{bmatrix}
        \bf0 & -\bfI & \bf0 & \bf0 & \bf0 & \bfI \\
    \end{bmatrix}
\end{align}

\subsection{Observability of the transformed system}
Observability refers to the ability of a system to recover its initial states using all available measurements. The set of states that cannot be recovered from measurements constitutes the unobservable subspace of the system. For the system described in \eqref{sys_sup:trans}, we can utilize the local observability matrix \cite[P. 180]{chenLinearSystemTheory1999} to conduct the observability analysis. Let $\bfM^*$ denote the local observability matrix of system \eqref{sys_sup:original}, then 
\begin{equation}
    \bfM^*
    = \begin{bmatrix}
        \bfM_0^*     \\
        \bfM_1^*     \\
        \vdots       \\
        \bfM_{n-1}^* \\
    \end{bmatrix} \label{equ_sup:O}
\end{equation}
where $\bfM_0 = \bfPi \bfH_{{e}} ^*$ and
\begin{equation}
    \bfM_{k+1}^* =   \bfM_{k}^* \bfF^* + \dot{\bfM}_k^* \quad k = 0, 1,..., n-1.
    \label{equ_sup:Mi}
\end{equation}
According to \eqref{equ_sup:Mi}, we can calculate
\begin{equation}
    \bfM_1^* = \bfPi \bfH_{{e}} ^* \bfF^* + \bfPi^{(1)}\bfH_{{e}} ^*
\end{equation}
\begin{subequations}
    \begin{align}
        \bfM_2^* & = \bfPi \bfH_{{e}} ^* {\bfF^*}^2 + \bfPi^{(1)}\bfH_{{e}} ^*\bfF^*   +\bfPi^{(1)}\bfH_{{e}} ^*{\bfF^*} +\bfPi\bfH_{{e}} ^*{\bfF^*}^{(1)} +  \bfPi^{(2)}\bfH_{{e}} ^* \\
                 & =  \bfPi (\bfH_{{e}} ^* {\bfF^*}^2+\bfPi\bfH_{{e}} ^*{\bfF^*}^{(1)}) + 2\bfPi^{(1)}\bfH_{{e}} ^*\bfF^*   + +  \bfPi^{(2)}\bfH_{{e}} ^*
    \end{align}
\end{subequations}
The observability matrix is written as
\begin{equation}
    \bfM^* = \begin{bmatrix}
        \bfPi       & \bf0         & \bf0  \\
        \bfPi^{(1)} & \bfPi        & \bf0  \\
        \bfPi^{(2)} & 2\bfPi^{(1)} & \bfPi \\
    \end{bmatrix}\begin{bmatrix}
        \bfH_{{e}} ^*        \\
        \bfH_{{e}} ^* \bfF^* \\
        \bfH_{{e}} ^* {\bfF^*}^2  +\bfH_{{e}} ^* {\bfF^*}^{(1)}
    \end{bmatrix}
\end{equation}
where
\begin{equation}
    \begin{bmatrix}
        \bfH_{{e}} ^*                                       \\
        \bfH_{{e}} ^* \bfF^*                                \\
        \bfH_{{e}} ^* {\bfF^*}^2  + \bfH_{{e}} ^* {\bfF^*}^{(1)} \\
    \end{bmatrix} = \begin{bmatrix}
        \bf0         & -\bfI & \bf0  & \cellcolor{blue!20}\bf0                                                                                                          & \cellcolor{blue!20} \bf0                                               & \bfI \\
        \bf0         & \bf0  & -\bfI & \cellcolor{blue!20}\skew{\hat\bfp - \hat\bfl}\hat\bfR                                                                            & \cellcolor{blue!20} \bf0                                               & \bf0 \\
        -\skew{\bfg} & \bf0  & \bf0  & \cellcolor{blue!20} \skew{\hat\bfv}\hat\bfR  +\skew{\hat\bfv}\hat\bfR + \skew{\hat\bfp - \hat\bfl}\hat\bfR \skew{\hat{\bfomega}} & \cellcolor{blue!20}                                           \hat\bfR & \bf0 \\
    \end{bmatrix}
    \label{equ_sup:conti}
\end{equation}
There exist four unobservable dimensions for the transformed linearized error-state system, s.t., $\bfM^*\bfN^*=\bf0$ with unobservable subspace
\begin{equation}
    \bfN^* =
    \left[
        \begin{array}{cc}
            \bf0_{3\times 3 }                      & \bfg                                 \\
            \bfI_{3}                               & \bf0_{3\times 1 }                    \\
            \bf0_{3\times 3 }                      & \bf0_{3\times 1 }                    \\
            \cellcolor{blue!20}  \bf0_{3\times 3 } & \cellcolor{blue!20}\bf0_{3\times 1 } \\
            \cellcolor{blue!20}  \bf0_{3\times 3 } & \cellcolor{blue!20}\bf0_{3\times 1 } \\
            \bfI_{3}                               & \bf0_{3\times 1 }                    \\
        \end{array}\right].
        \label{equ_sup:N*}
\end{equation}
The unobservable subspace of the transformed system is independent of the state.
When performing state estimation based on this transformed system, the erroneous reduction of unobservable dimensions is prevented, thus preserving consistency.

\begin{remark}
    For $k>2$, $\bfM_{k}^*$ is complicated but does not affect our result about the observability. This is because  only
    $\bfH_{{e}} ^*$, $ \bfH_{{e}} ^* {\bfF^*}$, and $ \bfH_{{e}} ^* {\bfF^*}^2$ contribute to observability analysis.
    If we continue to calculate $\bfM_{k}^*$ when $k>2$, the items such as $ \bfH_{{e}} ^* {\bfF^*}^m$ and $ \bfH_{{e}} ^* {\bfF^*}^m {\bfF^*}^{(n)}$ will also be included in the left side of \eqref{equ_sup:conti}. However, these terms only affect the blue-highlighted columns of \eqref{equ_sup:conti}. The nullspace blocks related to the blue-highlighted columns are always zeros, as shown in \eqref{equ_sup:N*}. Therefore, there is no need to calculate $\bfM_{k}^*$ for  $k>2$.
\end{remark}

\section{Transformed Error-State Kalman Filter}
\label{sec_sup:teskf}
\begin{figure}[!hbtp]
        \centerline{\includegraphics[width=0.5\textwidth]{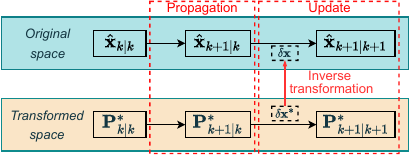}}
        \caption{Pipeline of T-ESKF.
                It propagates and updates the covariance estimates in the transformed space. The state estimate is propagated by integrating \eqref{equ_sup:dyn_nomi} and updated using the state correction derived from the transformed system. The correction is obtained in the transformed space and then transformed back to the original space.
        }
        \label{fig_sup:frames}
\end{figure}

In this section, we present the T-ESKF, a consistent VINS estimator based on the transformed linearized error-state system \eqref{sys_sup:trans}.
The pipeline of T-ESKF is shown in Fig. \ref{fig_sup:frames}.

The transformed error-state $\tilde{\bfx}^*$ matchs $\bfP^*$ and the error-state $\tilde{\bfx}$ matchs $\bfP$.
    Since $\bfP$ can always be obtained through the inverse transformation:
    \begin{equation}
        \bfP = \bfT(\hat{\bfx}) ^{-1} \bfP \bfT(\hat{\bfx}) ^{-\top}
    \end{equation}
    and it is not involved in the estimation process, we do not maintain it in T-ESKF.

\subsection{Propagation}

\begin{figure}[htbp]
    \centering
    \includegraphics[width=0.6\textwidth]{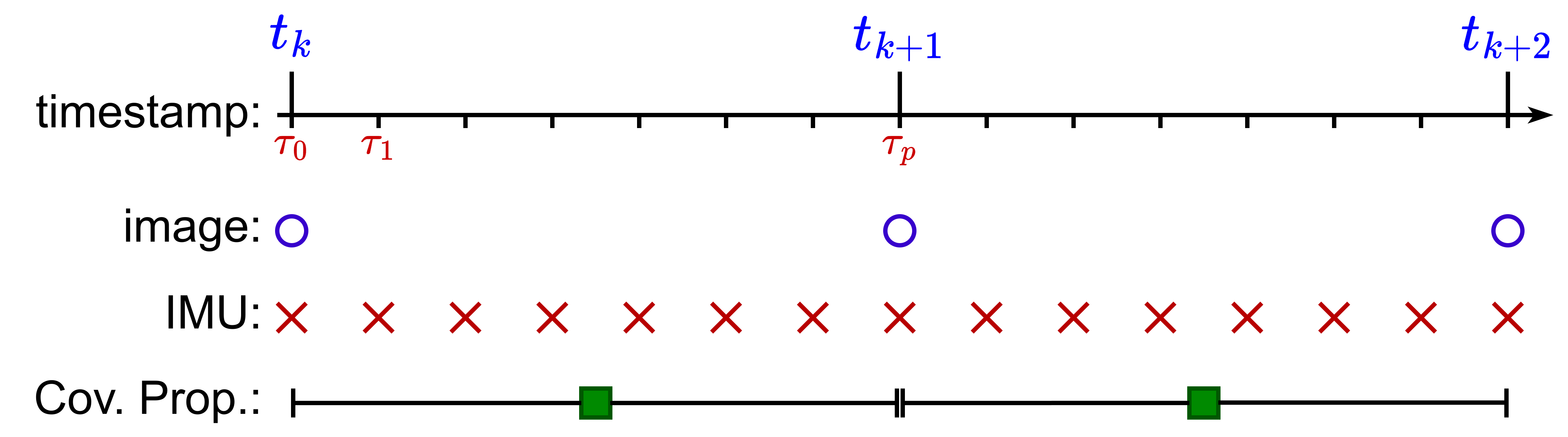}
    \caption{Different frequencies for camera and IMU measurements, along with covariance propagation. $\tau_0,\tau_1,\dots,$ are the IMU measurement timestamp and $t_k,t_{k+1},\dots$ are the camera measurement timestamp. For simplify, $\tau_0$ and $\tau_q$ coincide with $t_k$ and $t_{k+1}$, respectively.}
    \label{fig_sup:ffreq}
\end{figure}

Usually, the frequency of IMU measurement is much higher than that of the camera. Before performing update using visual measurement at timestamp $t_{k+1}$, the covariance matrix need to be propagated from $t_k$ to $t_{k+1}$, as follows:
\begin{align}
     \bfP_{k+1|k}^* & = \bfPhi_{k}^{*}  \bfP_{k|k}^* {\bfPhi_{k}^{*}}^\top + \bfQ_{k}^* 
     \label{equ_sup:P_prop}
\end{align}
where the transition matrix $\bfPhi^*_{k}\triangleq \bfPhi^*(\tau_p,\tau_0)$ is defined by the differential equation:
\begin{equation}
    \frac{\text{d}}{\text{d}\tau}{\bfPhi}^*(\tau,\tau_0) = \bfF^*{\bfPhi}^*(\tau,\tau_0)
        \label{equ_sup:int_PHI}
\end{equation}
with the initial condition ${\bfPhi}^*(\tau_0,\tau_0)= \bfI_{15+3m}.$
The accumulated noise matrix $ \bfQ_{k}^* \triangleq \bfQ^*(\tau_p,\tau_0)$ is defined as 
\begin{equation}
        \bfQ_{k}^* = \int_{\tau_0}^{\tau_{p}} \bfPhi ^* (\tau_p,\tau) \bfG^*_{\tau} \bfQ {\bfG^*_{\tau} }^\top {\bfPhi ^*(\tau_p,\tau)}^\top \text{d} \tau.
        \label{equ_sup:intG}
\end{equation}
In practice, the transition matrix $\bfPhi^*_{k}$ and the accumulated noise $\bfQ^*_{k}$ matrix are computed iteratively:
\begin{align}
    \bfPhi^*(\tau_{i+1},\tau_{0}) &=\bfPhi^*(\tau_{i+1},\tau_{i}) \bfPhi^*(\tau_{i},\tau_{0}), \quad i\in \{0, 1,.., p-1 \} \label{equ_sup:iterative_Phi}\\
    \bfQ^*(\tau_{i+1},\tau_0)   &=  \bfPhi^*(\tau_{i+1},\tau_{i})   \bfQ^*(\tau_{i},\tau_0)  \bfPhi^*(\tau_{i+1},\tau_{i})^\top + \bfG^*(\tau_{i+1},\tau_{i})   \bfQ_d  \bfG^*(\tau_{i+1},\tau_{i})^\top.   \label{equ_sup:iterative_Q}
\end{align}

When IMU bias and a large number of landmarks are included in the state vector, T-ESKF will encounter the same computational bottleneck in covariance propagation as RI-EKF \cite{yangDecoupledRightInvariant2022}.
To solve this computational bottleneck, 
instead of directly integrating \(\bfPhi^*_k\) and \(\bfQ_{k}^*\) using \eqref{equ_sup:iterative_Phi} and \eqref{equ_sup:iterative_Q} directly, we develop an efficient propagation technique to compute them. Specifically,
by leveraging \eqref{equ_sup:kdF} and \eqref{equ_sup:kdG}, we can decompose \(\bfPhi^*_k\) and \(\bfQ_{k}^*\) into the following computationally efficient form (derivations please refer to Appendix \ref{app:df3839}):
\begin{align}
      \bfPhi^*_{k} &= \bfT(\hat{\bfx}_{k+1|k}) \begin{bmatrix}
         \bfPhi_{k}^I &\bf0 \\
         \bf0 &\bfI\\
     \end{bmatrix}\bfT(\hat{\bfx}_{k|k})^{-1}, \label{equ_sup:38}  \\
     \bfQ_{k}^* &= \bfT(\hat{\bfx}_{k+1|k}) \begin{bmatrix}
         \bfQ_k^I &\bf0 \\
         \bf0 &\bfZo\\
     \end{bmatrix}\bfT(\hat{\bfx}_{k+1|k}) ^\top. \label{equ_sup:39}
\end{align}
where the IMU transition matrix $\bfPhi^I_{k} \triangleq \bfPhi^I(\tau_{p},\tau_{0}) \in \mathbb{R}^{15\times 15}$ and the IMU accumulated noise matrix $\bfQ_{k}^I \triangleq \bfQ^I(\tau_p,\tau_0) \in \mathbb{R}^{15\times 15}$ are computed iteratively:
\begin{align}
    \bfPhi^I(\tau_{i+1},\tau_{0}) &=\bfPhi^I(\tau_{i+1},\tau_{i}) \bfPhi^I(\tau_{i},\tau_{0}), \quad i\in \{0, 1,.., p-1 \} \label{equ_sup:iterative_Phi2}\\
    \bfQ^I(\tau_{i+1},\tau_0)   &=  \bfPhi^I(\tau_{i+1},\tau_{i})   \bfQ^I(\tau_{i},\tau_0)  \bfPhi^I(\tau_{i+1},\tau_{i})^\top + \bfG^I(\tau_{i+1},\tau_{i})   \bfQ_d  \bfG^I(\tau_{i+1},\tau_{i})^\top,  \label{equ_sup:iterative_Q2}
\end{align}
with
\begin{align}
    \bfPhi^I(\tau_{i+1},\tau_{i}) & = \begin{bmatrix}
        \bfI_3                                                                                          & \bf0   & \bf0            & -\hat{\bfR}_{i+1}\bfJ_r(\hat{\bfomega}_{k}\Delta t) \Delta t & \bf0                        & \bf0   \\
        \skew{\hat{\bfp}_{i}-\hat{\bfp}_{i+1}+\hat{\bfv}_{i}\Delta t + \frac{1}{2}\bfg\Delta t^2} & \bfI_3 & \bfI_3 \Delta t & \hat{\bfR}_{i}\bfXi_{4}                                      & -\hat{\bfR}_{i}\bfXi_{2} & \bf0   \\
        \skew{\hat{\bfv}_{i}-\hat{\bfv}_{i+1}+ \bfg\Delta t}                                       & \bf0   & \bfI_3          & \hat{\bfR}_{i}\bfXi_{3}                                      & -\hat{\bfR}_{i}\bfXi_{1} & \bf0   \\
        \bf0                                                                                            & \bf0   & \bf0            & \bfI_3                                                          & \bf0                        & \bf0   \\
        \bf0                                                                                            & \bf0   & \bf0            & \bf0                                                            & \bfI_3                      & \bf0   \\
        \bf0                                                                                            & \bf0   & \bf0            & \bf0                                                            & \bf0                        & \bfI_3 \\
    \end{bmatrix}, \\
    \bfG^I(\tau_{i+1},\tau_{i})  & = \begin{bmatrix}
        -{\upG}\hat{\bfR}_{i+1}\bfJ_r(\hat{\bfomega}_{k}\Delta t) \Delta t & \bf0                              & \bf0            & \bf0             \\
        {\upG}\hat{\bfR}_{i}\bfXi_{4}                                      & -{\upG}\hat{\bfR}_{i}\bfXi_{2} & \bf0            & \bf0             \\
        {\upG}\hat{\bfR}_{i}\bfXi_{3}                                      & -{\upG}\hat{\bfR}_{i}\bfXi_{1} & \bf0            & \bf0             \\
        \bf0                                                                  & \bf0                              & \bfI_3 \Delta t & \bf0             \\
        \bf0                                                                  & \bf0                              & \bf0            & \bfI_3  \Delta t \\
        \bf0                                                                  & \bf0                              & \bf0            & \bf0             \\
    \end{bmatrix}.
\end{align}
The computations of $\bfXi_1$, $\bfXi_2$, $\bfXi_3$, and $\bfXi_4$ are detailed in \href{https://docs.openvins.com/propagation_analytical.html#analytical_integration_components}{{Openvins: Analytical Integration Components}}.

Regardless of the number of landmarks included in the state vector, $\bfPhi_{k}^I$ and $\bfQ_k^I$ are always small and fixed-size matrices. Although \eqref{equ_sup:iterative_Phi2} and \eqref{equ_sup:iterative_Q2} need to be iterated $p$ times, the matrices are much smaller than those in \eqref{equ_sup:iterative_Phi} and \eqref{equ_sup:iterative_Q}. Thus, the computation for $\bfPhi^I_{k}$ and $\bfQ^I_{k}$ can be finished within a short timeframe.
 Additionally, the sparsity of $\bfT(\cdot)$ is also fully
exploited to accelerate the matrix multiplications in \eqref{equ_sup:38} and \eqref{equ_sup:39}.
Consequently, 
the covariance can be propagated efficiently in T-ESKF.

\begin{remark}
         \eqref{equ_sup:38} and \eqref{equ_sup:39} are derived in a continuous form and does not presuppose any assumptions regarding IMU model, such as the piecewise constant acceleration. Therefore, we can directly apply existing IMU integration theories, such as $\text{ACI}^2$\cite{ic2}, to compute $\bfPhi_k^I$ and $\bfQ_k^I$.
\end{remark}

% The solutions to \eqref{equ_sup:int_PHI} and \eqref{equ_sup:intG} are given in [to be cited]. 

\subsection{Update}
 After propagation, we have the prior estimation $\hat{\bfx}_{k+1|k}$ and the covariance $\bfP^*_{k+1|k}$ corresponding to the transformed error-state. The covariance corresponding to the original error-state can be calculated through the inverse transformation:
    \begin{equation}
        \bfP_{k+1|k}  = \bfT(\hat{\bfx}_{k+1|k})^{-1} \bfP^*_{k+1|k}\bfT(\hat{\bfx}_{k+1|k})^{-\top}.
    \end{equation}

    During the update steps, we start with deriving the state correction from the transformed space, then inversely transform it back into the original space.
    Let $\delta{\bfx}^*\triangleq \bfK^*\tilde{\bfy}$ denote the Kalman state correction for the transformed error-state, the transformed error-state is corrected as
    \begin{align}
        {\tilde{\bfx}^*_{k+1|k+1}} & = {\tilde{\bfx}^*_{k+1|k}} - \delta{\bfx}^*                                           \\
                                   & = {\tilde{\bfx}^*_{k+1|k}}-\bfK^*\tilde\bfy                                           \\
                                   & = {\tilde{\bfx}^*_{k+1|k}}-\bfK^*(\bfH^*\tilde{\bfx}^*_{k+1|k}+\boldsymbol{\epsilon}) \\
                                   & = (\bfI- \bfK^*\bfH^*){\tilde{\bfx}^*_{k+1|k}} - \bfK^*\boldsymbol{\epsilon}
    \end{align}
    with
    \begin{align}
        \bfP^*_{k+1|k+1} & = \bfE\left(
        ({\tilde{\bfx}^*_{k+1|k+1}}) ({\tilde{\bfx}^*_{k+1|k+1}})^\top
        \right)                                                                                                          \\
                         & =  (\bfI- \bfK^*\bfH^*)  \bfP^*_{k+1|k} (\bfI- \bfK^*\bfH^*)^\top + \bfK^* \bfV {\bfK^*}^\top.
        \label{equ_sup:P} \\
        \bfH^* &= \bfPi|_{\hat\bfx = \hat\bfx_{k+1|k}} \bfH_e^* \label{equ_sup:bfH*}
    \end{align}
    Note that $\bfP_{k+1|k+1}^*$ is the covariance for $\tilde{\bfx}^*_{k+1|k+1} = {\tilde{\bfx}^*_{k+1|k}} - \delta{\bfx}^*$. 
    The Kalman gain $\bfK^*$ is obtained by minisizing the trace of $\bfP_{k+1|k+1}^*$, and it has the same form as the standard Kalman gain:
    \begin{equation}
        \bfK^*  = \bfP^*_{k+1|k} {\bfH^*_{k+1}}^\top (\bfH^*_{k+1}\bfP^*_{k+1|k}{\bfH^*_{k+1}}^\top  + \bfV ) ^{-1}.
        \label{equ_sup:K}
    \end{equation}
    Substituting \eqref{equ_sup:K} into \eqref{equ_sup:P} yields
    \begin{equation}
        \bfP_{k+1|k+1}^*  =\bfP_{k+1|k}^* - \bfK^* \bfH^*_{k+1} \bfP_{k+1|k}^*.
    \end{equation}
    
    Subsequently, we derive the state correction in the original space.  Let $\delta\bfx$ denote the state correction in the original space, then we have
    \begin{equation}
        \hat{\bfx}_{k+1|k+1} = \hat{\bfx}_{k+1|k} \oplus \delta \bfx.
    \end{equation}
   Correspondingly, the error-state is updated as
    \begin{equation}
        \tilde{\bfx}_{k+1|k+1} = \tilde{\bfx}_{k+1|k} - \delta \bfx
        \label{equ_sup:error_o}.
    \end{equation}
    The covariance for $\tilde{\bfx}_{k+1|k+1}$ is computed by the inverse transformation:
    \begin{align}
        \bfP_{k+1|k+1} & =\bfT(\hat{\bfx}_{k+1|k+1})^{-1} \bfP_{k+1|k+1}^* \bfT(\hat{\bfx}_{k+1|k+1})^{-\top}
        \label{equ_sup:cov_t}
    \end{align}
    To ensure that $\tilde{\bfx}_{k+1|k+1}$ in \eqref{equ_sup:error_o} matches $\bfP_{k+1|k+1}$ in \eqref{equ_sup:cov_t}, $\delta \bfx$ must satisfy
    \begin{equation}
        \tilde{\bfx}_{k+1|k} - \delta \bfx= \bfT(\hat{\bfx}_{k+1|k+1})^{-1} (\tilde{\bfx}_{k+1|k}^* - \delta\bfx^*).
    \end{equation}
    Then we have
    \begin{equation}
        \delta\bfx = \Big(\bfI - \bfT(\hat{\bfx}_{k+1|k+1})^{-1}\bfT(\hat{\bfx}_{k+1|k})\Big) \tilde{\bfx}_{k+1|k} + \bfT(\hat{\bfx}_{k+1|k+1})^{-1}\delta\bfx^*
        \label{equ_sup:deltax}.
    \end{equation}
    Note that $(\bfI - \bfT(\hat{\bfx}_{k+1|k+1})^{-1}\bfT(\hat{\bfx}_{k+1|k})\Big) \approx \bf0$, thus \eqref{equ_sup:deltax} can be rewritten as
    \begin{equation}
        \delta\bfx = \bfT(\hat{\bfx}_{k+1|k+1})^{-1}\delta\bfx^*\label{equ_sup:tekf1}.
    \end{equation}

  The detailed derivation of the updating using \eqref{equ_sup:tekf1} is provided in Appendix \ref{app:E1}, which outlines the calculation of  $\hat{\bfx}_{k+1|k+1}$ through an analytical solution. 

Another updating method is using $\hat{\bfx}_{k+1|k}$ instead of $\hat{\bfx}_{k+1|k+1}$ in \eqref{equ_sup:tekf1}, that is
    \begin{equation}
        \delta\bfx = \bfT(\hat{\bfx}_{k+1|k})^{-1}\delta\bfx^* \label{equ_sup:tekf0}.
    \end{equation}
We also provide the derivation of updating with \eqref{equ_sup:tekf0} in Appendix \ref{app:E2}. 

We verified these two update equations. The results indicate that updating through \eqref{equ_sup:tekf1} or \eqref{equ_sup:tekf0} yields consistent performance, both showing nearly identical outcomes, despite being evaluated at different estimates. In the manuscript, we adopt the updating equation in \eqref{equ_sup:tekf0} as it is easy to follow. 
\subsection{T-ESKF properties}
\label{sec_sup:property}
We now show that T-ESKF has the optimal Jacobians and correct observability.
Since the Jacobians in T-ESKF, $\bfPhi^*$ and $\bfH^*$, are evaluated at the current best estimates,
the optimality of Jacobians is automatically preserved.
The observability matrix for T-ESKF is:
\begin{equation}
        \bfM_{\text{T-ESKF}} = \begin{bmatrix}
                \bfH^*_{0} \\
                \bfH^*_{1} \bfPhi^*(t_1,t_0) \\
                \vdots\\
                \bfH^*_{k} \bfPhi^*(t_k,t_{0}) \\
        \end{bmatrix}
\end{equation}

According \eqref{equ_sup:bfH*} and \eqref{equ_sup:38}, we have 
\begin{equation}
    \bfH^*_{i}\bfPhi^*(t_i,t_{0}) = \bfPi_{\hat{\bfx} = \hat{\bfx}_{i|i+1}} \begin{bmatrix}
        - (t_i-t_0)^2[\bfg]_\times & -\bfI_3 &-\bfI_3 (t_i-t_0) & \mathbf{\Gamma}_{i,1} &\mathbf{\Gamma}_{i,2} & \bfI_3
    \end{bmatrix}
\end{equation}
where $ \mathbf{\Gamma}_{i,1}$ and $ \mathbf{\Gamma}_{i,2} \in \mathbb{R}^{3\times 3}$.
We can directly find that the unobservable subspace for T-ESKF with $ \bfM_{\text{T-ESKF}}  \bfN_{\text{T-ESKF}} = \bf0$ is 
\begin{equation}
        \bfN_{\text{T-ESKF}} = \begin{bmatrix}
                \bf0_{3\times 3 } & \bfg              \\
                \bfI_{3}          & \bf0_{3\times 1 } \\
                \bf0_{3\times 3 } & \bf0_{3\times 1 } \\
                \bf0_{3\times 3 } & \bf0_{3\times 1 } \\
                \bf0_{3\times 3 } & \bf0_{3\times 1 } \\
                \bfI_{3}          & \bf0_{3\times 1 } \\
        \end{bmatrix}.
        \label{equ_sup:N_teskf}
\end{equation}

\eqref{equ_sup:N*} and \eqref{equ_sup:N_teskf} indicate that erroneous reductions of unobservable dimensions are prevented and  T-ESKF has the correct observability. Consequently, T-ESKF does not encounter estimation inconsistencies arising from observability mismatches.

\newpage
\section{Real-World Experiments}
\label{sec_sup:real}

\begin{figure}[!ht]
    \centering
    \subfigure[]{
        \includegraphics[width=0.32\linewidth]{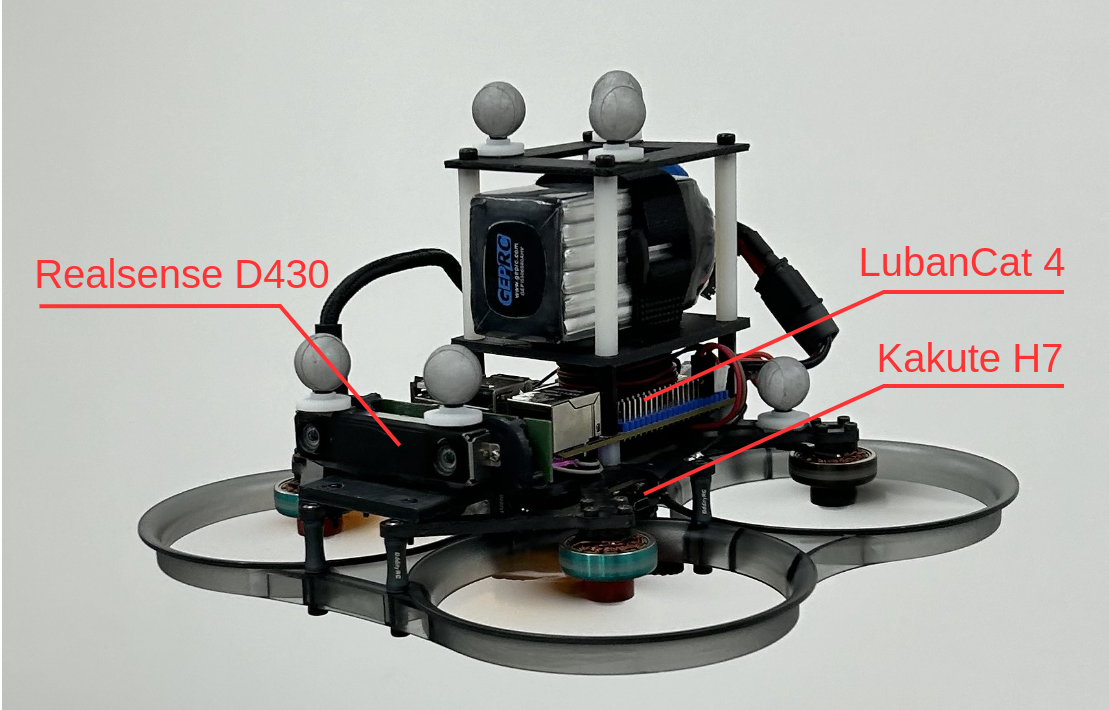}}
    \subfigure[]{
        \label{fig_sup:traj}
        \includegraphics[width=0.492\linewidth]{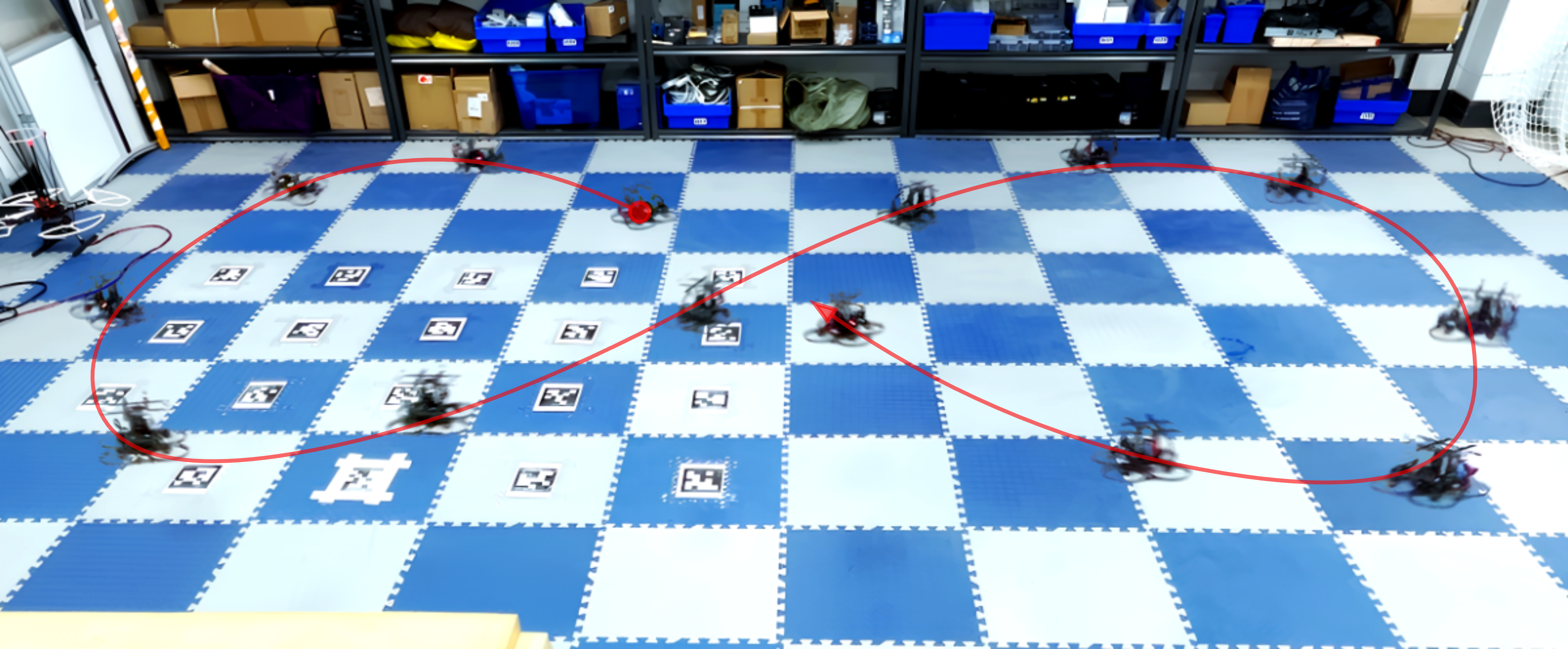}}
    \subfigure[]{
        \includegraphics[width=0.827\linewidth]{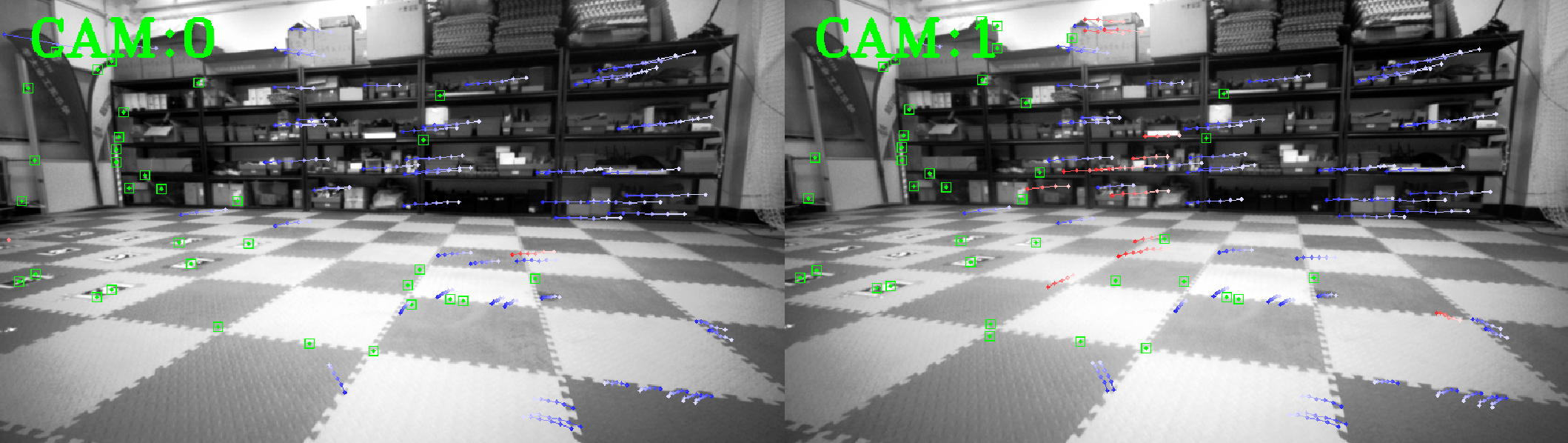}}
    \caption{(a) Aerial robot with a Realsense D430 stereo camera, a Kakute H7 flight controller (MPU6000), and an onboard computer, LubanCat 4. (b) Figure-eight flight trajectory with unfixed yaw. (c) A sample frame with tracked features in the experiment.
    }
    \label{fig_sup:platform}
\end{figure}

In addition to the dataset experiments, we further compare T-ESKF with  T-ESKF with ESKF\cite{eskf}, FEJ-ESKF\cite{huangAnalysisImprovementConsistency2008}, and RI-EKF\cite{zhangConvergenceConsistencyAnalysis2017} using our customized sensor platform mounted on an aerial robot, as depicted in Figure \ref{fig_sup:platform}. This platform provides stereo images at 30Hz with a resolution of 848 $\times$ 480 and IMU data at 200Hz, with the main parameters outlined in Table \ref{tab:exp_params}. The camera intrinsic and extrinsic parameters are calibrated using the offline calibration toolbox Kalibr. For robust performance, camera intrinsics and time offset calibrations are enabled.

% Besides the dataset experiments, we futher compare T-ESKF with ESKF\cite{eskf}, FEJ-ESKF\cite{huangAnalysisImprovementConsistency2008}, and RI-EKF\cite{zhangConvergenceConsistencyAnalysis2017} 
% using our custom platform on an aerial robot, as shown in Fig.  \ref{fig_sup:platform}. This platform provides stereo images at 30Hz with 848 $\times$ 480 resolution and IMU data at 200Hz. The main parameters are given in Table \ref{tab:exp_params}.
% The camera internal and external parameters is calibrated by the offline calibration toolbox Kalibr\cite{kalibr}.
% To achieve a robust performance, camera intrinsics and time offset calibrations are enable. 

% The aerial robot is commanded to track a figure-eight pattern or a circle pattern with  fixed or unfixed yaw, as llustrated in Fig.  \ref{fig_sup:designed_traj}. For each case, we collect three set of data, including the IMU and stereo camera measurements and the groundtruth obtained through the motion caption system.

\begin{table}[!htp]
    \caption{Real-world experiment configuration}
    \centering
    \begin{tabular}[]{cccc}
        \toprule
        { \textbf{Parameter}} & \textbf{Value} & \textbf{Parameter} & \textbf{Value} \\
        \midrule
        Accel. White Noise    & 6.33e-03       & Gyro White Noise   & 8.71e-04       \\
        Accel. Random Walk    & 2.87e-03       & Gyro Random Walk   & 3.34e-05       \\
        Pixel Noise           & 1              & IMU Freq.          & 200            \\
        Max Cam Pts/Frame     & 200            & Cam Freq.          & 30             \\
        Max Feats/Frame       & 60             & Cam Number         & Stereo         \\
        Max Clone Size        & 11             & Feat. Rep.         & Global XYZ     \\
        \bottomrule
    \end{tabular}
    \label{tab:exp_params}
\end{table}

\begin{figure}[!ht]
    \centering
    \subfigure[circle with fixed yaw]{
        \label{fig_sup:0N}
        \includegraphics[width=0.3\linewidth]{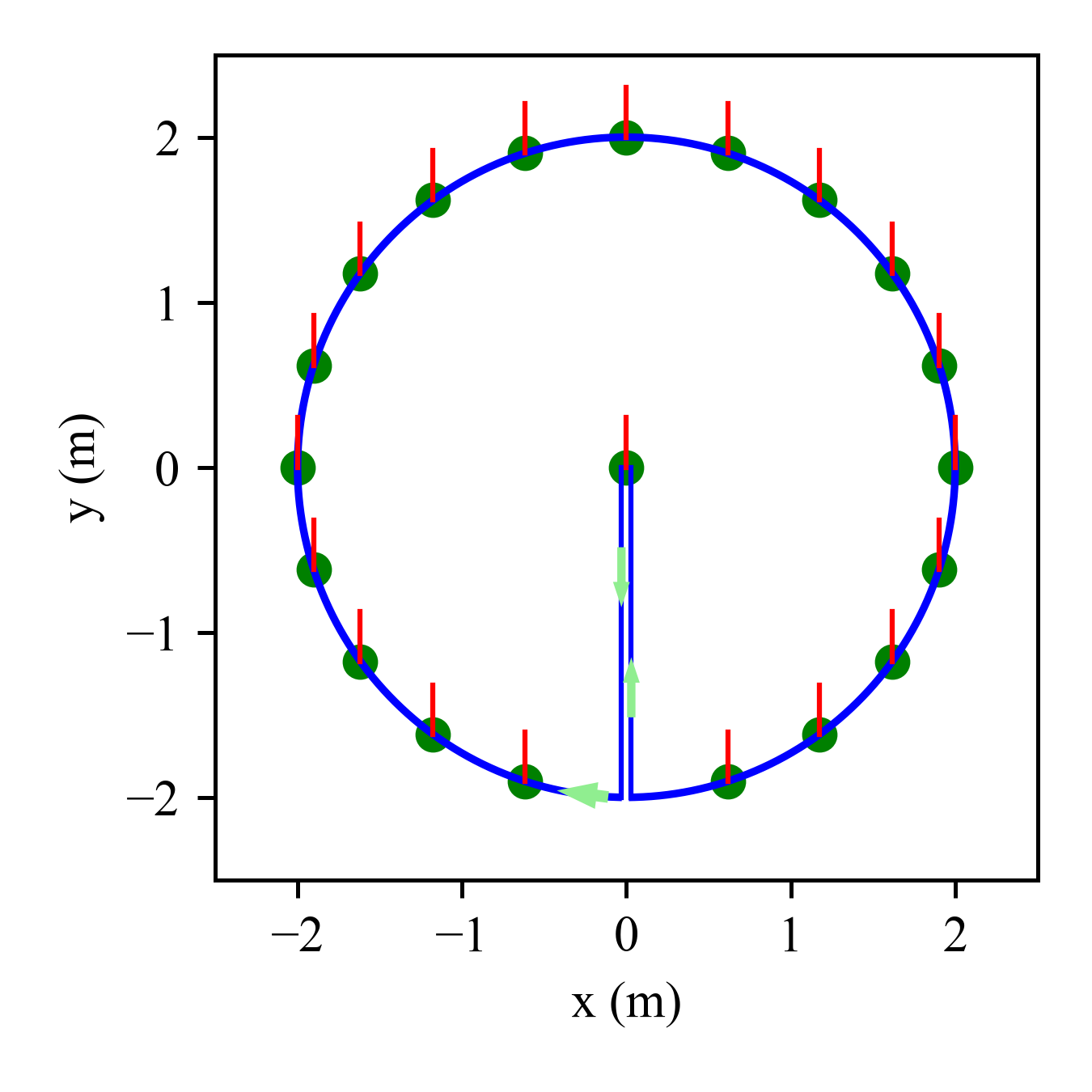}}
    \subfigure[ﬁgure-eight with fixed yaw]{
        \label{fig_sup:8N}
        \includegraphics[width=0.4\linewidth]{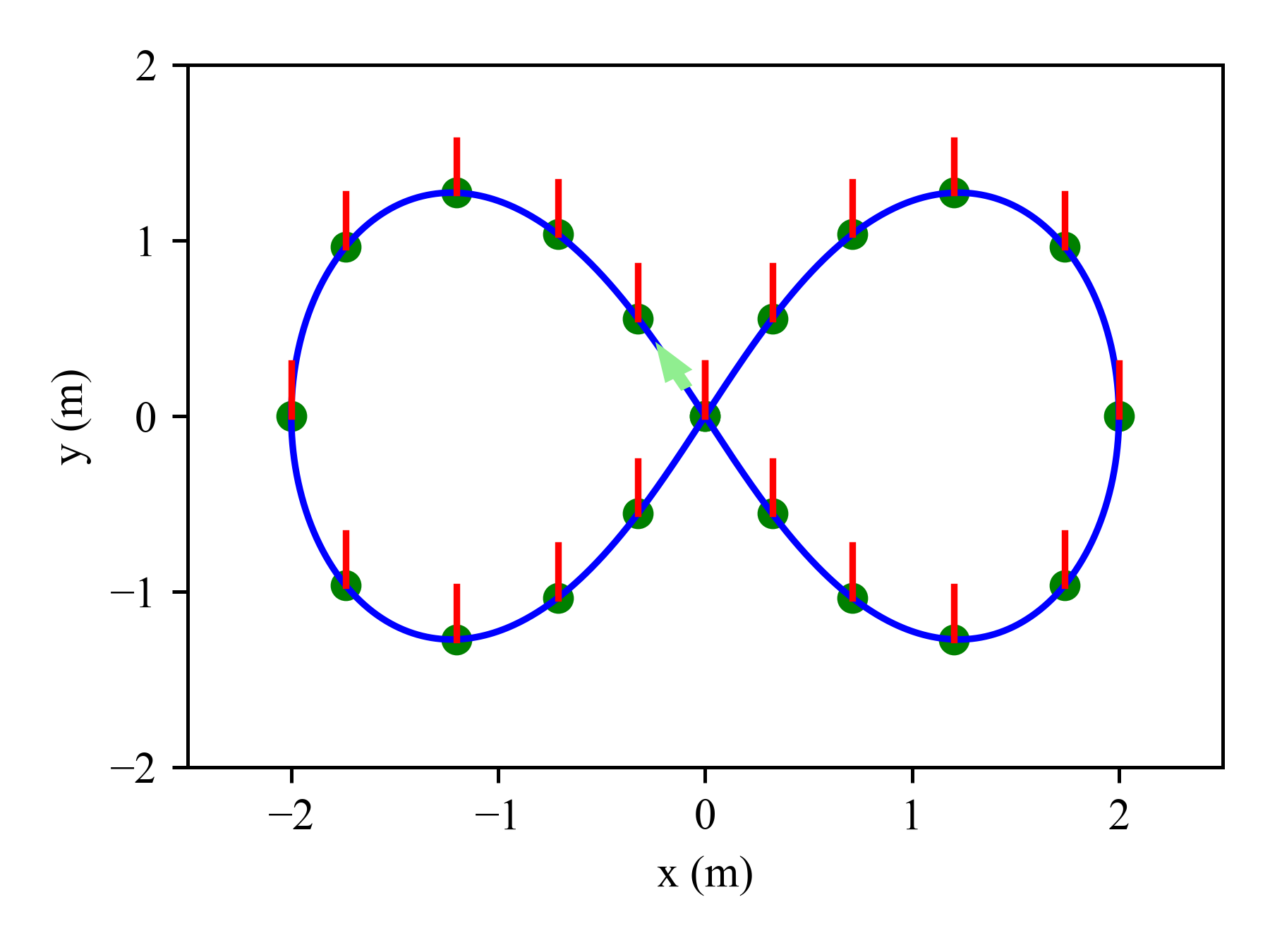}}
    \subfigure[circle with unfixed yaw]{
        \label{fig_sup:0Y}
        \includegraphics[width=0.3\linewidth]{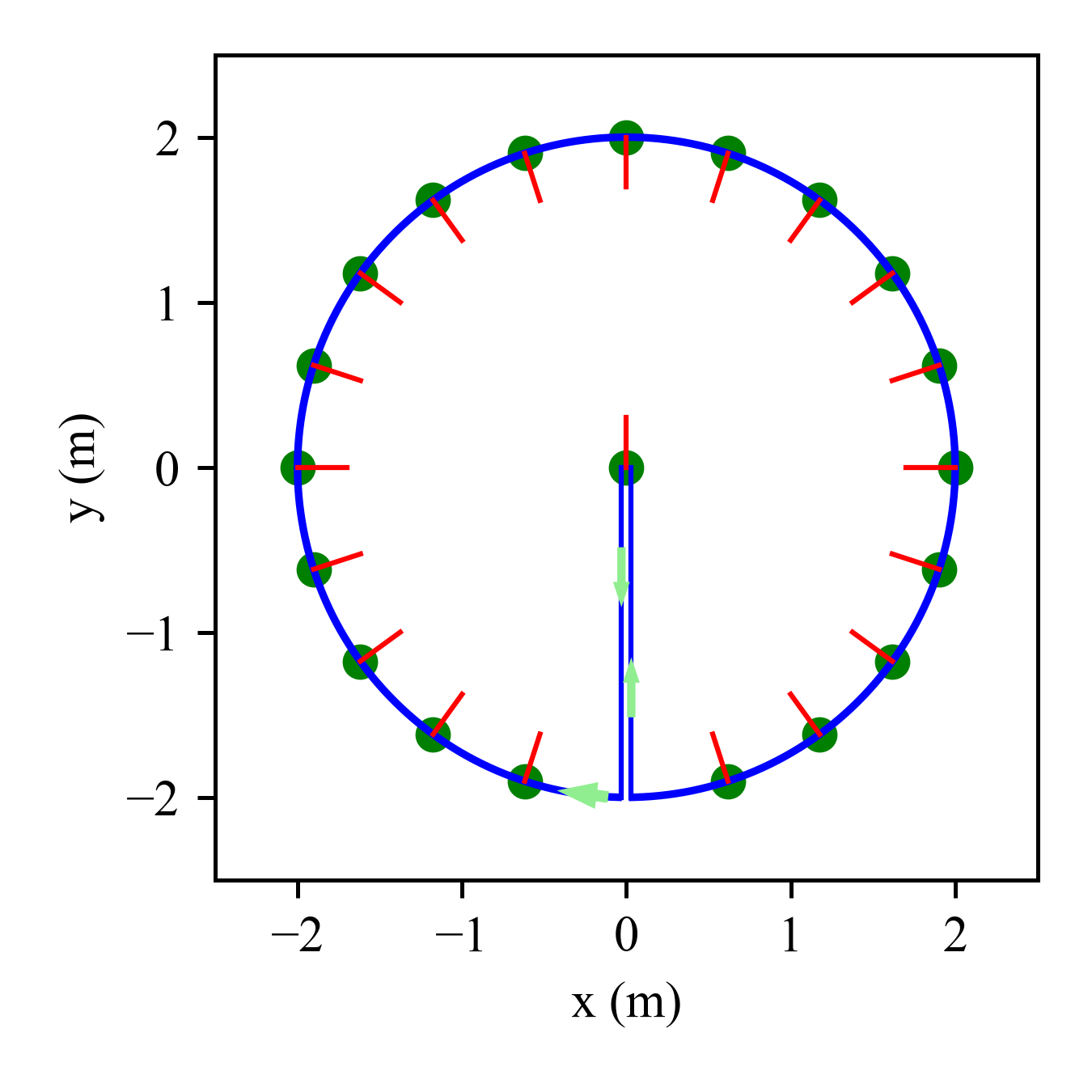}}
    \subfigure[ﬁgure-eight with unfixed yaw]{
        \label{fig_sup:8Y}
        \includegraphics[width=0.4\linewidth]{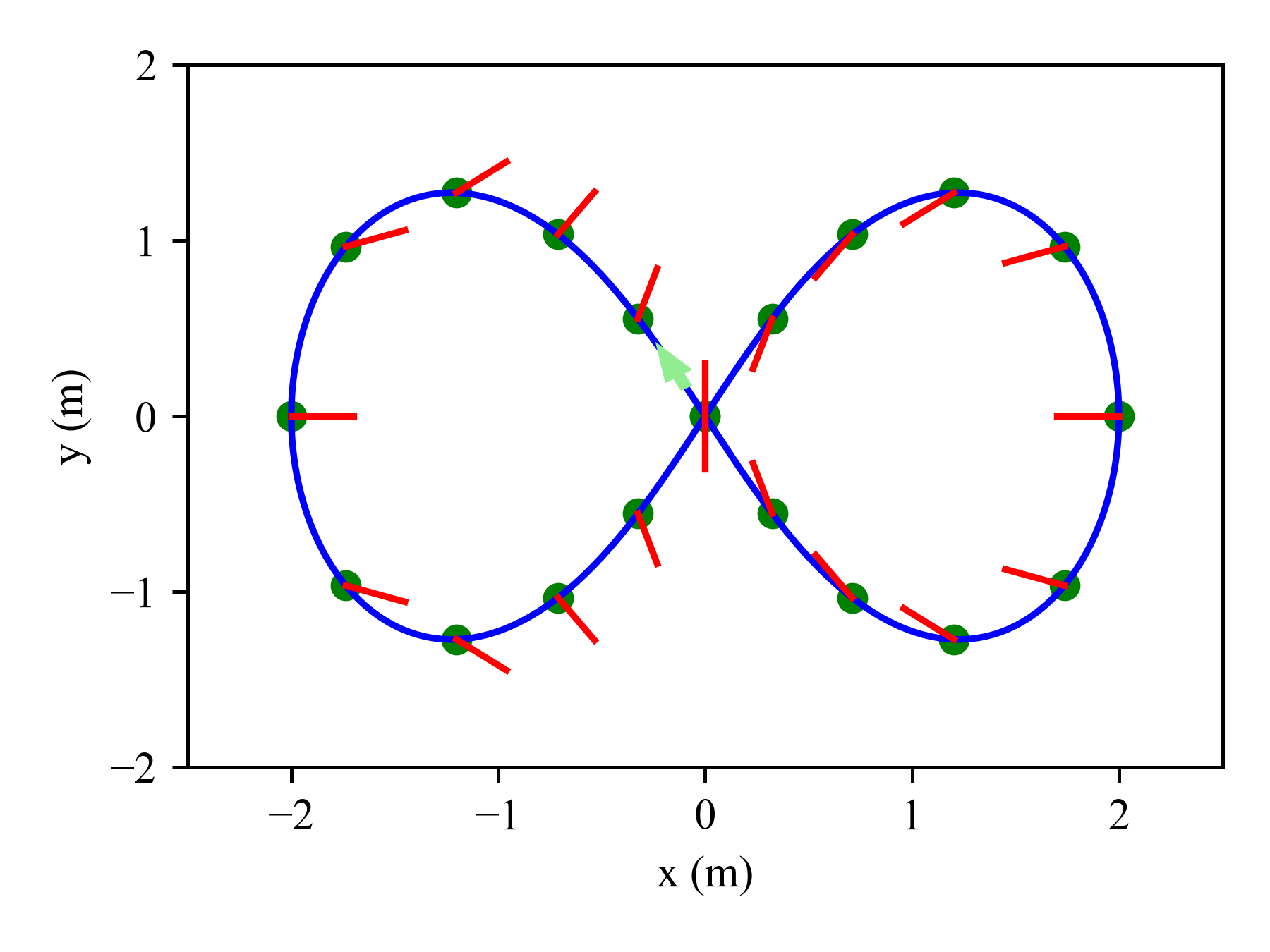}}
    \caption{
        Designed trajectories in real-world experiments.
        The circle or the figure-eight pattern is repeated 6 times in each trajectory.
        The red lines represent the orientation of the aerial robot. The lightgreen arrows indicate the direction of the trajectories.
    }
    \label{fig_sup:designed_traj}
\end{figure}

% Fig.  \ref{fig_sup:exp_time2} displays the processing time required for each frame. The
% the average time taken to process a single frame is 30.0 ms, indicating that T-ESKF is able to operate in real-time with limited onboard computational resources.

\begin{figure}[!ht]
    \centering
    \subfigure[Circle\_1 with fixed yaw]{
        \label{fig_sup:exp0N}
        \includegraphics[width=0.27\linewidth]{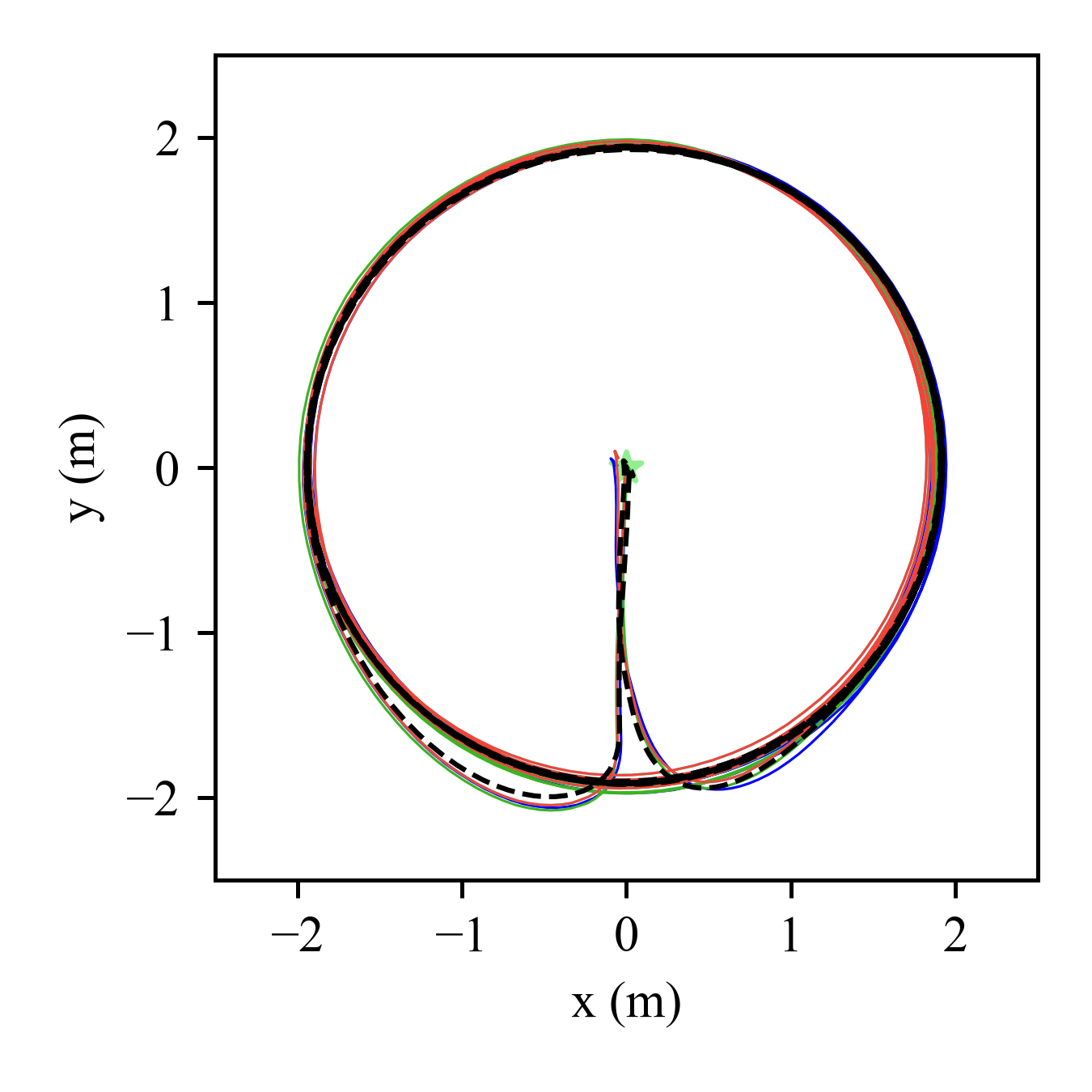}}
    \subfigure[Eight\_1 with fixed yaw]{
        \label{fig_sup:exp8N}
        \includegraphics[width=0.37\linewidth]{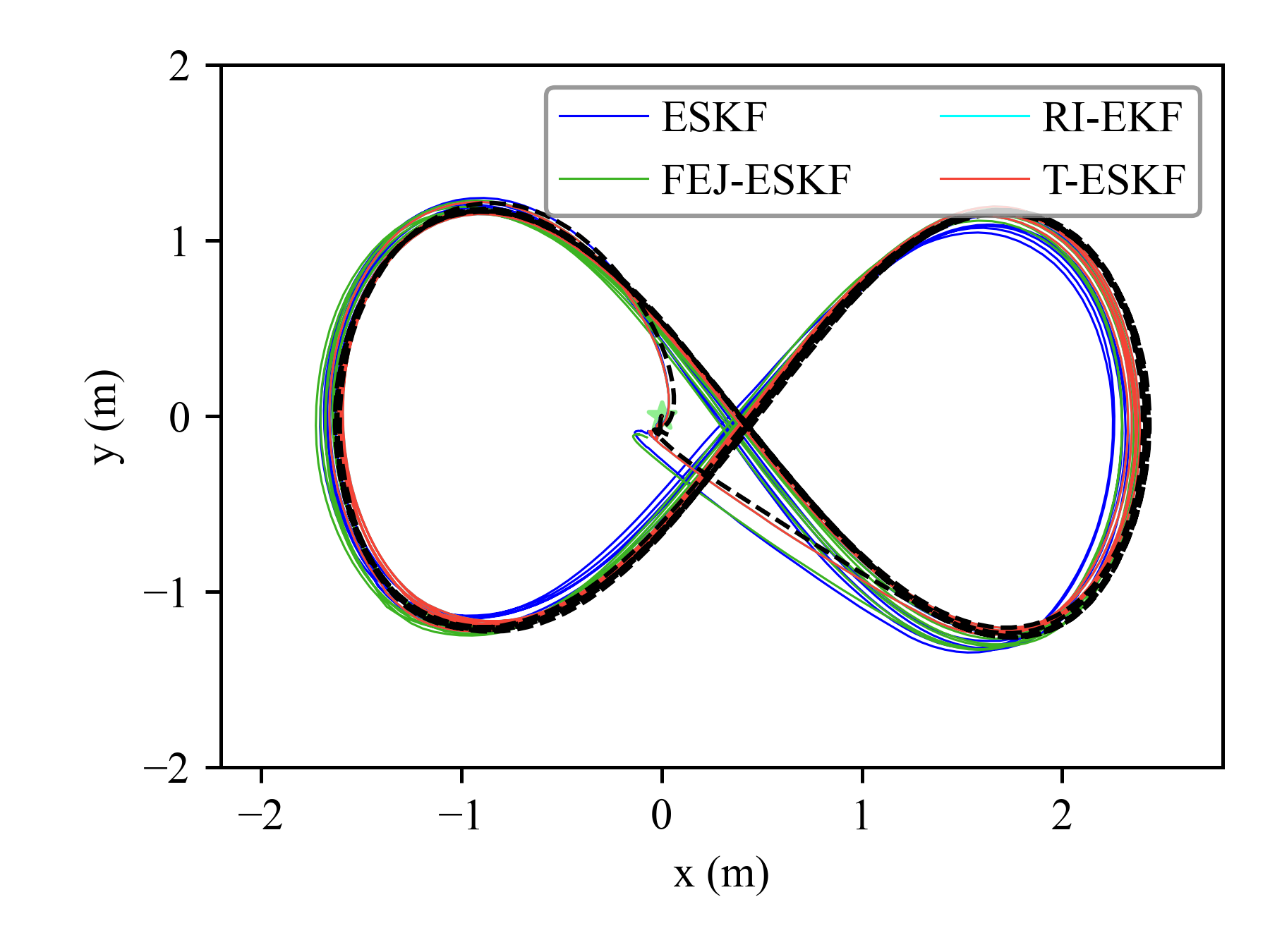}}\\
    \centering
    \subfigure[Circle\_1 with unfixed yaw]{
        \label{fig_sup:exp0Y}
        \includegraphics[width=0.27\linewidth]{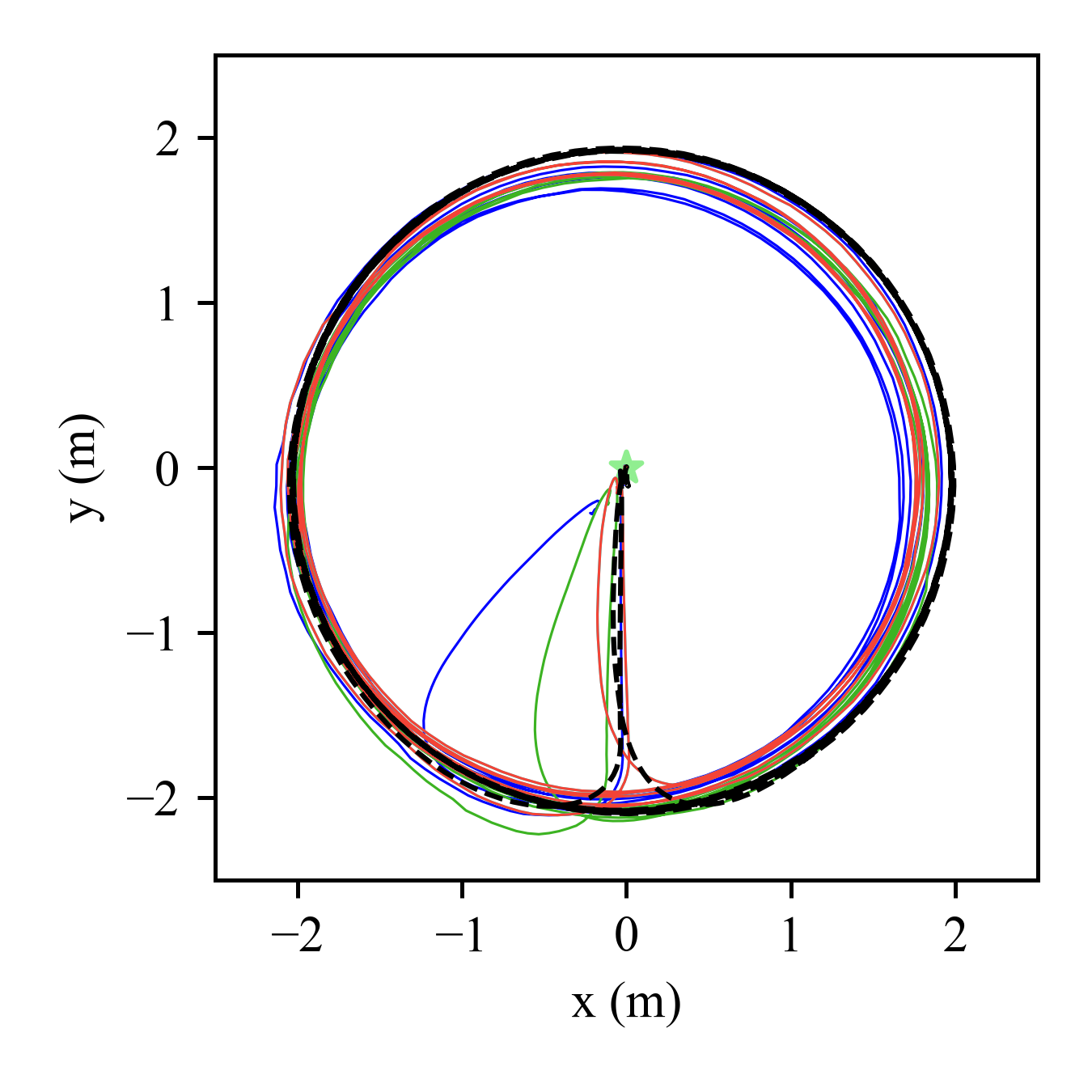}}
    \subfigure[Eight\_1 with unfixed yaw]{
        \label{fig_sup:exp8Y}
        \includegraphics[width=0.37\linewidth]{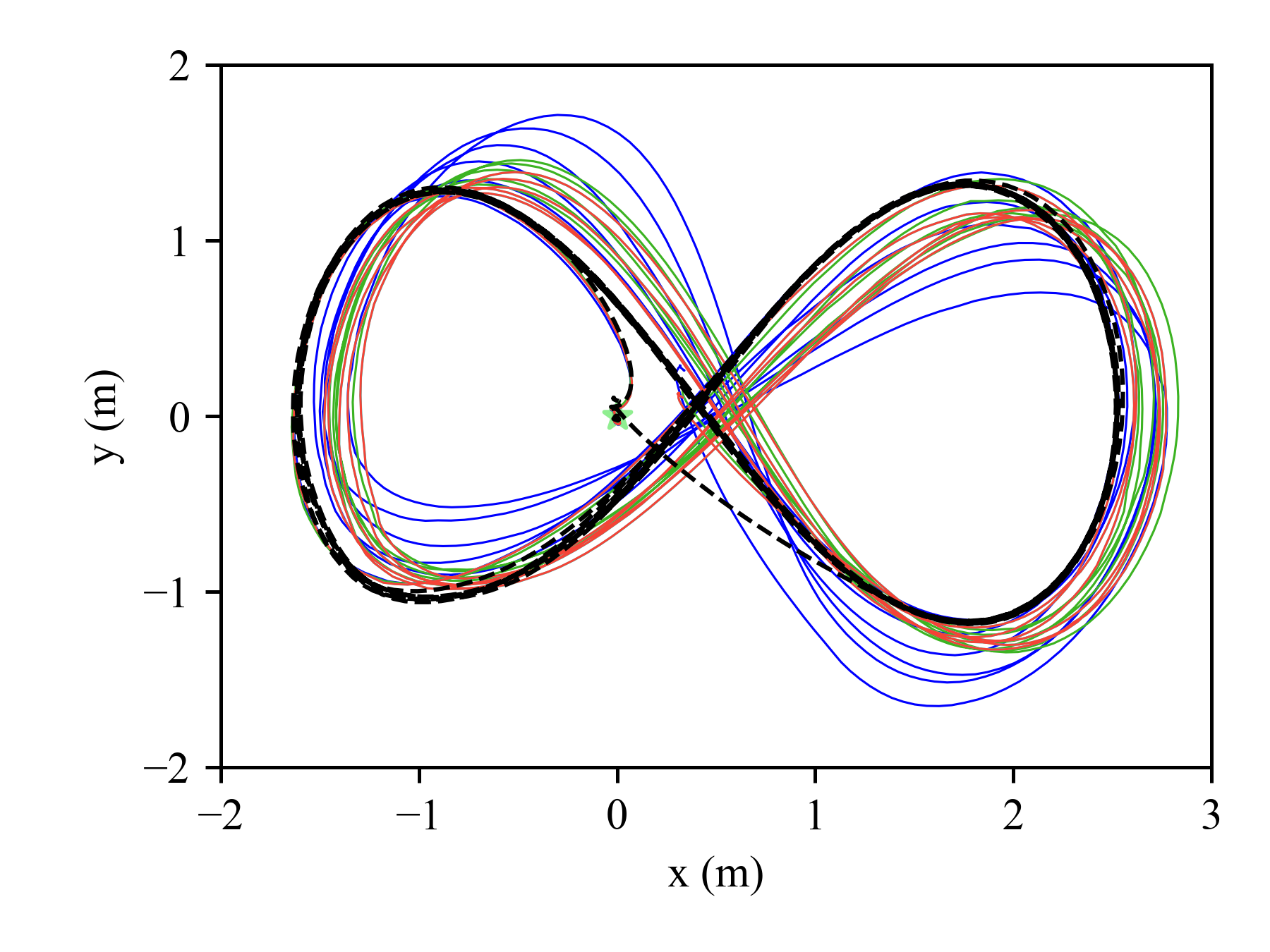}}
    \caption{
        Estimated trajectories on the first dataset in each case. 
        The black dashed line is the groundtruth trajectory with a green star marking the starting point.
        The estimated trajectories are aligned to the groundtruth trajectory (black dashed line) using the beginning frame.
    }
    \label{fig_sup:traj_exp}
\end{figure}

\begin{table}[!ht]
    \caption{Average orientation (deg) and position (meter) RMSE of the real-world experiments}
    \centering
    \begin{threeparttable}          %这行要添加
        \begin{tabular}{cccccc}
            \toprule
                                                         & {Set}            & {ESKF}          & {FEJ-ESKF}      & {RI-EKF}               & {T-ESKF}                 \\
            \midrule
\multirow{7}{*}{\rotatebox{90}{Fixed yaw}} &  Circle\_1            &  1.432 /  0.066 &  \textbf{1.261} /  \textbf{0.054} &  1.270 /  0.061 &  1.270 /  0.061\\
                                           &  Circle\_2            &  1.236 /  0.057 &  1.492 /  0.059 &  1.114 /  \textbf{0.047} &  \textbf{1.113} /  \textbf{0.047}\\
                                           & Circle\_3         &  5.309 /  0.165 &  1.469 /  0.076 &  1.216 /  \textbf{0.068} &  \textbf{1.198} /  0.075\\
                                           & Eight\_1          &  2.333 /  0.114 &  1.017 /  0.085 &  \textbf{0.878} /  \textbf{0.058} &  \textbf{0.878} /  \textbf{0.058}\\
                                           & Eight\_2        &  3.882 /  0.142 &  1.884 /  0.133 &  1.539 /  0.135 &  \textbf{1.434} /  \textbf{0.126}\\
                                           & Eight\_3         &  2.885 /  0.105 &  1.373 /  0.100 &  1.148 /  \textbf{0.096} &  \textbf{0.940} /  0.103\\
                                           & Average               &  2.846 /  0.108 &  1.416 /  0.084 &  1.194 /  \textbf{0.078} &  \textbf{1.139} /  \textbf{0.078}\\
            \midrule
\multirow{7}{*}{\rotatebox{90}{Unfixed yaw}} & Circle\_1            & 26.617 /  0.665 & 11.655 /  0.402 &  \textbf{2.734} /  \textbf{0.128} &  2.747 /  \textbf{0.128}\\
                                             &  Circle\_2          & 25.674 /  0.727 &  2.334 /  \textbf{0.104} &  \textbf{2.180} / 0.137 &  \textbf{2.180} /  0.137\\
                                             &  Circle\_3         & 25.023 /  0.689 &  2.056 /  0.134 &  2.633 /  \textbf{0.124} &  \textbf{1.555} /  0.125\\
                                             & Eight\_1          & 12.833 /  0.366 &  4.712 /  0.245 &  \textbf{3.702} /  \textbf{0.216} &  3.716 /  \textbf{0.216}\\
                                             &Eight\_2         & 12.447 /  0.397 &  6.360 /  0.349 &  6.456 /  0.277 &  \textbf{6.138} /  \textbf{0.273}\\
                                             & Eight\_3 & 15.979 /  0.440 &  6.534 /  \textbf{0.355} &  6.440 /  0.368 &  \textbf{6.438} /  0.368\\
                                             & average               & 19.762 /  0.547 &  5.609 /  0.265 &  4.024 /  \textbf{0.208} &  \textbf{3.796} /  \textbf{0.208}\\
            \bottomrule
        \end{tabular} \label{tab:rmse}
    \end{threeparttable}       %这行要添加，到这里结束
\end{table}

\newpage
The aerial robot is commanded to follow either a circular or a figure-eight trajectory with fixed or unfixed yaw, as illustrated in Figure \ref{fig_sup:designed_traj}.
The time taken to track a circle is 6.28 seconds, and for a figure-eight, it is 10.47 seconds.
Each trajectory includes six repetitions of the circular or figure-eight patterns.
Three datasets are gathered for each scenario, consisting of IMU and stereo camera measurements, in addition to groundtruth acquired from the motion capture system.

To evaluate the estimated results, we align the estimated trajectories and the ground truth based on the initial frame, as depicted in Figure \ref{fig_sup:traj_exp}. The RMSE of these estimators is detailed in Table \ref{tab:rmse}. As seen, the performance of these estimators is close when the yaw is fixed. Nonetheless, T-ESKF surpasses ESKF, particularly in orientation estimation. In the case of unfixed yaw, T-ESKF exhibits superior performance compared to ESKF and FEJ-ESKF.

\appendices
\newpage
\section{Derivation for (28) in the manuscript}
\label{app:a}
To make the transformed Jacobians $\bfF^*$ and $\bfH_{{{e}}}^*$ independent of states, an intuitive idea is to transform these green-highlighted sub-matrices exclusively while leaving the remaining sub-matrices unchanged. Thus, the transformation can be designed as a lower triangular matrix with the diagonal being the identity:
\begin{equation}
        \bfT = \left[
                \begin{array}{c:ccc}
                        \bfI_3                         & \bf0   & \bf0   & \bf0   \\
                        \hdashline
                        \cellcolor{green!0}  \bfT_\bfp & \bfI_3 & \bf0   & \bf0   \\
                        \cellcolor{green!0}  \bfT_\bfv & \bf0   & \bfI_3 & \bf0   \\
                        \cellcolor{green!0}  \bfT_\bfl & \bf0   & \bf0   & \bfI_3 \\
                \end{array}
                \right]
        \label{equ_sup:Tk},
\end{equation}
where $\bfT_\bfp , \bfT_\bfv, \bfT_\bfl$ are sub-matrices to be determined such that $\bfF^*$ and $\bfH_{{{e}}}^*$ are independent of the states.

\begin{align}
        \bfF^*   & = \left[
                \begin{array}{c:ccc}
                        \bf0                          & \bf0 & \bf0   & \bf0 \\
                        \hdashline
                        \cellcolor{green!20}   \dot{\bfT}_{\bfp} - {\bfT}_{\bfv}& \bf0 & \bfI_3 & \bf0 \\
                        \cellcolor{green!20}    \dot{\bfT}_{\bfv} - \skew{\hat{\bfR}\bfa_m}   & \bf0 & \bf0   & \bf0 \\
                        \cellcolor{green!20}  \dot{\bfT}_{\bfl} & \bf0 & \bf0   & \bf0 \\
                \end{array}
                \right],  \label{equ_sup:bfF1} \\
        \bfH_{{{e}}}^* & =\left[
                \begin{array}{c:ccc}
                        \cellcolor{green!20} \skew{\upG \hat\bfl- \upG \hat\bfp } +{\bfT}_{\bfp} -{\bfT}_{\bfl} & -\bfI_3 & \bf0 & \bfI_3
                \end{array} \label{equ_sup:bfHv2}
                \right].
\end{align}
To ensure $\bfF^*$ and $\bfH_{{{e}}}^*$ independent of states,  
the green-highlighted blocks in \eqref{equ_sup:bfF1} and \eqref{equ_sup:bfHv2} are required to be constant, i.e.,

\begin{align}
    \bfC_1 & = \dot{\bfT}_{\bfp} - {\bfT}_{\bfv}                                  \label{equ_sup:c1}  \\
    \bfC_2 & = \dot{\bfT}_{\bfv} - \skew{\hat{\bfR}\bfa_m}                        \label{equ_sup:c2}  \\
    \bfC_3 & = \dot{\bfT}_{\bfl}                                                   \label{equ_sup:c3} \\
    \bfC_4 & = \skew{\upG \hat\bfl- \upG \hat\bfp } +{\bfT}_{\bfp} -{\bfT}_{\bfl} \label{equ_sup:c4}
\end{align}
Note that $\bfC_1,\bfC_2,\bfC_3$, and $\bfC_4$ are constant matrices.
Taking the derivative of \eqref{equ_sup:c1} and \eqref{equ_sup:c3} and the second order derivative of \eqref{equ_sup:c4}, we have
\begin{align}
    \bf0 & = \ddot{\bfT}_\bfp- \dot{\bfT}_\bfv \label{equ_sup:c1d}                             \\
    \bf0 & = \ddot{\bfT}_\bfl      \label{equ_sup:c3d}                                         \\
    \bf0 & = -\skew{\upG \hat\bfa } + \ddot{\bfT}_\bfp -\ddot{\bfT}_\bfl  \label{equ_sup:c4dd}
\end{align}
By combining these equations with \eqref{equ_sup:c2}+\eqref{equ_sup:c1d}-\eqref{equ_sup:c3d}-\eqref{equ_sup:c4dd}, we get $\bfC_2$:
\begin{equation}
    \bfC_2 = \skew{\hat{\bfa}-\hat{\bfR}\bfa_m}=\skew{\bfg}.
    \label{equ_sup:C2}
\end{equation}

Back substituting \eqref{equ_sup:C2} into \eqref{equ_sup:c1} - \eqref{equ_sup:c4}, we can obtain the solution as follows:
\begin{equation}
        \left\{
    \begin{array}{l}
             \bfT_\bfp = [\hat{\bfp}]_\times + t\bfC_3 +\bfC_5 \\
        \bfT_\bfv = [\hat{\bfv}]_\times + \bfC_3 -\bfC_1\\
        \bfT_\bfl = [\hat{\bfl}]_\times + t\bfC_3  -\bfC_4 + \bfC_5
    \end{array}
    \right.
    \label{equ_sup:solu}
\end{equation}
where $\bfC_5$ is an integral constant. 
Among these constant sub-matrices, $\bfC_1,\bfC_3,\bfC_4,\bfC_5$ can be arbitrarily chosen. To simplify the transformation matrix, we choose 
\begin{equation}
    \bfC_1=\bf0, \quad \bfC_3=\bf0, \quad  \bfC_4=\bf0, \quad \bfC_5=\bf0.
    \label{equ_sup:c15}
\end{equation}
Substituting \eqref{equ_sup:c15} into \eqref{equ_sup:solu}, we get:
\begin{equation}
    \bfT_\bfp = [\hat{\bfp}]_\times, \quad  \bfT_\bfv= [\hat{\bfv}]_\times,\quad  \bfT_\bfl = [\hat{\bfl}]_\times.
    \label{equ_sup:Tpvl}
\end{equation}

\section{Derivation for \eqref{equ_sup:38} and \eqref{equ_sup:39}}
\label{app:df3839}
Let $\bfPhi_k$ and $\bfQ_k$ be the transition matrix and accumulated noise matrix of ESKF. 
According to the definations of $\bfPhi$ and $\bfPhi^*$ and the relationship between $\bfF$ and $\bfF^*$, i.e.,
\begin{equation}
  \left\{ 
    \begin{array}{l}
        \frac{\text{d}}{\text{d}\tau}{\bfPhi}(\tau,\tau_0)   = \bfF_{\tau}{\bfPhi}(\tau,\tau_0)         \\
        \frac{\text{d}}{\text{d}\tau}{\bfPhi}^*(\tau,\tau_0) = \bfF_{\tau}^*{\bfPhi}^*(\tau,\tau_0)    \\
        \bfF^*_\tau             = \dot{\bfT}_\tau \bfT^{-1}_\tau+\bfT_\tau \bfF_\tau  \bfT^{-1}_\tau
    \end{array}
  \right.
\end{equation}
we can verify that 
% \begin{align}
%     \frac{\text{d}}{\text{d}\tau}  {\bfPhi}^*(\tau,\tau_0) = \bfF_{\tau}^*{\bfPhi}^*(\tau,\tau_0) 
% \end{align}
\begin{subequations}
    \begin{align}
        \frac{\text{d}}{\text{d}\tau}  \Big(\bfT_\tau {\bfPhi}(\tau,\tau_0) \bfT_{\tau_0}^{-1}\Big) &= \dot\bfT_\tau {\bfPhi}(\tau,\tau_0)\bfT_k^{-1} +  
        \bfT_\tau \bfF_{\tau}{\bfPhi}(\tau,\tau_0) \bfT_{\tau_0}^{-1}\\
        & = \Big( \dot\bfT_\tau \bfT_{\tau}^{-1} +  \bfT_\tau \bfF_{\tau} \bfT_{\tau}^{-1}\Big)
        \bfT_\tau {\bfPhi}(\tau,\tau_0) \bfT_{\tau_0}^{-1} \\
        & =  \bfF_{\tau}^*{\bfPhi}^*(\tau,\tau_0) \\
        &= \frac{\text{d}}{\text{d}\tau}{\bfPhi}^*(\tau,\tau_0).
    \end{align} 
    \label{equ_sup:phi2*}%
\end{subequations} 
According to \eqref{equ_sup:phi2*} and the initial condition ${\bfPhi}^*(\tau_0,\tau_0) = \bfT_{\tau_0} {\bfPhi}(\tau_0,\tau_0) \bfT_{\tau_0}^{-1} = \bfI_{15+3m}$, we have:
\begin{equation}
    {\bfPhi}^*(\tau,\tau_0) = \bfT_\tau { \bfPhi}(\tau,\tau_0) \bfT_{\tau_0}^{-1} 
    \label{equ_sup:relation1}
\end{equation}

Similarly, from 
\begin{equation}
    \left\{ 
      \begin{array}{l}
        \bfQ_{k} = \int_{\tau_0}^{\tau_{p}} \bfPhi  (\tau_p,\tau) \bfG_{\tau} \bfQ {\bfG_{\tau} }^\top {\bfPhi (\tau_p,\tau)}^\top \text{d} \tau \\
        \bfQ_{k}^* =  \int_{\tau_0}^{\tau_{p}} \bfPhi ^* (\tau_p,\tau) \bfG^*_{\tau} \bfQ {\bfG^*_{\tau} }^\top {\bfPhi ^*(\tau_p,\tau)}^\top \text{d} \tau\\
          \bfG_\tau^* = \bfT_\tau \bfG_\tau
      \end{array}
    \right.
  \end{equation}
we can imply that 
\begin{subequations}
    \begin{align}
        \bfQ_{k}^* & = \int_{\tau_0}^{\tau_{p}} \bfPhi ^* (\tau_{p},\tau) \bfG^*_{\tau} \bfQ {\bfG^*_{\tau} }^\top {\bfPhi ^*(\tau_{p},\tau)}^\top \text{d} \tau  \\
        & = \int_{\tau_0}^{\tau_{p}}  \bfT_{\tau_p}  \bfPhi  (\tau_{p},\tau)  \bfT_{\tau}^{-1}  \bfT_\tau\bfG_{\tau} \bfQ {\bfG_{\tau} }^\top \bfT_{\tau}^{\top} \bfT_{\tau}^{-\top} {\bfPhi (\tau_{p},\tau)}^\top \bfT_{\tau_p}^{\top} \text{d} \tau \\
        & =\bfT_{\tau_p}  \left(\int_{\tau_0}^{\tau_{p}}    \bfPhi  (\tau_{p},\tau)  \bfG_{\tau} \bfQ {\bfG_{\tau} }^\top  {\bfPhi (\tau_{p},\tau)}^\top  \text{d} \tau \right)\bfT_{\tau_p}^{\top} \\
        & = \bfT_{\tau_p}\bfQ_{k} \bfT_{\tau_p}^{\top}
    \end{align} 
    \label{equ_sup:relation2}
\end{subequations}

Moreover, $\bfPhi_k$ and $\bfQ_k$ are diagonal block matrices,
\begin{align}
    \bfPhi_k & = \begin{bmatrix}
        \bfPhi^I_k &\bf0 \\
        \bf0 &\bfI_{3m}
    \end{bmatrix} \\
    \bfQ_k & = \begin{bmatrix}
        \bfQ_k^I &\bf0\\
        \bf0 &\bfZo_{3m\times 3m}
    \end{bmatrix}
\end{align}
where  $\bfQ_{k}^I\in \mathbb{R}^{15\times 15}$ and $\bfPhi^I_{k}  \in \mathbb{R}^{15\times 15}$  the IMU transition matrix and the IMU accumulated noise matrix. Substituting above two equations to \eqref{equ_sup:relation1} and \eqref{equ_sup:relation2} 
and replacing $(\bfT_{t_0}, \bfT_{t_p})$ with $\bfT(\hat{\bfx}_{k|k}, \bfT(\hat{\bfx}_{k+1|k}))$
yield \eqref{equ_sup:38} and \eqref{equ_sup:39}.

\section{T-ESKF state update equation}
\label{app:E}
\subsection{Update with \eqref{equ_sup:tekf0}\label{app:E1}: $\hat{\bfx}_{k+1|k+1} = \hat{\bfx}_{k+1|k} \oplus \left(\bfT(\hat{\bfx}_{k+1|k})^{-1}\delta \bfx^*\right)$}
\begin{align}
    \hat{\bfR}_{k+1|k+1} &= \text{Exp}(\delta \bftheta^*)    \hat{\bfR}_{k+1|k}\\
    \hat{\bfp}_{k+1|k+1} &=   (\bfI_3 + [\delta \bftheta^*]_{\times}) \hat{\bfp}_{k+1|k} +  \delta \bfp^* \\
    \hat{\bfv}_{k+1|k+1} &=   (\bfI_3 + [\delta \bftheta^*]_{\times}) \hat{\bfv}_{k+1|k} +  \delta \bfv^* \\
    \hat{\bfb}_{g,k+1|k+1} & =  \hat{\bfb}_{g,k+1|k} + \delta \bfb_{g}^*\\
    \hat{\bfb}_{a,k+1|k+1} & =  \hat{\bfb}_{a,k+1|k} + \delta \bfb_{a}^*\\
    \hat{\bfl}_{k+1|k+1} &=   (\bfI_3 + [\delta \bftheta^*]_{\times}) \hat{\bfl}_{k+1|k} +  \delta \bfl^* .
\end{align}

\subsection{Update with \eqref{equ_sup:tekf1}\label{app:E2}: $\hat{\bfx}_{k+1|k+1} = \hat{\bfx}_{k+1|k}\oplus \left(\bfT(\hat{\bfx}_{k+1|k+1})^{-1}\delta \bfx^*\right)$}
\begin{align}
    \hat{\bfR}_{k+1|k+1} &= \text{Exp}(\delta \bftheta^*)    \hat{\bfR}_{k+1|k}\\
    \hat{\bfp}_{k+1|k+1} &=   \bfA( \hat{\bfp}_{k+1|k} +  \delta \bfp^*) \\
    \hat{\bfv}_{k+1|k+1} &=   \bfA(  \hat{\bfv}_{k+1|k} +  \delta \bfv^* )\\
    \hat{\bfb}_{g,k+1|k+1} & =  \hat{\bfb}_{g,k+1|k} + \delta \bfb_{g}^*\\
    \hat{\bfb}_{a,k+1|k+1} & =  \hat{\bfb}_{a,k+1|k} + \delta \bfb_{a}^*\\
    \hat{\bfl}_{k+1|k+1} &=   \bfA( \hat{\bfl}_{k+1|k} +  \delta \bfl^*) 
\end{align}
with 
\begin{equation}
    \bfA=\frac{\bfI_3 + [\delta \bftheta^*]_\times  +  \delta \bftheta^* {\delta \bftheta^*}^\top }{1+ {\delta \bftheta^*}^\top \delta \bftheta^* }.
\end{equation}

% \bibliography{ref_sup}

\end{document}